\documentclass[10pt,journal,cspaper,compsoc]{IEEEtran}
\ifCLASSOPTIONcompsoc
\else
\fi
\ifCLASSINFOpdf
\else
\fi

\usepackage{subfigure}
\usepackage{amssymb}
\usepackage{enumerate}
\usepackage{picinpar}

\usepackage{algorithm}
\usepackage{algorithmic}
\usepackage{subfigure}
\usepackage{amsmath}
\usepackage{mathrsfs}
\usepackage{graphicx}
\usepackage{url}
\usepackage{latexsym}
\usepackage{amsfonts}
\usepackage{makecell}
\usepackage{wrapfig}
\usepackage{multirow}
\usepackage{arydshln}
\usepackage{epsfig}
\usepackage{booktabs}
\usepackage{color}

\newcommand{\xv}{\mathbf{x}}

\newcommand{\yv}{\mathbf{y}}

\newcommand{\ytilde}{\tilde{y}}

\newcommand{\zv}{\mathbf{z}}
\newcommand{\Zv}{\mathbf{Z}}

\newcommand{\wv}{\mathbf{w}}
\newcommand{\Wv}{\mathbf{W}}

\newcommand{\diag}{\mathrm{diag}}

\newcommand{\risk}{\mathcal{R}}

\newcommand{\zvbar}{\bar{\zv}}

\newcommand{\iTrain}{\mathcal{I}_{\textrm{tr}}}

\newcommand{\defEq}{\stackrel{\textrm{def}}{=}}

\newcommand{\etav}{\boldsymbol \eta}
\newcommand{\muv}{\boldsymbol \mu}

\newcommand{\alphav}{\boldsymbol \alpha}
\newcommand{\betav}{\boldsymbol \beta}
\newcommand{\thetav}{\boldsymbol \theta}
\newcommand{\Thetav}{\boldsymbol \Theta}

\newcommand{\lambdav}{\boldsymbol \lambda }
\newcommand{\Sigmav}{\boldsymbol \Sigma }

\newcommand{\Phiv}{\boldsymbol \Phi }

\newcommand{\ep}{\mathbb{E}}

\newcommand{\KL}{\mathrm{KL}}

\newcommand{\data}{\mathcal{D}}

\def\indicator{{\mathbb I}}

\newtheorem{theorem}{\bf{Theorem}}
\newtheorem{definition}[theorem]{\bf{Definition}}

\newtheorem{lemma}[theorem]{\bf{Lemma}}

\newtheorem{remark}[theorem]{\bf{Remark}}

\begin{document}
%
\title{Discriminative Relational Topic Models}

\author{Ning~Chen,
        Jun~Zhu,~\IEEEmembership{Member,~IEEE,}
        Fei~Xia,
        and~Bo~Zhang
\IEEEcompsocitemizethanks{\IEEEcompsocthanksitem N. Chen$^{\dagger}$, J. Zhu$^{\dagger}$, F. Xia$^{\ddagger}$ and B. Zhang$^{\dagger}$ are with the Department of Computer Science and Technology, National Lab of Information Science and Technology, State Key Lab of Intelligent Technology and Systems, Tsinghua University, Beijing, 100084 China.\protect\\
E-mail: $^{\dagger}$\{ningchen, dcszj, dcszb\}@mail.tsinghua.edu.cn,\protect\\
 $^{\ddagger}$xia.fei09@gmail.com.}}

%
%

\markboth{submitted to IEEE Transaction on Pattern Analysis and Machine Intelligence, August~2013}%
{Shell \MakeLowercase{\textit{et al.}}: Bare Demo of IEEEtran.cls for Computer Society Journals}
%


\IEEEcompsoctitleabstractindextext{%
\begin{abstract}
Many scientific and engineering fields involve analyzing network data. For document networks, relational topic models (RTMs) provide a probabilistic generative process to describe both the link structure and document contents, and they have shown promise on predicting network structures and discovering latent topic representations. However, existing RTMs have limitations in both the restricted model expressiveness and incapability of dealing with imbalanced network data. 
To expand the scope and improve the inference accuracy of RTMs, this paper presents three extensions: 1) unlike the common link likelihood with a diagonal weight matrix that allows the-same-topic interactions only, we generalize it to use a full weight matrix that captures all pairwise topic interactions and is applicable to asymmetric networks; 2) instead of doing standard Bayesian inference, we perform regularized Bayesian inference (RegBayes) with a regularization parameter to deal with the imbalanced link structure issue in common real networks and improve the discriminative ability of learned latent representations; and 3) instead of doing variational approximation with strict mean-field assumptions, we present collapsed Gibbs sampling algorithms for the generalized relational topic models by exploring data augmentation without making restricting assumptions. Under the generic RegBayes framework, we carefully investigate two popular discriminative loss functions, namely, the logistic log-loss and the max-margin hinge loss. Experimental results on several real network datasets demonstrate the significance of these extensions on improving the prediction performance, and the time efficiency can be dramatically improved with a simple fast approximation method. 
\end{abstract}\vspace{-0.2cm}

}\vspace{-0.3cm}

\maketitle

\IEEEdisplaynotcompsoctitleabstractindextext

%
\IEEEpeerreviewmaketitle

\vspace{-0.3cm}
\section{Introduction}
%
%
\IEEEPARstart{M}{any} scientific and engineering fields involve analyzing large collections of data that can be well described by networks, where vertices represent entities and edges represent relationships or interactions between entities; and to name a few, such data include online social networks, communication networks, protein interaction networks, academic paper citation and coauthorship networks, etc. As the availability and scope of network data increase, statistical network analysis (SNA) has attracted a considerable amount of attention (see~\cite{Goldenberg:2010} for a comprehensive survey). Among the many tasks studied in SNA, link prediction~\cite{liben_nowell,backstrom} is a most fundamental one that attempts to estimate the link structure of networks based on partially observed links and/or entity attributes (if exist). Link prediction could provide useful predictive models for suggesting friends to social network users or citations to scientific articles.

Many link prediction methods have been proposed, including the early work on designing good similarity measures~\cite{liben_nowell} that are used to rank unobserved links and those on learning supervised classifiers with well-conceived features~\cite{Hasan:2006,Lichtenwalter:2010}. Though specific domain knowledge can be used to design effective feature representations, feature engineering is generally a labor-intensive process. In order to expand the scope and ease of applicability of machine learning methods, fast growing interests have been spent on learning feature representations from data~\cite{Bengio:2012}. Along this line, recent research on link prediction has focused on learning latent variable models, including both parametric~\cite{Hoff:02,Hoff:07,Airoldi:nips08} and nonparametric Bayesian methods~\cite{Miller:nips09,Zhu:ICML12}. Though these methods could model the network structures well, little attention has been paid to account for observed attributes of the entities, such as the text contents of papers in a citation network or the contents of web pages in a hyperlinked network. One work that accounts for both text contents and network structures is the relational topic models (RTMs)~\cite{Chang:RTM09}, an extension of latent Dirichlet allocation~(LDA)~\cite{Blei:03} to predicting link structures among documents as well as discovering their latent topic structures.


Though powerful, existing RTMs have some assumptions that could limit their applicability and inference accuracy. First, RTMs define a symmetric link likelihood model with a diagonal weight matrix that allows the-same-topic interactions only, and the symmetric nature could also make RTMs unsuitable for asymmetric networks. Second, by performing standard Bayesian inference under a generative modeling process, RTMs do not explicitly deal with the common imbalance issue in real networks, which normally have only a few observed links while most entity pairs do not have links, and the learned topic representations could be weak at predicting link structures. Finally, RTMs and other variants~\cite{LiuYan:09} apply variational methods to estimate model parameters with mean-field assumptions~\cite{Jordan:99}, which are normally too restrictive to be realistic in practice.

To address the above limitations, this paper presents discriminative relational topic models, which consist of three extensions to improving RTMs:
\begin{enumerate}
\item we relax the symmetric assumption and define the generalized relational topic models (gRTMs) with a full weight matrix that allows all pairwise topic interactions and is more suitable for asymmetric networks;
\item we perform regularized Bayesian inference (RegBayes)~\cite{Zhu:nips11} that introduces a regularization parameter to deal with the imbalance problem in common real networks;
\item we present a collapsed Gibbs sampling algorithm for gRTMs by exploring the classical ideas of data augmentation~\cite{Dempster1977,Tanner:1987,DykMeng2001}.
\end{enumerate}
Our methods are quite generic, in the sense that we can use various loss functions to learn discriminative latent representations. In this paper, we particularly focus on two types of popular loss functions, namely, logistic log-loss and max-margin hinge loss. For the max-margin loss, the resulting max-margin RTMs are themselves new contributions to the field of statistical network analysis.

For posterior inference, we present efficient Markov Chain Monte Carlo (MCMC) methods for both types of loss functions by introducing auxiliary variables. Specifically, for the logistic log-loss, we introduce a set of Polya-Gamma random variables~\cite{Polson:arXiv12}, one per training link, to derive an exact mixture representation of the logistic link likelihood; while for the max-margin hinge loss, we introduce a set of generalized inverse Gaussian variables~\cite{Devroye:book1986} to derive a mixture representation of the corresponding unnormalized pseudo-likelihood. Then, we integrate out the intermediate Dirichlet variables and derive the local conditional distributions for collapsed Gibbs sampling analytically. These ``augment-and-collapse" algorithms are simple and efficient. More importantly, they do not make any restricting assumptions on the desired posterior distribution. Experimental results on several real networks demonstrate that these extensions are important and can significantly improve the performance.

The rest paper is structured as follows. Section 2 summarizes the related work. Section 3 presents the generalized RTMs with both the log-loss and hinge loss. Section 4 presents the ``augment-and-collapse" Gibbs sampling algorithms for both types of loss functions. Section 5 presents experimental results. Finally, Section 6 concludes with future directions discussed.

\begin{table*}
\caption{Learned diagonal weight matrix of 10-topic RTM and representative words corresponding with topics.}\label{table:latentTopic-RTM}\vspace{-.1cm}
\begin{center}
\scalebox{1.3}{\setlength{\tabcolsep}{1.8pt}
       \hspace{-.7cm}\begin{tabular}{c l}
        \cline{2-2}
        {\multirow{3}{*}{\includegraphics[width=1.6in,height=1.35in]{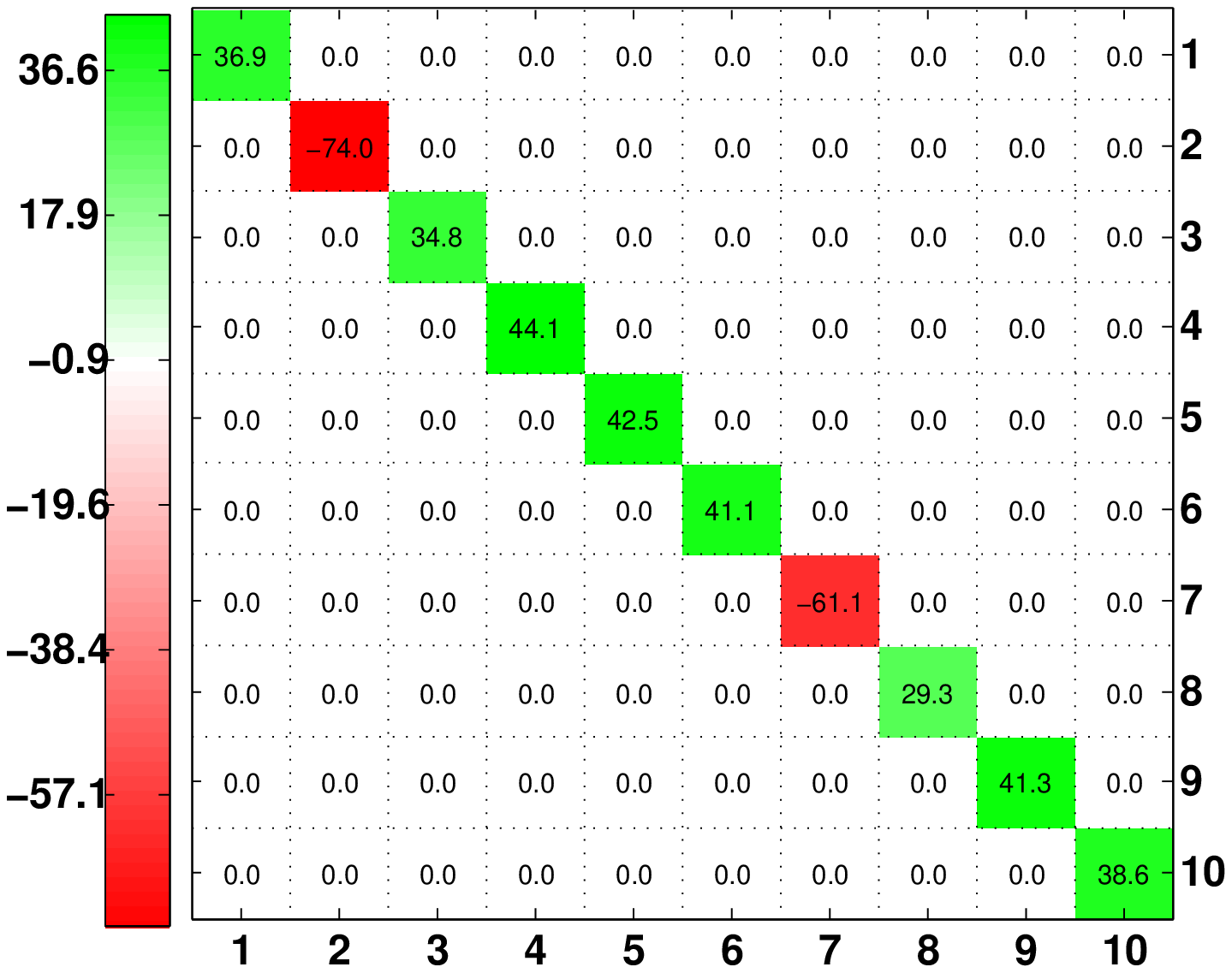}}} & {learning, bound, PAC, hypothesis, algorithm}        \\
{}& {numerical, solutions, extensions, approach, remark }\\
{}& {mixtures, experts, EM, Bayesian, probabilistic}\\
{}& {features, selection, case-based, networks, model}\\
{}& {planning, learning, acting, reinforcement, dynamic}\\
{}& {genetic, algorithm, evolving, evolutionary, learning}\\
{}& {plateau, feature, performance, sparse, networks}\\
{}& {modulo, schedule, parallelism, control, processor}\\
{}& {neural, cortical, networks, learning, feedforward}\\
{}&{markov, models, monte, carlo, Gibbs, sampler }\\
        \cline{2-2}
        \end{tabular}}
\end{center}
\end{table*}
\begin{table*}
\caption{Learned weight matrix of 10-topic gRTM and representative words corresponding with topics.}\label{table:latentTopic-gRTM}\vspace{-.1cm}
\begin{center}
\scalebox{1.3}{\setlength{\tabcolsep}{1.8pt}
       \hspace{-.7cm}\begin{tabular}{c l}
        \cline{2-2}
        \cline{2-2}
        {\multirow{3}{*}{\includegraphics[width=1.6in,height=1.35in]{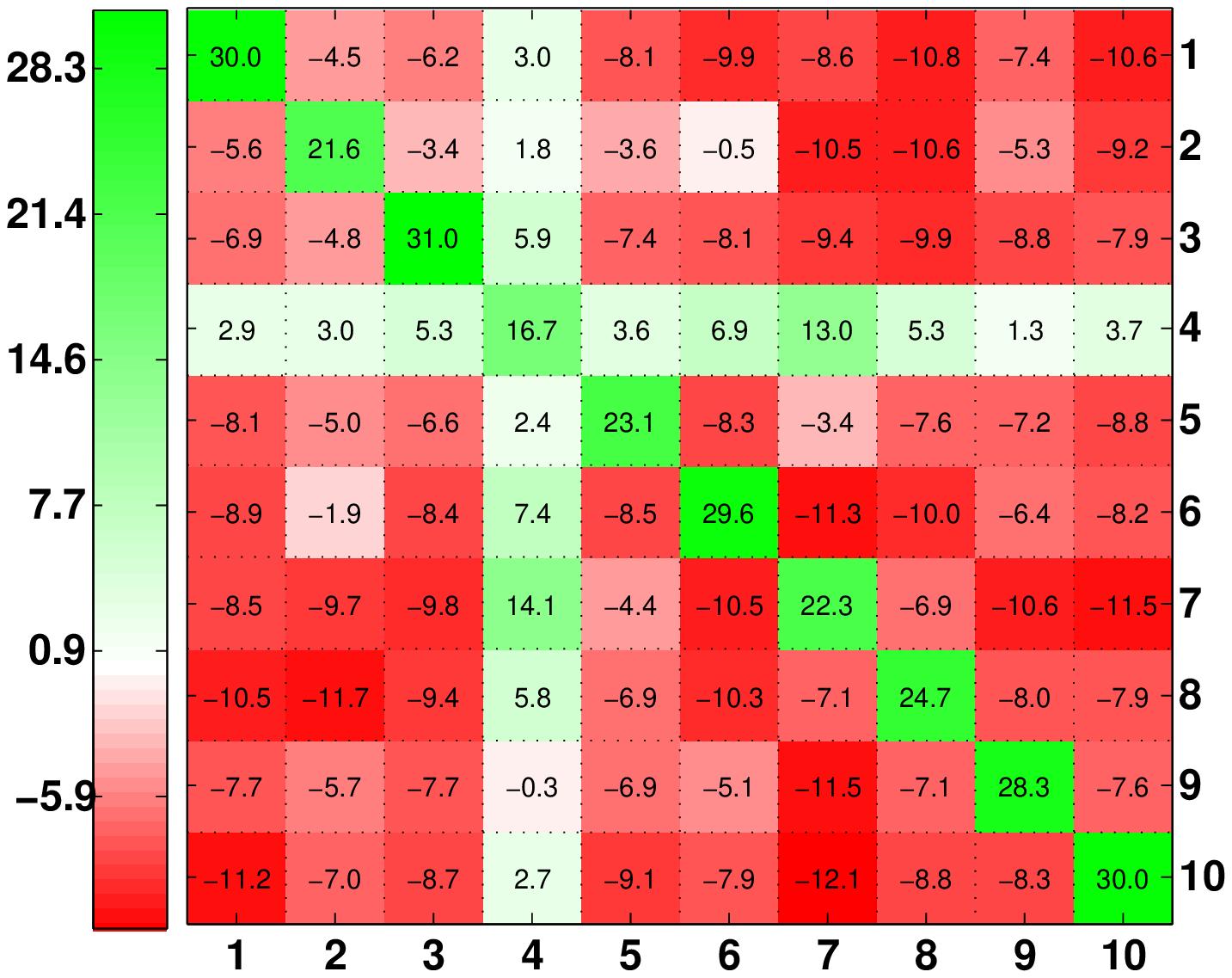}}} & {genetic, evolving, algorithm, coding, programming}        \\
{}& {logic, grammars, FOIL, EBG, knowledge, clauses}\\
{}& {reinforcement, learning, planning, act, exploration}\\
{}& {mixtures, EM, Bayesian, networks, learning, genetic}\\
{}& {images, visual, scenes, mixtures, networks, learning}\\
{}& {decision-tree, rules, induction, learning, features}\\
{}& {wake-sleep, learning, networks, cortical, inhibition}\\
{}& {monte, carlo, hastings, markov, chain, sampler}\\
{}& {case-based, reasoning, CBR, event-based, cases } \\
{}&{markov, learning, bayesian, networks, distributions}\\
        \cline{2-2}
        \cline{2-2}
        \end{tabular}}
\end{center}
\end{table*}

\vspace{-0.1cm}
\section{Related Work}


Probabilistic latent variable models, e.g., latent Dirichlet allocation (LDA)~\cite{Blei:03}, have been widely developed for modeling link relationships between documents, as they share nice properties on dealing with missing attributes as well as discovering representative latent structures. For instance, RTMs~\cite{Chang:RTM09} capture both text contents and network relations for document link prediction; Topic-Link LDA~\cite{LiuYan:09} performs topic modeling and author community discovery in one unified framework; Link-PLSA-LDA~\cite{Cohen:08} combines probabilistic latent semantic analysis (PLSA)~\cite{Hofmann:99} and LDA into a single framework to explicitly model the topical relationship between documents; Others include Pairwise-Link-LDA~\cite{Nallappati:2008}, Copycat and Citation Influence models~\cite{Dietz:2007}, Latent Topic Hypertext Models (LTHM)~\cite{Gruber:2008}, Block-LDA models~\cite{SDM:2011}, etc. One shared goal of the aforementioned models is link prediction. For static networks, our focus in this paper, this problem is usually formulated as inferring the missing links given the other observed ones. However, very few work explicitly imposes discriminative training, and many models suffer from the common imbalance issue in sparse networks (e.g., the number of unobserved links is much larger than that of the observed ones). In this paper, we build our approaches by exploring the nice framework of regularized Bayesian inference (RegBayes)~\cite{Zhu:arXiv12}, under which one could easily introduce posterior regularization and do discriminative training in a cost sensitive manner.

Another under-addressed problem in most probabilistic topic models for link prediction~\cite{Chang:RTM09,LiuYan:09} is the intractability of posterior inference due to the non-conjugacy between the prior and link likelihood (e.g., logistic likelihood). Existing approaches using variational inference with mean field assumption are often too restrictive in practice. Recently, \cite{Polson:arXiv12} and \cite{Polson:BA11} show that by making use of the ideas of data augmentation, the intractable likelihood (either a logistic likelihood or the one induced from a hinge loss) could be expressed as a marginal of a higher-dimensional distribution with augmented variables that leads to a scale mixture of Gaussian components. These strategies have been successfully explored to develop efficient Gibbs samplers for supervised topic models~\cite{Zhu:ACL13,Zhu:ICML13}. This paper further explores data augmentation techniques to do collapsed Gibbs sampling for our discriminative relational topic models. Please note that our methods could also be applied to many of the aforementioned relational latent variable models. Finally, this paper is a systematical generalization of the conference paper~\cite{Chen:ijcai13}.

\vspace{-0.1cm}
\section{Generalized RTMs}\label{section:RTM}


We consider document networks with binary link structures. Let $\mathcal{D} = \{ (\wv_i, \wv_j, y_{ij}) \}_{(i,j) \in \mathcal{I}}$ be a labeled training set, where $\wv_i = \{w_{in}\}_{n=1}^{N_i}$ denote the words within document $i$ and the response variable $y_{ij}$ takes values from the binary output space $\mathcal{Y} = \{0, 1\}$. A relational topic model (RTM) consists of two parts --- an LDA model~\cite{Blei:03} for describing the words $\Wv = \{\wv_i\}_{i=1}^D$ and a classifier for considering link structures $\yv =\{ y_{ij} \}_{(i,j) \in \mathcal{I}}$. Let $K$ be the number of topics and each topic $\Phiv_k$ is a multinomial distribution over a $V$-word vocabulary. For Bayesian RTMs, the topics are samples drawn from a prior, e.g., $\Phiv_k \sim \textrm{Dir}(\betav)$, a Dirichlet distribution. The generating process can be described as
\begin{enumerate}
\item For each document $i = 1,2, \dots, D$:
\begin{enumerate}
    \item draw a topic mixing proportion $\thetav_i \sim \mathrm{Dir}(\alphav)$
    \item for each word $n = 1,2, \dots, N_i$:
    \begin{enumerate}
        \item draw a topic assignment $z_{in} \sim \mathrm{Mult}(\thetav_i)$
        \item draw the observed word $w_{in} \sim \mathrm{Mult}(\Phiv_{z_{in}})$
    \end{enumerate}
\end{enumerate}
\item For each pair of documents $(i, j) \in \mathcal{I}$:
    \begin{enumerate}
    \item draw a link indicator $y_{ij} \sim p(.| \zv_i, \zv_j, \etav)$, where $\zv_i = \{z_{in}\}_{n=1}^{N_i}$. 
    \end{enumerate}
\end{enumerate}
We have used $\mathrm{Mult}(\cdot)$ to denote a multinomial distribution; and used $\Phiv_{z_{in}}$ to denote the topic selected by the non-zero entry of $z_{in}$, a $K$-dimensional binary vector with only one entry equaling to 1.

Previous work has defined the link likelihood as
\begin{eqnarray}
p(y_{ij} = 1 | \zv_i, \zv_j, \etav) = \sigma\left( \etav^\top(\bar{\zv}_i \circ \bar{\zv}_j) \right),
\end{eqnarray}
\noindent where $\zvbar_i = \frac{1}{N_i}\sum_{n=1}^{N_i} \zv_{in}$ is the average topic assignments of document $i$; $\sigma$ is the sigmoid function; and $\circ$ denotes elementwise product. In~\cite{Chang:RTM09}, other choices of $\sigma$ such as the exponential function and the cumulative distribution function of the normal distribution were also used, as long as it is a monotonically increasing function with respect to the weighted inner product between $\bar{\zv}_i$ and $\bar{\zv}_j$. Here, we focus on the commonly used logistic likelihood model~\cite{Miller:nips09,LiuYan:09}, as no one has shown consistently superior performance than others.

\vspace{-0.3cm}
\subsection{The Full RTM Model}
Since $\etav^\top (\zvbar_i \circ \zvbar_j) = \zvbar_i^\top \diag(\etav) \zvbar_j$, the standard RTM learns a diagonal weight matrix which only captures the-same-topic interactions (i.e., there is a non-zero contribution to the link likelihood only when documents $i$ and $j$ have the same topic). One example of the fitted diagonal matrix on the Cora citation network~\cite{Chang:RTM09} is shown in Table~\ref{table:latentTopic-RTM}, where each row corresponds to a topic and we show the representative words for the topic at the right hand side. Due to the positiveness of the latent features (i.e., $\zvbar_i$) and the competition between the diagonal entries, some of $\eta_k$ will have positive values while some are negative. The negative interactions may conflict our intuitions of understanding a citation network, where we would expect that papers with the same topics tend to have citation links. Furthermore, by using a diagonal weight matrix, the model is symmetric, i.e., the probability of a link from document $i$ to $j$ is the same as the probability of a link from $j$ to $i$. The symmetry property does not hold for many networks, e.g., citation networks.

To make RTMs more expressive and applicable to asymmetric networks, the first simple extension is to define the link likelihood as
\begin{eqnarray}\label{eq:gRTM-lhood}
p(y_{ij} = 1 | \zv_i, \zv_j, U) = \sigma\left( \zvbar_i^\top U \zvbar_j \right), 
\end{eqnarray}
using a full $K \times K$ weight matrix $U$. Using the algorithm to be presented, an example of the learned $U$ matrix on the same Cora citation network is shown in Table~\ref{table:latentTopic-gRTM}. We can see that by allowing all pairwise topic interactions, all the diagonal entries are positive, while most off-diagonal entries are negative. This is consistent with our intuition that documents with the same topics tend to have citation links, while documents with different topics are less likely to have citation links. We also note that there are some documents with generic topics (e.g., topic 4) that have positive link interactions with almost all others.

\vspace{-0.1cm}
\subsection{Regularized Bayesian Inference}


Given $\data$, we let $\Zv = \{\zv_i\}_{i=1}^D$ and $\Thetav = \{\thetav_i\}_{i=1}^D$ denote all the topic assignments and mixing proportions respectively. To fit RTM models, maximum likelihood estimation (MLE) has been used with an EM algorithm~\cite{Chang:RTM09}. We consider Bayesian inference~\cite{Hoff:07,Miller:nips09} to get the posterior distribution $$p(\Thetav, \Zv, \Phiv, U | \data) \propto p_0(\Thetav, \Zv, \Phiv, U) p(\data | \Zv, \Phiv, U),$$
where $p(\data | \Zv, \Phiv, U) = p(\Wv|\Zv,\Phiv) p(\yv | \Zv, U)$ is the likelihood of the observed data and $p_0(\Thetav, \Zv, \Phiv, U) = p_0(U) [\prod_i p(\thetav_i | \alphav) \prod_n p(z_{in}|\thetav_i)] \prod_k p(\Phiv_k|\betav)$ is the prior distribution defined by the model. One common issue with this estimation is that real networks are highly imbalanced---the number of positive links is much smaller than the number of negative links. For example, less than $0.1\%$ document pairs in the Cora network have positive links.

To deal with this imbalance issue, we propose to do regularized Bayesian inference (RegBayes)~\cite{Zhu:nips11} which offers an extra freedom to handle the imbalance issue in a cost-sensitive manner. Specifically, we define a Gibbs classifier for binary links as follows.

\begin{enumerate}
\item {\bf A Latent Predictor}: If the weight matrix $U$ and topic assignments $\Zv$ are given, we build a classifier using the likelihood~(\ref{eq:gRTM-lhood}) and the {\it latent} prediction rule is
\begin{eqnarray}\label{eq:predict-rule}
\hat{y}_{ij}|_{ \zv_i, \zv_j, U } 
= \indicator( \zvbar_i^\top U \zvbar_j > 0),
\end{eqnarray}
where $\indicator(\cdot)$ is an indicator function that equals to 1 if predicate holds otherwise 0. Then, the training error of this latent prediction rule is
$$\textrm{Err}(U, \Zv) = \sum_{(i,j) \in \mathcal{I}} \indicator( y_{ij}  \neq  \hat{y}_{ij}|{ \zv_i, \zv_j, U }  ). $$
Since directly optimizing the training error is hard, a convex surrogate loss is commonly used in machine learning. Here, we consider two popular examples, namely, the logistic log-loss and the hinge loss
\begin{eqnarray}
\risk_1(U, \Zv) &=& -\sum_{(i,j) \in \mathcal{I}} \log p(y_{ij} | \zv_i, \zv_j, U), \nonumber \\
\risk_2(U, \Zv) &=& \sum_{(i,j) \in \mathcal{I}} \max\left( 0, \ell - \ytilde_{ij} \zv_i^\top U \zv_j \right), \nonumber
\end{eqnarray}
where $\ell (\geq 1)$ is a cost parameter that penalizes a wrong prediction and $\ytilde_{ij} = 2 y_{ij} - 1$ is a transformation of the $0/1$ binary links to be $-1/+1$ for notation convenience.

\item {\bf Expected Loss}: Since both $U$ and $\Zv$ are hidden variables, we infer a posterior distribution $q(U, \Zv)$ that has the minimal expected loss
\begin{eqnarray}
\risk_1(q(U, \Zv)) &=& \ep_q\left[ \risk_1(U, \Zv) \right] \\ 
\risk_2(q(U, \Zv)) &=& \ep_q\left[ \risk_2(U, \Zv) \right]. \label{eq:ExpHingeLoss} 
\end{eqnarray}
\end{enumerate}

\begin{remark} Note that both loss functions $\risk_1(U, \Zv)$ and $\risk_2(U, \Zv)$ are convex over the parameters $U$ when the latent topics $\Zv$ are fixed. The hinge loss is an upper bound of the training error, while the log-loss is not. Many comparisons have been done in the context of classification~\cite{Rosasco:04}. Our results will provide a careful comparison of these two loss functions in the context of relational topic models.
\end{remark}

\begin{remark} Both $\risk_1(q(U, \Zv))$ and $\risk_2(q(U, \Zv))$ are good surrogate loss for the expected link prediction error $$\textrm{Err}(q(U, \Zv)) = \ep_q\left[ \textrm{Err}(U, \Zv) \right], $$
of a Gibbs classifier that randomly draws a model $U$ from the posterior distribution $q$ and makes predictions~\cite{McAllester2003}\cite{Germain:icml09}. The expected hinge loss $\risk_2(q(U, \Zv))$ is also an upper bound of $\textrm{Err}(q(U, \Zv))$. 
\end{remark}

With the above Gibbs classifiers, we define the generalized relational topic models (gRTM) as solving the regularized Bayesian inference problem
\setlength\arraycolsep{1pt}  \begin{eqnarray}\label{problem:sLDA}
\min_{q(U, \Thetav, \Zv, \Phiv) \in \mathcal{P} } && \mathcal{L}(q(U, \Thetav, \Zv, \Phiv)) + c \risk(q(U, \Zv)) 
\end{eqnarray}
where $\mathcal{L}(q) = \mathrm{KL}(q(U, \Thetav, \Zv, \Phiv)||p_0(U, \Thetav, \Zv, \Phiv)) - \ep_q[\log p(\Wv | \Zv, \Phiv)]$ is an information theoretical objective; $c$ is a positive regularization parameter controlling the influence from link structures; and $\mathcal{P}$ is the space of normalized distributions. In fact, minimizing the single term of $\mathcal{L}(q)$ results in the posterior distribution of the vanilla LDA without considering link information. For the second term, we have used $\mathcal{R}$ to denote a generic loss function, which can be either the log-loss $\risk_1$ or the hinge-loss $\risk_2$ in this paper. Note that the Gibbs classifiers and the LDA likelihood are coupled by sharing the latent topic assignments $\Zv$, and the strong coupling makes it possible to learn a posterior distribution that can describe the observed words well and make accurate predictions.

To better understand the above formulation, we define the un-normalized pseudo-likelihood\footnote{Pseudo-likelihood has been used as an approximate maximum likelihood estimation procedure~\cite{Strauss:1990}. Here, we use it to denote an unnormalized likelihood of empirical data.} for links:
\setlength\arraycolsep{1pt} \begin{eqnarray}\label{eq:pseudo-likelihood}
\psi_1(y_{ij} | \zv_i, \zv_j, U) &=&  p^c(y_{ij} | \zv_i, \zv_j, U) = \frac{ e^{c y_{ij} \omega_{ij}} }{ (1 + e^{ \omega_{ij} })^c}, \\
\psi_2(y_{ij} | \zv_i, \zv_j, U) &=& \exp\left( -2c \max(0, 1-y_{ij}\omega_{ij})\right),
\end{eqnarray}
where $\omega_{ij} = \zvbar_i^\top U \zvbar_j$ is the discriminant function value. The pseudo-likelihood $\psi_1$ is un-normalized if $c \neq 1$. 
Then, the inference problem~(\ref{problem:sLDA}) can be written as
\setlength\arraycolsep{-2pt} \begin{eqnarray}\label{problem:sLDA2}
&& \min_{q(U, \Thetav, \Zv, \Phiv) \in \mathcal{P} } \mathcal{L}(q(U, \Thetav, \Zv, \Phiv)) - \ep_q\left[ \log \psi(\yv | \Zv, U) \right] 
\end{eqnarray}
where $\psi(\yv|\Zv, U) = \prod_{(i,j) \in \mathcal{I}}  \psi_1(y_{ij} | \zv_i, \zv_j, U)$ if using log-loss and $\psi(\yv|\Zv, U) = \prod_{(i,j) \in \mathcal{I}}  \psi_2(y_{ij} | \zv_i, \zv_j, U)$ if using hinge loss.

We can show that the optimum solution of problem (\ref{problem:sLDA}) or the equivalent problem (\ref{problem:sLDA2}) is the posterior distribution with link information
\setlength\arraycolsep{1pt} \begin{eqnarray}
q(U, \Thetav, \Zv, \Phiv) = \frac{p_0(U, \Thetav, \Zv, \Phiv) p(\Wv|\Zv,\Phiv) \psi(\yv| \Zv, U)}{\phi(\yv, \Wv)}. \nonumber
\end{eqnarray}
where $\phi(\yv, \Wv)$ is the normalization constant to make $q$ as a normalized distribution.

Therefore, by solving problem~(\ref{problem:sLDA}) or (\ref{problem:sLDA2}) we are in fact doing Bayesian inference with a generalized pseudo-likelihood, which is a powered version of the likelihood~(\ref{eq:gRTM-lhood}) in the case of using the log-loss. The flexibility of using regularization parameters can play a significant role in dealing with imbalanced network data as we shall see in the experiments. For example, we can use a larger $c$ value for the sparse positive links, while using a smaller $c$ for the dense negative links. This simple strategy has been shown effective in learning classifiers~\cite{Akbani:ecml04} and link prediction models~\cite{Zhu:ICML12} with highly imbalanced data. Finally, for the logistic log-loss an ad hoc generative story can be described as in RTMs, where $c$ can be understood as the pseudo-count of a link.

\section{Augment and Collapse Sampling}\label{section:GibbsSLDA}

For gRTMs with either the log-loss or the hinge loss, exact posterior inference is intractable due to the non-conjugacy between the prior and pseudo-likelihood. Previous inference methods for the standard RTMs use variational techniques with mean-field assumptions. For example, a variational EM algorithm was developed in~\cite{Chang:RTM09} with the factorization assumption that $q(U, \Thetav, \Zv, \Phiv) = q(U) \big\lbrack \prod_i q(\thetav_i) \prod_n q(z_{in})\big\rbrack \prod_k q(\Phiv_k)$ which can be too restrictive to be realistic in practice. In this section, we present simple and efficient Gibbs sampling algorithms without any restricting assumptions on $q$. Our ``augment-and-collapse" sampling algorithm relies on a data augmentation reformulation of the RegBayes problem~(\ref{problem:sLDA2}). 

Before a full exposition of the algorithms, we summarize the high-level ideas. For the pseudo-likelihood $\psi(\yv|\Zv, U)$, it is not easy to derive a sampling algorithm directly. Instead, we develop our algorithms by introducing auxiliary variables, which lead to a scale mixture of Gaussian components and analytic conditional distributions for Bayesian inference without an accept/reject ratio. Below, we present the algorithms for the log-loss and hinge loss in turn.


\subsection{Sampling Algorithm for the Log-Loss}

For the case with the log-loss, our algorithm represents an extension of Polson et al.'s approach~\cite{Polson:arXiv12} to deal with the highly non-trivial Bayesian latent variable models for relational data analysis.
\subsubsection{Formulation with Data Augmentation}

Let us first introduce the Polya-Gamma variables~\cite{Polson:arXiv12}.
\begin{definition} A random variable $X$ has a Polya-Gamma distribution, denoted by $X \!\sim\! \mathcal{PG}(a, b)$, if
\begin{eqnarray}
X = \frac{1}{2 \pi^2}\sum_{m=1}^\infty \frac{g_m}{(m-1/2)^2 + b^2/(4\pi^2)}, \nonumber
\end{eqnarray}
\noindent where $(a>0, b\in\mathcal{R})$ are parameters and each $g_m \sim \mathcal{G}(a, 1)$ is an independent Gamma random variable.
\end{definition}

Then, using the ideas of data augmentation~\cite{Polson:arXiv12}, we have the following results
\begin{lemma} \label{lemma:SoM} The pseudo-likelihood can be expressed as
\setlength\arraycolsep{1pt} \begin{eqnarray}
\psi_1(y_{ij} | \zv_i, \zv_j, U) = \frac{1}{2^c} e^{ (\kappa_{ij} \omega_{ij} )} \! \int_0^\infty \!\! e^{( \! - \frac{\lambda_{ij} \omega_{ij}^2}{2} )} p(\lambda_{ij} | c, 0) d \lambda_{ij}, \nonumber
\end{eqnarray}
\noindent where $\kappa_{ij} = c(y_{ij} - 1/2)$ and $\lambda_{ij}$ is a Polya-Gamma variable with parameters $a=c$ and $b=0$.
\end{lemma}

Lemma~\ref{lemma:SoM} indicates that the posterior distribution of the generalized Bayesian logistic relational topic models, i.e., $q(U, \Thetav, \Zv, \Phiv)$, can be expressed as the marginal of a higher dimensional distribution that includes the augmented variables $\lambdav$. The complete posterior distribution is
\setlength\arraycolsep{1pt}\begin{eqnarray}
q(U, \lambdav, \Thetav, \Zv, \Phiv) = \frac{ p_0(U, \Thetav, \Zv, \Phiv) p(\Wv | \Zv, \Phiv) \psi(\yv, \lambdav | \Zv, U) }{\phi(\yv, \Wv)}, \nonumber
\end{eqnarray}
\noindent where $\psi(\yv, \lambdav | \Zv, U) = \prod_{(i,j) \in \mathcal{I}} \exp\big( \kappa_{ij} \omega_{ij} - \frac{\lambda_{ij} \omega_{ij}^2}{2} \big) p(\lambda_{ij} | c, 0)$ is the joint pseudo-distribution\footnote{Not normalized appropriately.} of $\yv$ and $\lambdav$.

\subsubsection{Inference with Collapsed Gibbs Sampling}


Although we can do Gibbs sampling to infer the complete posterior $q(U, \lambdav,\Thetav, \Zv, \Phiv)$ and thus $q(U, \Thetav, \Zv, \Phiv)$ by ignoring $\lambdav$, the mixing rate would be slow due to the large sample space. An effective way to reduce the sample space and improve mixing rates is to integrate out the intermediate Dirichlet variables $(\Thetav, \Phiv)$ and build a Markov chain whose equilibrium distribution is the collapsed distribution $q(U, \lambdav, \Zv)$. Such a collapsed Gibbs sampling procedure has been successfully used in LDA \cite{Griffiths:04}. For gRTMs, the collapsed posterior distribution is
\setlength\arraycolsep{1pt} \begin{eqnarray}
 q(U, \lambdav, \Zv)  && \propto  p_0(U) p(\Wv, \Zv|\alphav, \betav) \psi(\yv, \lambdav|\Zv, U)  \nonumber \\
             && = p_0(U) \prod_{k=1}^{K}\frac{\delta(\mathbf{C}_k + \betav)}{\delta(\betav)}  \prod_{i=1}^{D}  \frac{\delta(\mathbf{C}_i + \alphav)}{\delta(\alphav)}   \nonumber \\
             && ~~~ \times \prod_{(i,j) \in \mathcal{I}} \exp\Big( \kappa_{ij} \omega_{ij} - \frac{\lambda_{ij} \omega_{ij}^2}{2} \Big) p(\lambda_{ij} | c, 0) , \nonumber
\end{eqnarray}
\noindent where $\delta(\xv)=\frac{\prod_{i=1}^{\mathrm{dim}(\xv)}\Gamma(x_i)}{\Gamma(\sum_{i=1}^{\mathrm{dim}(\xv)} x_i)}$, $C_k^t$ is the number of times the term $t$ being assigned to topic $k$ over the whole corpus and $\mathbf{C}_k=\{C_k^t\}_{t=1}^{V}$; $C_i^k$ is the number of times that terms are associated with topic $k$ within the $i$-th document and $\mathbf{C}_i = \{C_i^k\}_{k=1}^{K}$. Then, the conditional distributions used in collapsed Gibbs sampling are as follows.

{\bf For $U$}: for notation clarity, we define $\zvbar_{ij} = \textrm{vec}(\zvbar_i \zvbar_j^\top)$ and $\etav = \textrm{vec}(U)$, where $\textrm{vec}(A)$ is a vector concatenating the row vectors of matrix $A$. Then, we have the discriminant function value $\omega_{ij} = \etav^\top \zvbar_{ij}$. For the commonly used isotropic Gaussian prior $p_0(U) = \prod_{kk^\prime} \mathcal{N}(U_{kk^\prime}; 0, \nu^2)$, i.e., $p_0(\etav) = \prod_m^{K^2} \mathcal{N}(\eta_m; 0, \nu^2)$, we have
\begin{eqnarray}\label{eq:GibbsEta}
q(\etav | \Zv, \lambdav ) 
                        && \propto p_0(\etav) \prod_{(i,j) \in \mathcal{I}} \exp\left( \kappa_{ij} \etav^\top \zvbar_{ij} - \frac{\lambda_{ij} (\etav^\top \zvbar_{ij})^2}{2} \right) \nonumber \\
                         &&= \mathcal{N}(\etav; \muv, \Sigmav),
\end{eqnarray}
where $\Sigmav = \big( \frac{1}{\nu^2}I + \sum_{(i,j) \in \mathcal{I}} \lambda_{ij} \zvbar_{ij} \zvbar_{ij}^\top \big)^{-1}$ and $
\muv = \Sigmav \big( \sum_{(i,j) \in \mathcal{I}} \kappa_{ij} \zvbar_{ij} \big)$. Therefore, we can easily draw a sample from a $K^2$-dimensional multivariate Gaussian distribution. The inverse can be robustly done using Cholesky decomposition.
Since $K$ is normally not large, the inversion is relatively efficient, especially when the number of documents is large. We will provide empirical analysis in the experiment section. Note that for large $K$ this step can be a practical limitation. But fortunately, there are good parallel algorithms for Cholesky decomposition~\cite{ParallelCholeskyDeCompo:1986}, which can be used for applications with large $K$ values. 

{\bf For $\Zv$}: the conditional distribution of $\Zv$ is
\begin{eqnarray}
q(\Zv | U, \lambdav ) \propto \! \prod_{k=1}^{K} \! \frac{\delta(\mathbf{C}_k + \betav)}{\delta(\betav)} \! \prod_{i=1}^{D} \! \frac{\delta(\mathbf{C}_i + \alphav)}{\delta(\alphav)} \!\!\! \prod_{(i,j) \in \mathcal{I}} \!\! \psi_1(y_{ij} | \lambdav, \Zv) \nonumber
\end{eqnarray}
\noindent where $\psi_1(y_{ij} | \lambdav, \Zv) = \exp( \kappa_{ij} \omega_{ij} - \frac{\lambda_{ij} \omega_{ij}^2}{2} )$. By canceling common factors, we can derive the local conditional of one variable $z_{in}$ given others $\Zv_{\neg}$ as:
\setlength\arraycolsep{0pt} \begin{eqnarray}\label{eqn:transitionProb}
 q(z_{in}^k = 1 &&| \Zv_{\neg}, U, \lambdav, w_{in}=t ) \nonumber \\
&& \propto \frac{ (C_{k,\neg n}^{t}+\beta_t) (C_{i,\neg n}^{k}+\alpha_k) }{\sum_t C_{k,\neg n}^t + \sum_{t=1}^V \beta_t} \nonumber \\
&& ~~\times \prod_{j \in \mathcal{N}_i^+}  \psi_1(y_{ij}|\lambdav, \Zv_\neg, z_{in}^k=1) \nonumber \\
&& ~~ \times \prod_{j \in \mathcal{N}_i^-} \psi_1(y_{ji} | \lambdav, \Zv_\neg, z_{in}^k=1),
\end{eqnarray}
\noindent where $C_{\cdot,\neg n}^{\cdot}$ indicates that term $n$ is excluded from the corresponding document or topic; and $\mathcal{N}_i^+ = \{j: (i,j) \in \mathcal{I} \} $ and $\mathcal{N}_i^- = \{ j: (j, i) \in \mathcal{I} \}$ denote the neighbors of document $i$ in the training network. For symmetric networks, $\mathcal{N}_i^+ = \mathcal{N}_i^-$, only one part is sufficient. We can see that the first term is from the LDA model for observed word counts and the second term is from the link structures $\yv$.

\begin{algorithm}[t]
\caption{Collapsed Gibbs Sampling Algorithm for Generalized RTMs with Logistic Log-loss}\label{alg:GibbsAlg}
\begin{algorithmic}[1]
   \STATE {\bfseries Initialization:} set $\lambdav = 1$ and randomly draw $z_{dn}$ from a uniform distribution.
   \FOR{$m=1$ {\bfseries to} $M$}
   \STATE draw the classifier from the normal distribution~(\ref{eq:GibbsEta})
    \FOR{$i=1$ {\bfseries to} $D$}
        \FOR{each word $n$ in document $i$}
             \STATE draw the topic using distribution~(\ref{eqn:transitionProb})
        \ENDFOR
   \ENDFOR
    \FOR{$(i,j) \in \mathcal{I} $}
       \STATE draw $\lambda_{ij}$ from distribution~(\ref{eq:GibbsLambda}).
   \ENDFOR
   \ENDFOR
\end{algorithmic}
\end{algorithm}

{\bf For $\lambdav$}: the conditional distribution of the augmented variables $\lambdav$ is a Polya-Gamma distribution
\setlength\arraycolsep{1pt} \begin{eqnarray}\label{eq:GibbsLambda}
  q(\lambda_{ij} | \Zv, U) && \propto \exp\left( \! - \frac{\lambda_{ij} \omega_{ij}^2}{2} \right) p\left(\lambda_{ij} | c, 0\right) \nonumber \\
                         && =  \mathcal{PG}\left(\lambda_{ij}; c, \omega_{ij} \right).
\end{eqnarray}
\noindent The equality is achieved by using the construction definition of the general $\mathcal{PG}(a, b)$ class through an exponential tilting of the $\mathcal{PG}(a,0)$ density~\cite{Polson:arXiv12}. To draw samples from the Polya-Gamma distribution, a naive implementation using the infinite sum-of-Gamma representation is not efficient and it also involves a potentially inaccurate step of truncating the infinite sum. Here we adopt the method 
proposed in~\cite{Polson:arXiv12}, which draws the samples from the closely related exponentially tilted Jacobi distribution.

With the above conditional distributions, we can construct a Markov chain which iteratively draws samples of $\etav$ (i.e, $U$) using Eq. (\ref{eq:GibbsEta}), $\Zv$ using Eq. (\ref{eqn:transitionProb}) and $\lambdav$ using Eq. (\ref{eq:GibbsLambda}) as shown in Alg.~\ref{alg:GibbsAlg}, with an initial condition. In our experiments, we initially set $\lambdav=1$ and randomly draw $\Zv$ from a uniform distribution. In training, we run the Markov chain for $M$ iterations (i.e., the so called burn-in stage). Then, we draw a sample $\hat{U}$ as the final classifier to make predictions on testing data. As we shall see in practice, the Markov chain converges to stable prediction performance with a few burn-in iterations.


\subsection{Sampling Algorithm for the Hinge Loss}

Now, we present an ``augment-and-collapse" Gibbs sampling algorithm for the gRTMs with the hinge loss. The algorithm represents an extension of the recent techniques~\cite{Zhu:ICML13} to relational data analysis.

\vspace{-0.1cm}
\subsubsection{Formula with Data Augmentation}

As we do not have a closed-form of the expected margin loss, it is hard to deal with the expected hinge loss in Eq.~(\ref{eq:ExpHingeLoss}). Here, we develop a collapsed Gibbs sampling method based on a data augmentation formulation of the expected margin loss to infer the posterior distribution
\begin{eqnarray}
q(U, \Thetav, \Zv, \Phiv) = \frac{p_0(U, \Thetav, \Zv, \Phi) p(\Wv| \Zv,\Phi) \psi(\yv|\Zv, U)}{\phi(\yv, \Wv)}, \nonumber
\end{eqnarray}
where $\phi(\yv, \Wv)$ is the normalization constant and $\psi(\yv | \Zv, U) = \prod_{(i,j) \in \mathcal{I}} \psi_2 (y_{ij}|\zv_i, \zv_j, U)$ in this case. Specifically, we have the following data augmentation representation of the pseudo-likelihood: \begin{eqnarray}\label{eq:dataaug}
&& \psi_2(y_{ij}| \zv_i, \zv_j, U ) \nonumber \\
&& ~~~ = \int_0^\infty \! \frac{1}{\sqrt{2\pi\lambda_{ij}}} \exp\Big\{\! -\frac{(\lambda_{ij} + c\zeta_{ij})^2}{2\lambda_{ij}} \! \Big\} d\lambda_{ij},
\end{eqnarray}
where $\zeta_{ij} = \ell - y_{ij} \omega_{ij}$. Eq.~(\ref{eq:dataaug}) can be derived following \cite{Polson:BA11}, and it indicates that the posterior distribution $q(U, \Thetav, \Zv, \Phiv)$ can be expressed as the marginal of a higher dimensional posterior distribution that includes the augmented variables $\lambdav$:
\begin{eqnarray}
q(U, \lambdav, \Thetav, \Zv, \Phiv) = \frac{p_0(U, \Thetav, \Zv, \Phiv) p(\Wv | \Zv, \Phiv) \psi(\yv, \lambdav|\Zv, U)}{\phi(\yv, \Wv)}, \nonumber
\end{eqnarray}
where the unnormalized distribution of $\yv$ and $\lambdav$ is
\begin{eqnarray}
\psi(\yv, \lambdav | \Zv, U) = \prod_{(i,j) \in \mathcal{I}} \frac{1}{\sqrt{2\pi\lambda_{ij}}} \exp\left( - \frac{(\lambda_{ij} + c\zeta_{ij})^2}{2\lambda_{ij}} \right). \nonumber
\end{eqnarray}

\vspace{-0.2cm}
\subsubsection{Inference with Collapsed Gibbs Sampling}

Similar as in the log-loss case, although we can sample the complete distribution $q(U, \lambdav, \Thetav, \Zv, \Phiv)$, the mixing rate would be slow due to the high dimensional sample space. Thus, we reduce the sample space and improve mixing rate by integrating out the intermediate Dirichlet variables $(\Thetav, \Phiv)$ and building a Markov chain whose equilibrium distribution is the resulting marginal distribution $q(U, \lambdav, \Zv)$. Specifically, the collapsed posterior distribution is
\begin{eqnarray}
q(U, \lambdav, \Zv) &\propto& p_0(U) p(\Wv, \Zv|\alphav, \betav) \prod_{i,j} \phi(y_{ij}, \lambda_{ij}|\zv_i, \zv_j, U) \nonumber\\
&=& p_0(U)\prod_{i=1}^D \frac{\delta(\mathbf{C}_i + \alphav)}{\delta(\alphav)} \times \prod_{k=1}^K\frac{\delta(\mathbf{C}_k + \betav)}{\delta(\betav)}  \nonumber \\
&& \times \prod_{(i,j)\in \mathcal{I}}\frac{1}{\sqrt{2\pi\lambda_{ij}}} \exp\Big\{ -\frac{(\lambda_{ij} + c \zeta_{ij})^2}{2\lambda_{ij}} \Big\}. \nonumber
\end{eqnarray}
Then we could get the conditional distribution using the collapsed Gibbs sampling as following:

{\bf{For $U$}}: we use the similar notations, $\etav = \textrm{vec}(U)$ and $\zvbar_{ij} = \textrm{vec}(\zvbar_i \zvbar_j^\top)$. For the commonly used isotropic Gaussian prior, $p_0(U) = \prod_{k,k^\prime} \mathcal{N}(U_{k, k^\prime}; 0, \nu^2)$, the posterior distribution of $q(U |\Zv, \lambdav)$ or $q(\etav|\Zv, \lambdav)$ is still a Gaussian distribution:
\begin{eqnarray}\label{eq:sampleU2}
q(\etav | \Zv, \lambda) &\propto& p_0(U) \prod_{(i,j)\in \mathcal{I} } \exp\Big( -\frac{(\lambda_{ij} + c\zeta_{ij})^2}{2\lambda_{ij}} \Big) \nonumber\\
&=&\mathcal{N}(\etav; \muv, \Sigmav)
\end{eqnarray}
where $\Sigmav = \big( \frac{1}{\sigma^2}I + c^2 \sum_{i,j} \frac{\zvbar_{ij} \zvbar_{ij}^\top} {\lambda_{ij}} \big)^{-1}$ and $\muv = \Sigmav \big( c \sum_{i,j}y_{ij}\frac{(\lambda_{ij} + c \ell)}{\lambda_{ij}} \zvbar_{ij} \big)$.


{\bf{For $\Zv$}}: the conditional posterior distribution of $\Zv$ is
\begin{eqnarray}\label{eq:sampleZ}
q(\Zv | U, \lambdav) \propto \! \prod_{i=1}^D \! \frac{\delta(\mathbf{C}_i + \alphav)}{\delta(\alphav)} \! \prod_{k=1}^K \! \frac{ \delta(\mathbf{C}_k + \betav)}{\delta(\betav)} \!\!\! \prod_{(i,j) \in \mathcal{I}} \!\!\! \psi_2(y_{ij} | \lambdav, \Zv), \nonumber
\end{eqnarray}
where $\psi_2(y_{ij} | \lambdav, \Zv) = \exp (  -\frac{(\lambda_{ij} + c \zeta_{ij})^2}{2\lambda_{ij}} )$. By canceling common factors, we can derive the local conditional of one variable $z_{in}$ given others $\Zv_{\neg}$ as:
\setlength\arraycolsep{0pt} \begin{eqnarray}\label{eqn:transitionProb2}
 q(z_{in}^k = 1 &&| \Zv_{\neg}, U, \lambdav, w_{in}=t ) \nonumber \\
&& \propto \frac{ (C_{k,\neg n}^{t}+\beta_t) (C_{i,\neg n}^{k}+\alpha_k) }{\sum_t C_{k,\neg n}^t + \sum_{t=1}^V \beta_t} \nonumber \\
&& ~~\times \prod_{j \in \mathcal{N}_i^+}  \psi_2(y_{ij}|\lambdav, \Zv_\neg, z_{in}^k=1) \nonumber \\
&& ~~ \times \prod_{j \in \mathcal{N}_i^-} \psi_2(y_{ji} | \lambdav, \Zv_\neg, z_{in}^k=1).
\end{eqnarray}
Again, we can see that the first term is from the LDA model for observed word counts and the second term is from the link structures $\yv$.

{\bf{For $\lambdav$}}: due to the independence structure among the augmented variables when $\Zv$ and $U$ are given, we can derive the conditional posterior distribution of each augmented variable $\lambda_{ij}$ as:
\begin{eqnarray}\label{eq:sampleLambda}
q(\lambda_{ij} | \Zv, U) &\propto& \frac{1}{\sqrt{2\pi\lambda_{ij}}} \exp \left( -\frac{(\lambda_{ij} + c \zeta_{ij})^2}{2\lambda_{ij}} \right) \nonumber\\
& = & \mathcal{GIG}\left(\lambda_{ij}; \frac{1}{2}, 1, c^2\zeta_{ij}^2\right)
\end{eqnarray}
where $\mathcal{GIG}(x; p, a,b) = C(p, a,b)x^{p-1}\exp(-\frac{1}{2}(\frac{b}{x} + a x))$ is a generalized inverse Gaussian distribution~\cite{Devroye:book1986} and $C(p,a,b)$ is a normalization constant. Therefore, we can derive that $\lambda_{ij}^{-1}$ follows an inverse Gaussian distribution $$p(\lambda_{ij}^{-1}|\Zv, U) = \mathcal{IG}\left(\lambda_{ij}^{-1}; \frac{1}{c| \zeta_{ij}|}, 1\right),$$
where $\mathcal{IG}(x; a, b) = \sqrt{ \frac{b}{2\pi x^3} }\exp(-\frac{b(x - a)^2}{2 a^2 x})$ for $a,b>0$.

With the above conditional distributions, we can construct a Markov chain which iteratively draws samples of the weights $\etav$ (i.e., $U$) using Eq. (\ref{eq:sampleU2}), the topic assignments $\Zv$ using Eq. (\ref{eqn:transitionProb2}) and the augmented variables $\lambdav$ using Eq. (\ref{eq:sampleLambda}), with an initial condition which is the same as in the case of the logistic log-loss. To sample from an inverse Gaussian distribution, we apply the efficient transformation method with multiple roots~\cite{Michael:IG76}. 

\begin{remark}
We note that the Gibbs sampling algorithms for both the hinge loss and logistic loss have a similar structure. But they have different distributions for the augmented variables. As we shall see in experiments, drawing samples from the different distributions for $\lambdav$ will have different efficiency.
\end{remark}

\subsection{Prediction}

Since gRTMs account for both text contents and network structures, we can make predictions for each of them conditioned on the other~\cite{Chang:RTM09}. For link prediction, given a test document $\wv$, we need to infer its topic assignments $\zv$ in order to apply the classifier~(\ref{eq:predict-rule}). This can be done with a collapsed Gibbs sampling method, where the conditional distribution is $$p(z_n^k=1|\zv_{\neg n}) \propto \hat{\phi}_{kw_{n}}(C_{\neg n}^{k} + \alpha_k);$$
$C_{\neg n}^{k}$ is the times that the terms in this document $\wv$ are assigned to topic $k$ with the $n$-th term excluded; and $\hat{\Phiv}$ is a point estimate of the topics, with $\hat{\phi}_{kt} \propto {C_k^t} + \beta_t$. To initialize, we randomly set each word to a topic, and then run the Gibbs sampler until some stopping criterion is met, e.g., the relative change of likelihood is less than a threshold (e.g., 1e-4 in our experiments).

For word prediction, we need to infer the distribution $$p(w_n|\yv, \data, \hat{\Phiv}, \hat{U}) = \sum_{k} \hat{\phi}_{k w_n} p(z_n^k=1 | \yv, \data, \hat{U}).$$ This can be done by drawing a few samples of $z_n$ and compute the empirical mean of $\hat{\phi}_{k w_n}$ using the sampled $z_n$. The number of samples is determined by running a Gibbs sampler until some stopping criterion is met, e.g., the relative change of likelihood is less than 1e-4 in our experiments.

\vspace{-0.1cm}
\section{Experiments}

Now, we present both quantitative and qualitative results on several real network datasets to demonstrate the efficacy of the generalized discriminative relational topic models. We also present extensive sensitivity analysis with respect to various parameters.
\vspace{-0.2cm}
\subsection{Data sets and Setups}\label{section:dataset}
We present experiments on three public datasets of document networks\footnote{http://www.cs.umd.edu/projects/linqs/projects/lbc/index.html}:
\begin{enumerate}
\item The {\it Cora} data~\cite{McCallum:cora2000} consists of abstracts of 2,708 computer science research papers, with links between documents that cite each other. In total, the Cora citation network has 5,429 positive links, and the dictionary consists of 1,433 words.
\item The {\it WebKB} data~\cite{Craven:1998} contains 877 webpages from the computer science departments of different universities, with links between webpages that are hyper-linked. In total, the WebKB network has 1,608 positive links and the dictionary has 1,703 words.
\item The {\it Citeseer} data~\cite{sen:aimag08} consists of 3,312 scientific publications with 4,732 positive links, and the dictionary contains 3,703 unique words.
\end{enumerate}

\begin{figure*}
\centering
\subfigure[link rank]{\includegraphics[height=1.3in, width=1.25in]{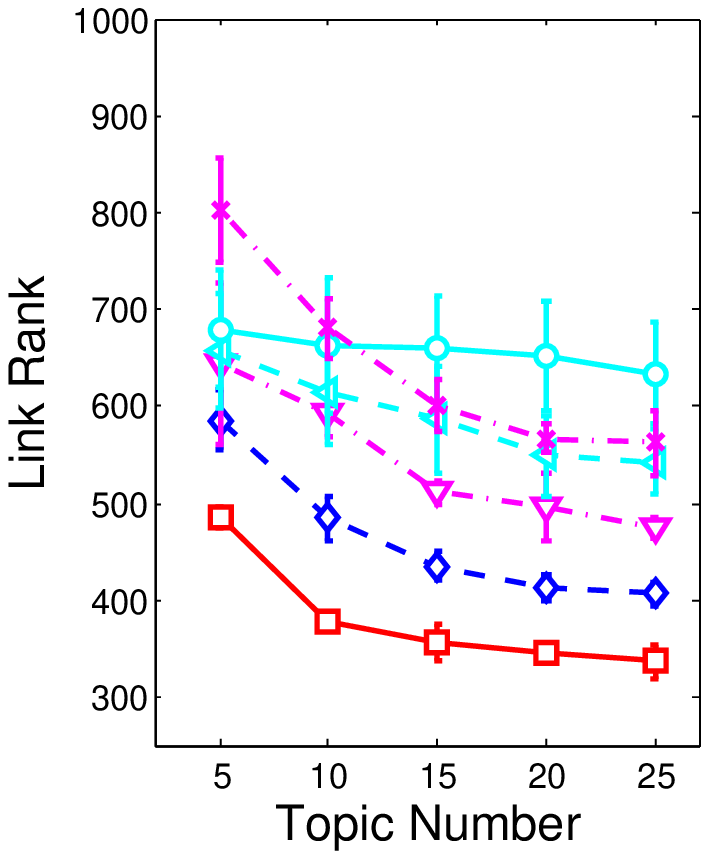}}\label{fig:cora1}
\subfigure[word rank]{\includegraphics[height=1.3in, width=1.25in]{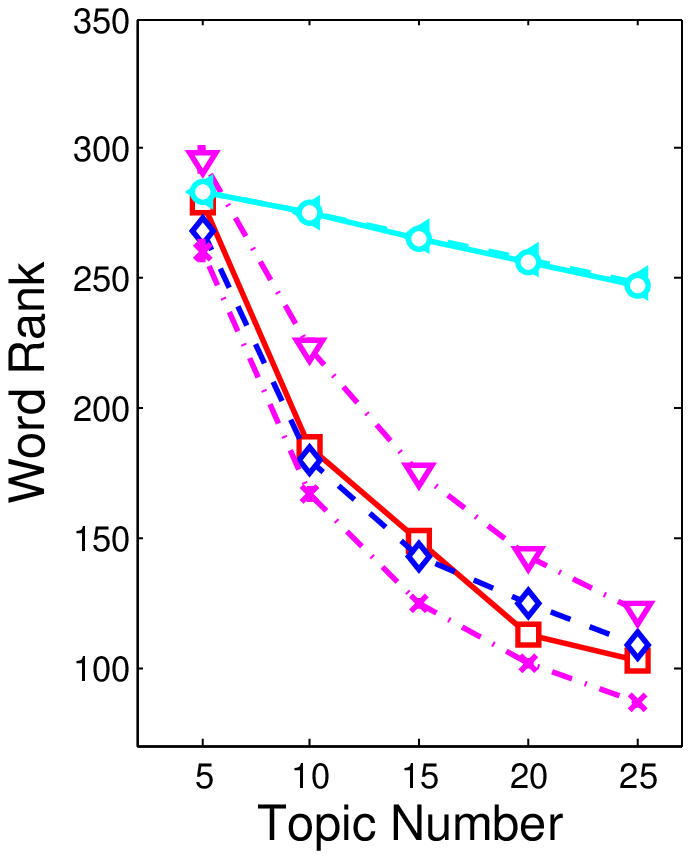}}\label{fig:cora2}
\subfigure[AUC score]{\includegraphics[height=1.3in, width=1.25in]{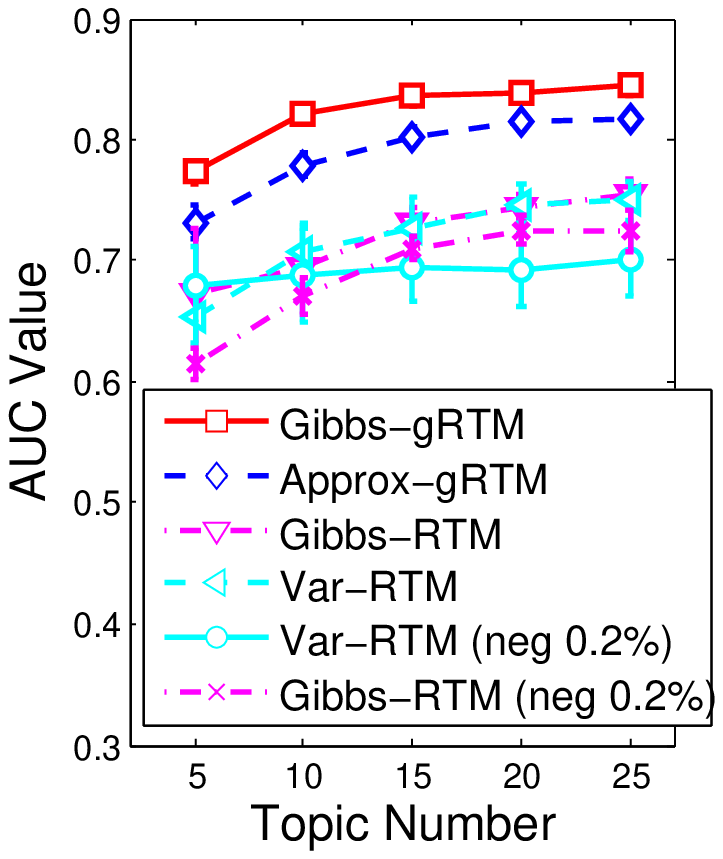}}\label{fig:cora3}
\subfigure[train time]{\includegraphics[height=1.3in, width=1.25in]{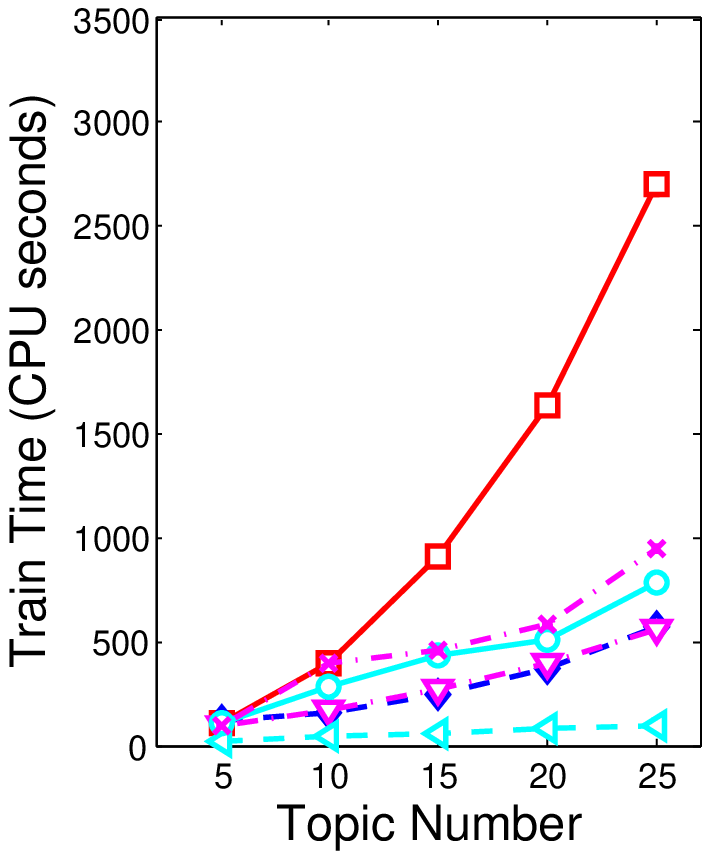}}\label{fig:cora4}
\subfigure[test time]{\includegraphics[height=1.3in, width=1.25in]{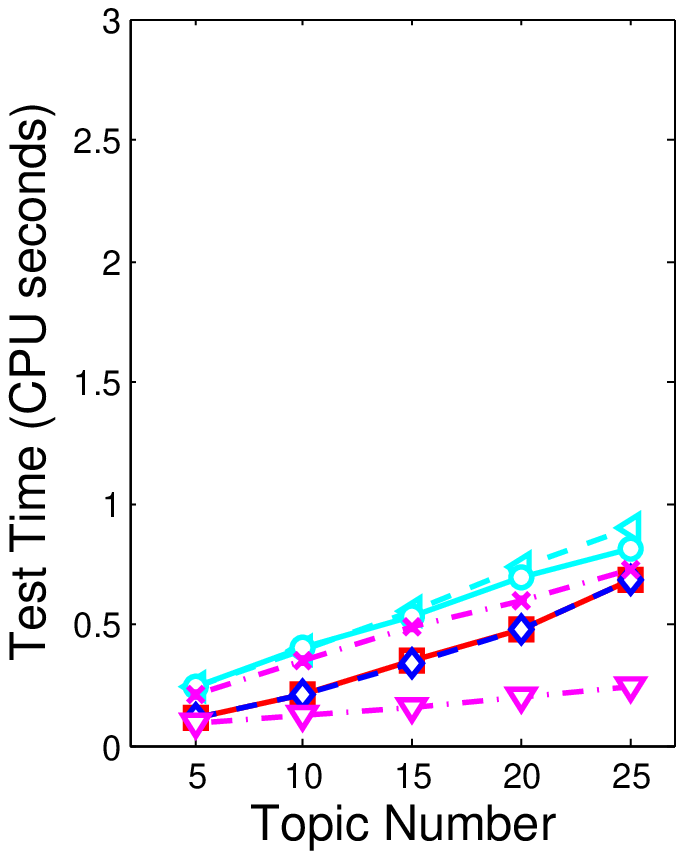}}\label{fig:cora5}
\caption{Results of various models with different numbers of topics on the Cora citation dataset.}
\label{fig:cora}
\end{figure*}
\begin{figure*}
\centering
\subfigure[link rank]{\includegraphics[height=1.3in, width=1.25in]{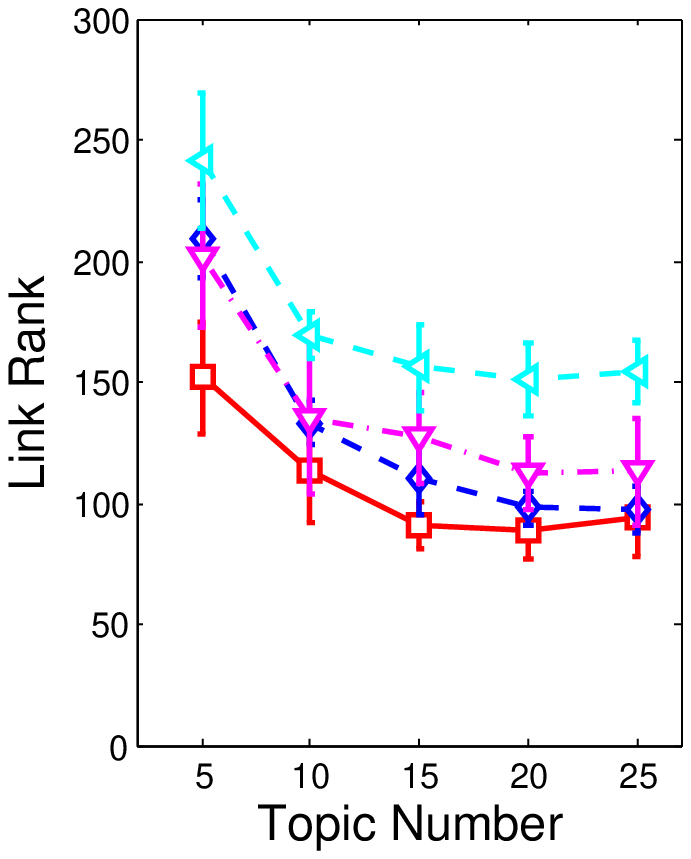}}\label{fig:web1}
\subfigure[word rank]{\includegraphics[height=1.3in, width=1.25in]{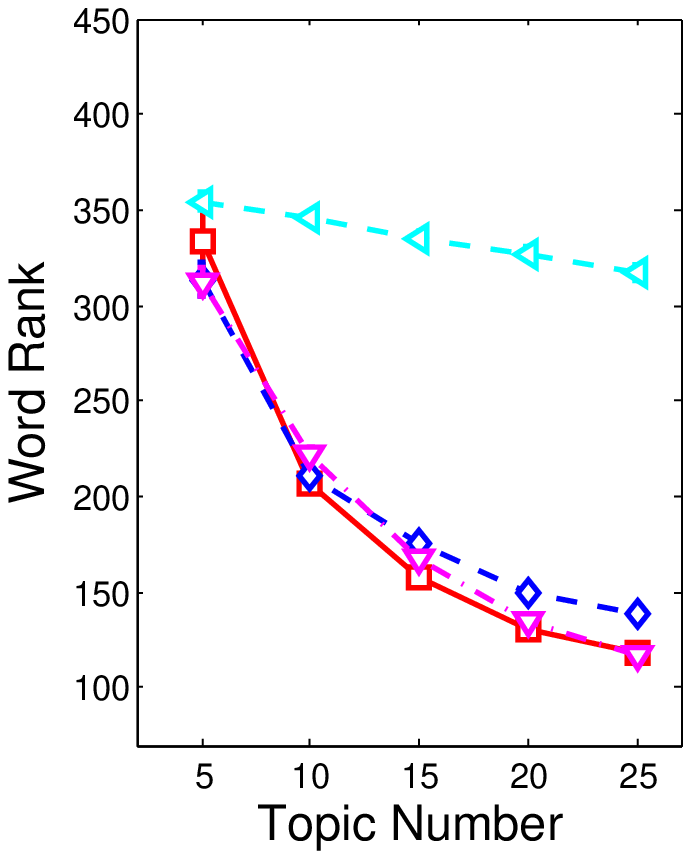}}\label{fig:web2}
\subfigure[AUC score]{\includegraphics[height=1.3in, width=1.25in]{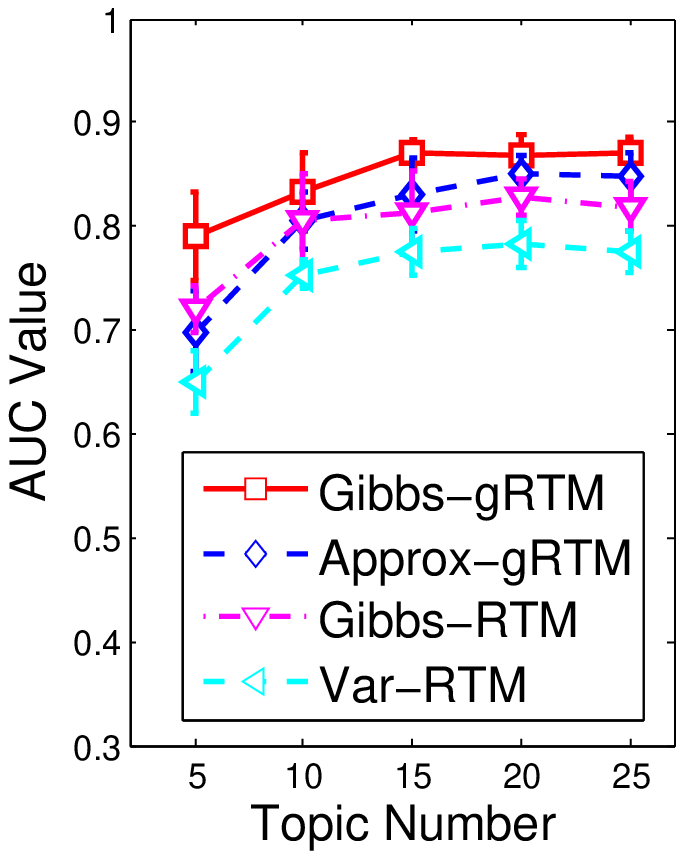}}\label{fig:web3}
\subfigure[train time]{\includegraphics[height=1.3in, width=1.25in]{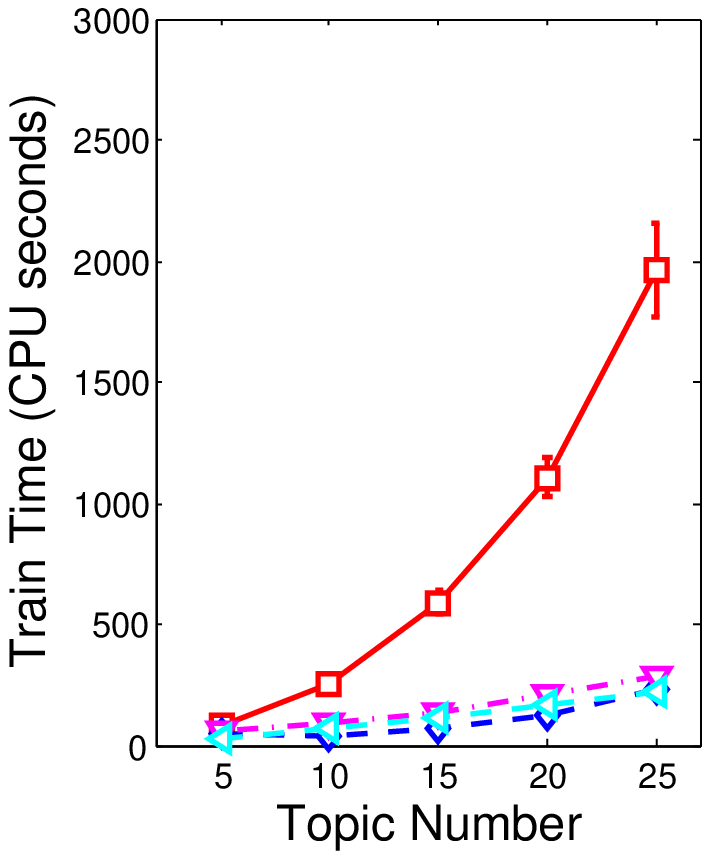}}\label{fig:web4}
\subfigure[test time]{\includegraphics[height=1.3in, width=1.25in]{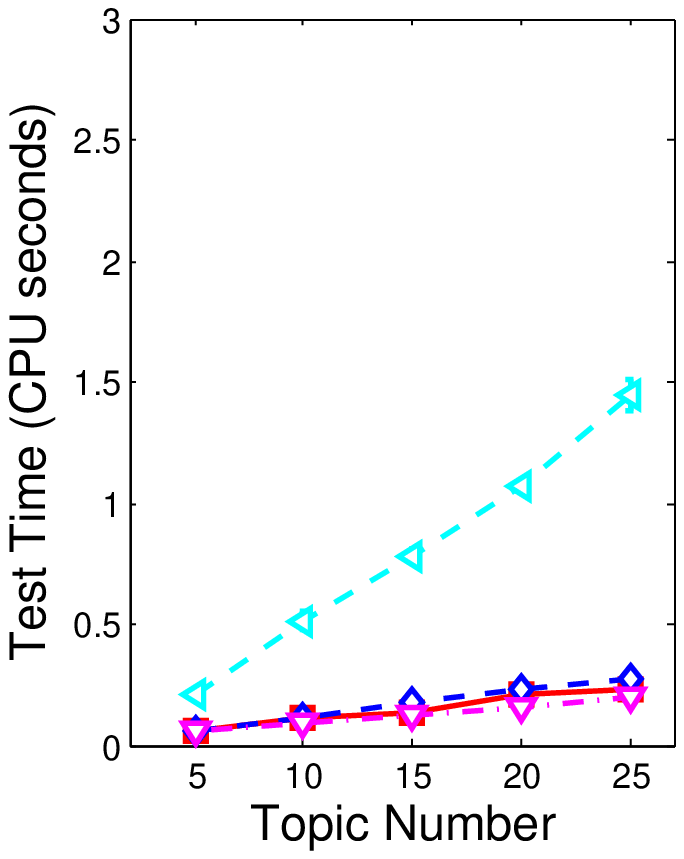}}\label{fig:web5}
\caption{Results of various models with different numbers of topics on the WebKB dataset.}
\label{fig:web}
\end{figure*}
\begin{figure*}
\centering
\subfigure[link rank]{\includegraphics[height=1.3in, width=1.25in]{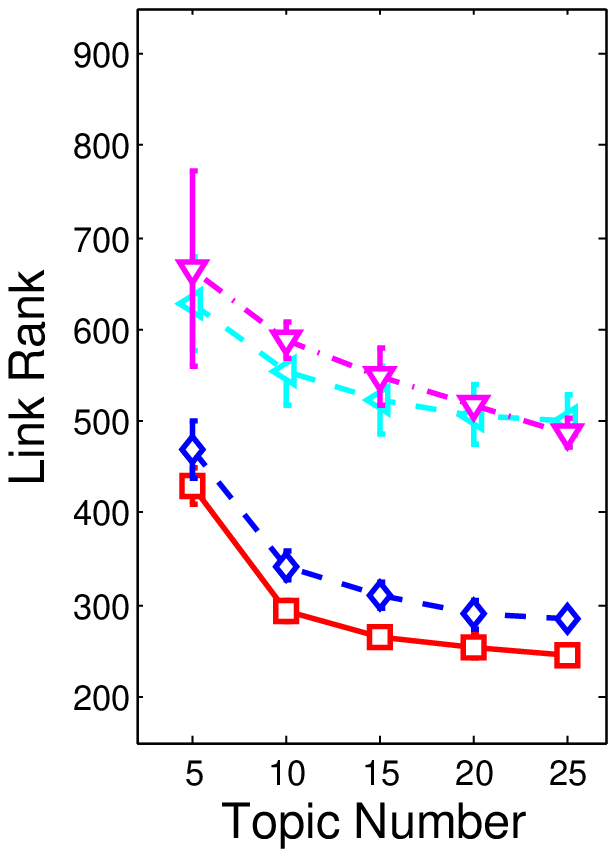}\label{fig:cite1}}
\subfigure[word rank]{\includegraphics[height=1.3in, width=1.25in]{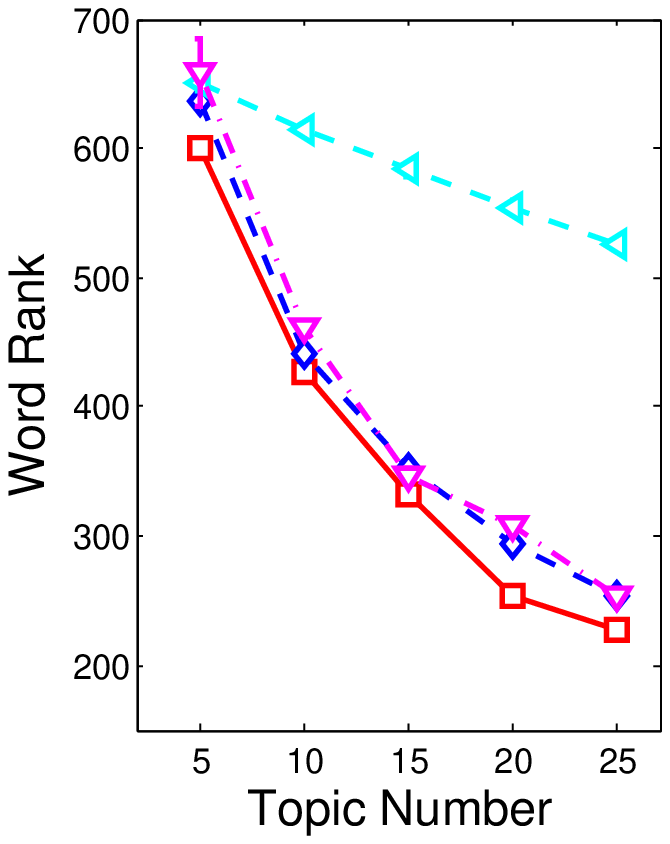}\label{fig:cite2}}
\subfigure[AUC score]{\includegraphics[height=1.3in, width=1.25in]{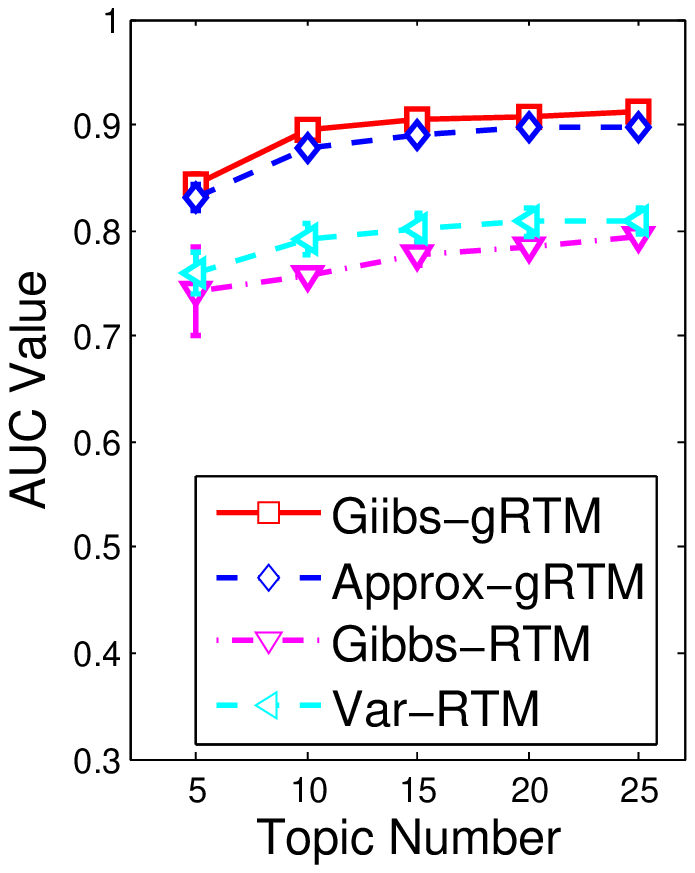}\label{fig:cite3}}
\subfigure[train time]{\includegraphics[height=1.3in, width=1.25in]{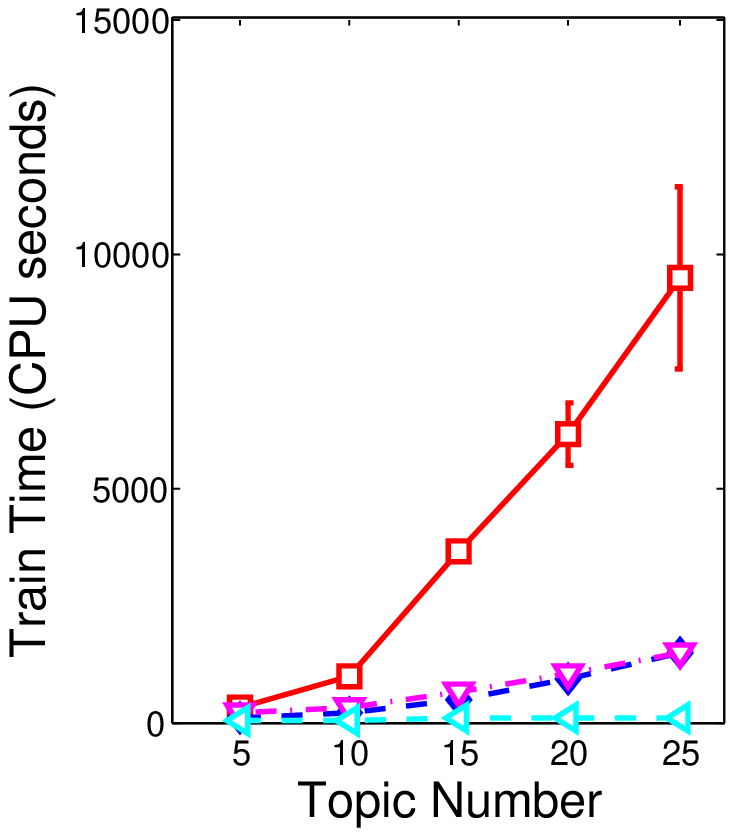}\label{fig:cite4}}
\subfigure[test time]{\includegraphics[height=1.3in, width=1.25in]{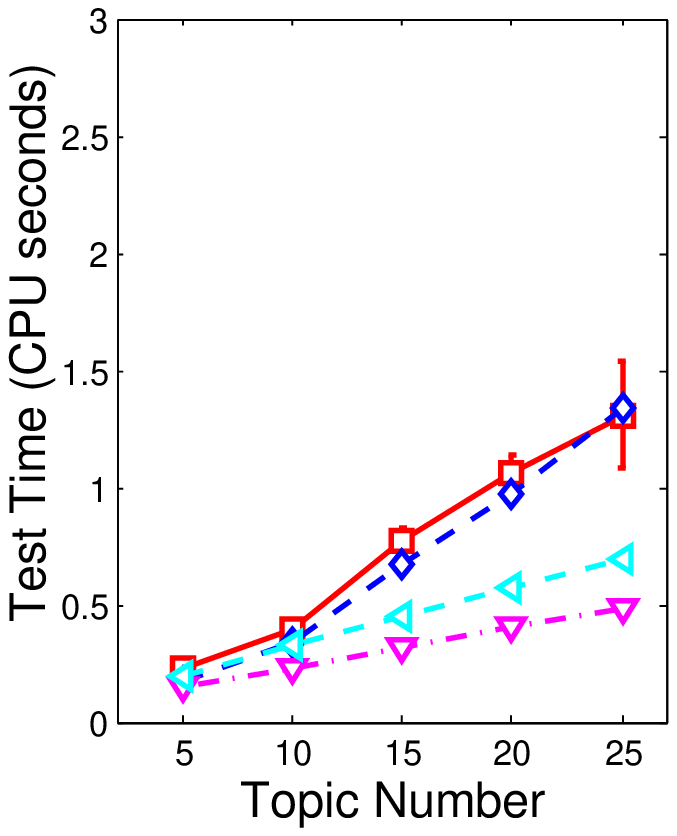}\label{fig:cite5}}
\caption{Results of various models with different numbers of topics on the Citeseer dataset.}
\label{fig:citeseer}
\end{figure*}


Since many baseline methods have been outperformed by RTMs on the same datasets~\cite{Chang:RTM09}, we focus on evaluating the effects of the various extensions in the discriminative gRTMs with log-loss (denoted by Gibbs-gRTM) and hinge loss (denoted by Gibbs-gMMRTM) by comparing with various special cases:
\begin{enumerate}
\item {\bf Var-RTM}: the standard RTMs (i.e., $c=1$) with a diagonal logistic likelihood and a variational EM algorithm with mean-field assumptions~\cite{Chang:RTM09};
\item {\bf Gibbs-RTM}: the Gibbs-RTM model with a diagonal weight matrix and a Gibbs sampling algorithm for the logistic link likelihood;
\item{\bf Gibbs-gRTM}: the Gibbs-gRTM model with a full weight matrix and a Gibbs sampling algorithm for the logistic link likelihood;
\item {\bf Approx-gRTM}: the Gibbs-gRTM model with fast approximation on sampling $\Zv$, by computing the link likelihood term in Eq. (\ref{eq:GibbsEta}) for once and caching it for sampling all the word topics in each document;
\item{\bf Gibbs-MMRTM}: the Gibbs-MMRTM model with a diagonal weight matrix and a Gibbs sampling algorithm for the hinge loss;
\item{\bf Gibbs-gMMRTM}: the Gibbs-gMMRTM model with a full weight matrix and a Gibbs sampling algorithm for the hinge loss;
\item{\bf Approx-gMMRTM}: the Gibbs-gMMRTM model with fast approximation on sampling $\Zv$, which is similar to Approx-gRTM.
\end{enumerate}

For Var-RTM, we follow the setup~\cite{Chang:RTM09} and use positive links only as training data; to deal with the one-class problem, a regularization penalty was used, which in effect injects some number of pseudo-observations (each with a fixed uniform topic distribution). For the other proposed models, including Gibbs-gRTM, Gibbs-RTM, Approx-gRTM, Gibbs-gMMRTM, Gibbs-MMRTM, and Approx-gMMRTM, we instead draw some unobserved links as negative examples. Though subsampling normally results in imbalanced datasets, the regularization parameter $c$ in our discriminative gRTMs can effectively address it, as we shall see. Here, we fix $c$ at $1$ for negative examples, while we tune it for positive examples. All the training and testing time are fairly calculated on a desktop computer with four 3.10GHz processors and 4G RAM.

\vspace{-0.2cm}
\subsection{Quantitative Results}
We first report the overall results of {\it link rank}, {\it word rank} and {\it AUC} (area under the ROC curve) to measure the prediction performance, following the setups in~\cite{Chang:RTM09}. Link rank is defined as the average rank of the observed links from the held-out test documents to the training documents, and word rank is defined as the average rank of the words in testing documents given their links to the training documents. Therefore, lower link rank and word rank are better, and higher AUC value is better. The test documents are completely new that are not observed during training. In the training phase all the words along with their links of the test documents are removed.

\vspace{-0.1cm}
\subsubsection{Results with the Log-loss}

Fig.~\ref{fig:cora}, Fig.~\ref{fig:web} and Fig.~\ref{fig:citeseer} show the 5-fold average results and standard deviations of various models on all the three datasets with varying numbers of topic. For the RTM models using collapsed Gibbs sampling, we randomly draw $1\%$ of the unobserved links as negative training examples, which lead to imbalanced training sets. We can see that the generalized Gibbs-gRTM can effectively deal with the imbalance and achieve significantly better results on link rank and AUC scores than all other competitors. For word rank, all the RTM models using Gibbs sampling perform better than the RTMs using variational EM methods when the number of topics is larger than 5.

\begin{figure*}
\centering
\subfigure[link rank]{\includegraphics[height=1.3in, width=1.25in]{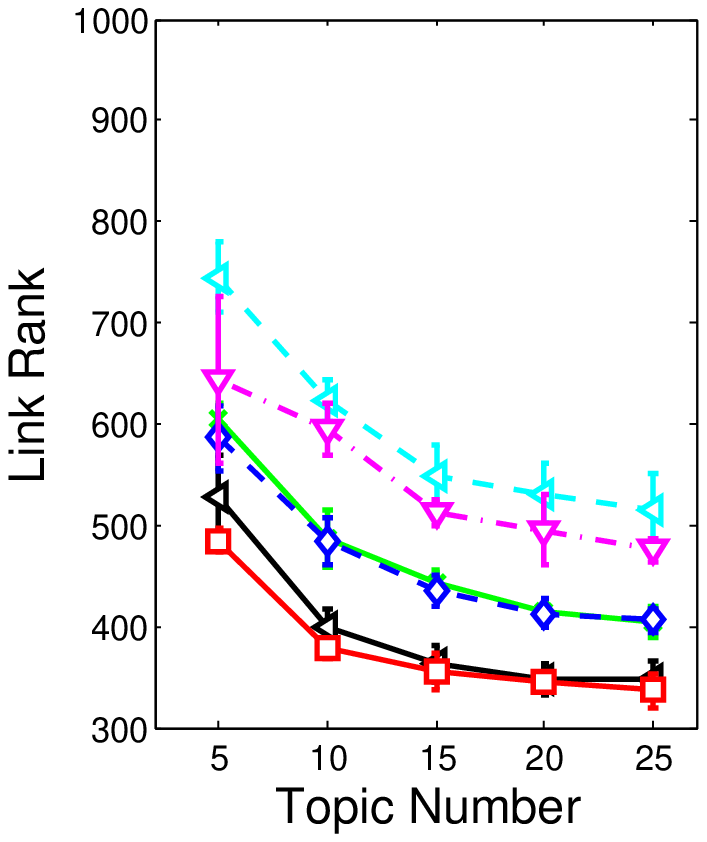}}\label{fig:cora_mmrtm}
\subfigure[word rank]{\includegraphics[height=1.3in, width=1.25in]{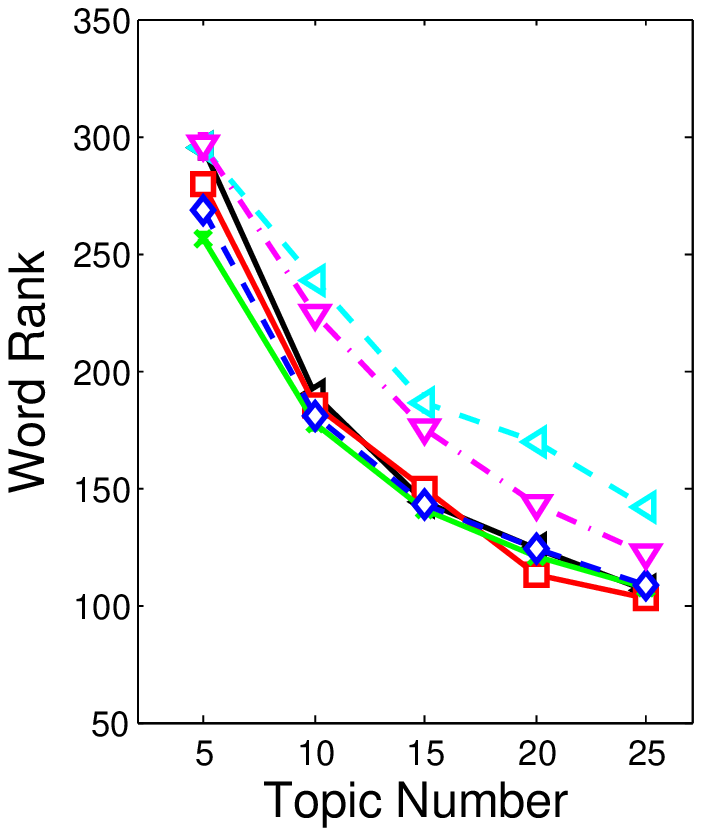}}\label{fig:cora_mmrtm}
\subfigure[AUC score]{\includegraphics[height=1.3in, width=1.25in]{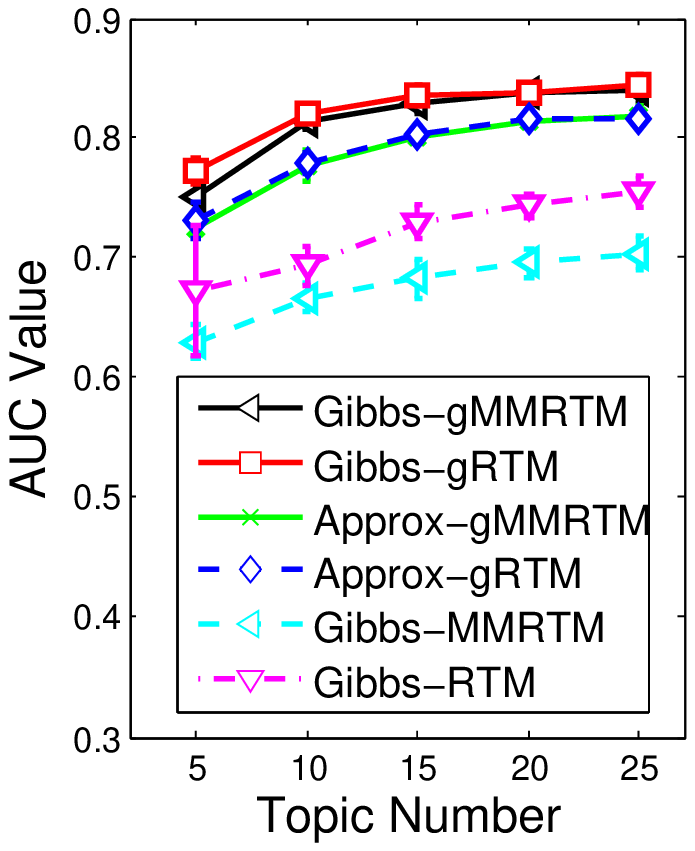}}\label{fig:cora_mmrtm}
\subfigure[train time]{\includegraphics[height=1.3in, width=1.25in]{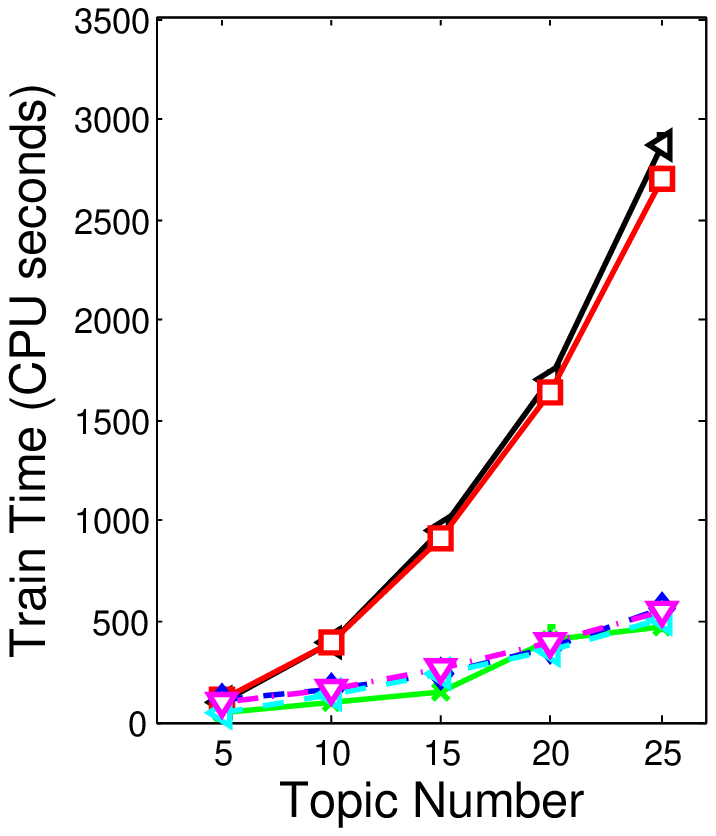}}\label{fig:cora_mmrtm}
\subfigure[test time]{\includegraphics[height=1.3in, width=1.25in]{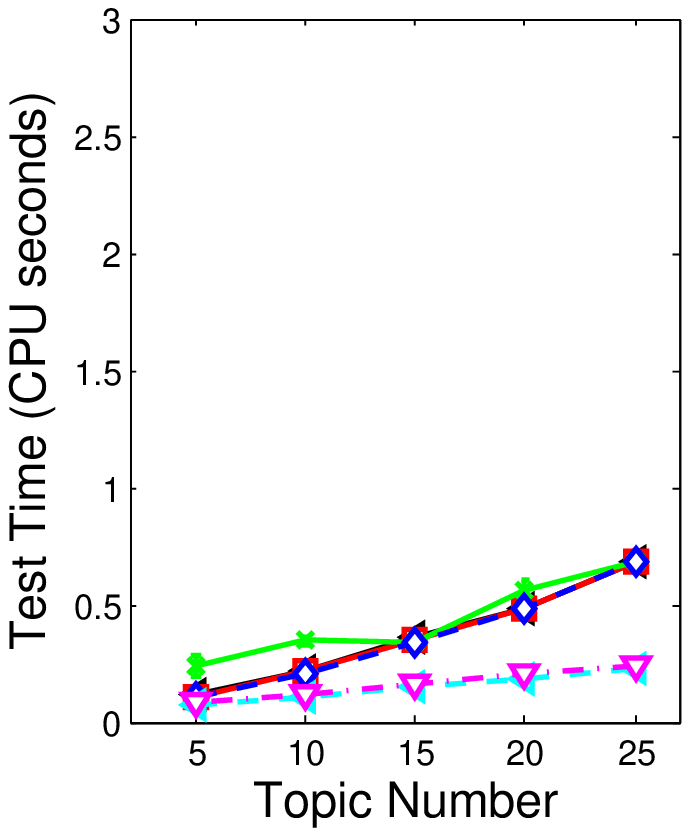}}\label{fig:cora_mmrtm}
\caption{Results of various models with different numbers of topics on the Cora dataset.}
\label{fig:cora_mmrtm}
\end{figure*}


\begin{figure*}
\centering
\subfigure[link rank]{\includegraphics[height=1.3in, width=1.25in]{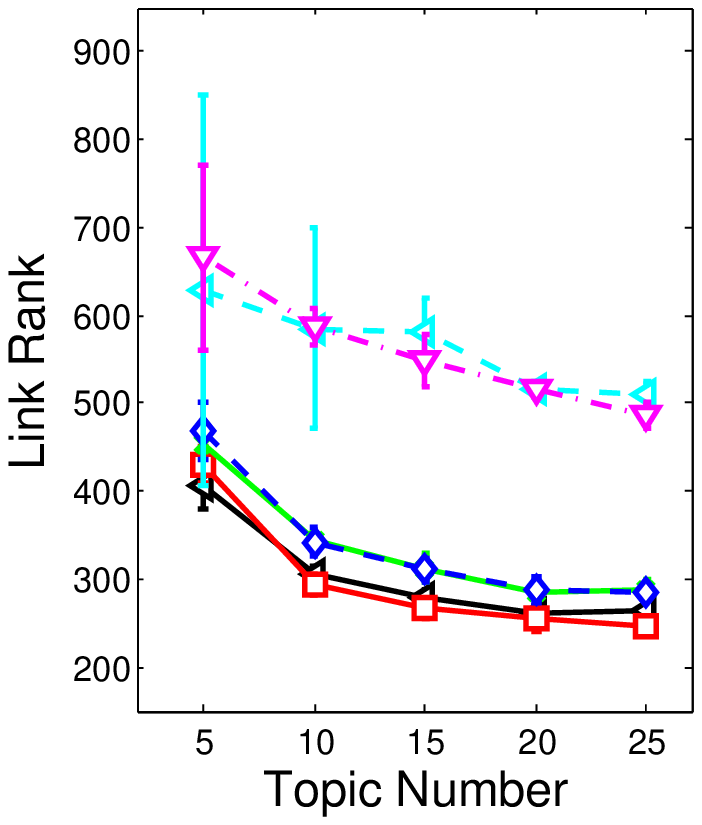}}\label{fig:cite_mmrtm}
\subfigure[word rank]{\includegraphics[height=1.3in, width=1.25in]{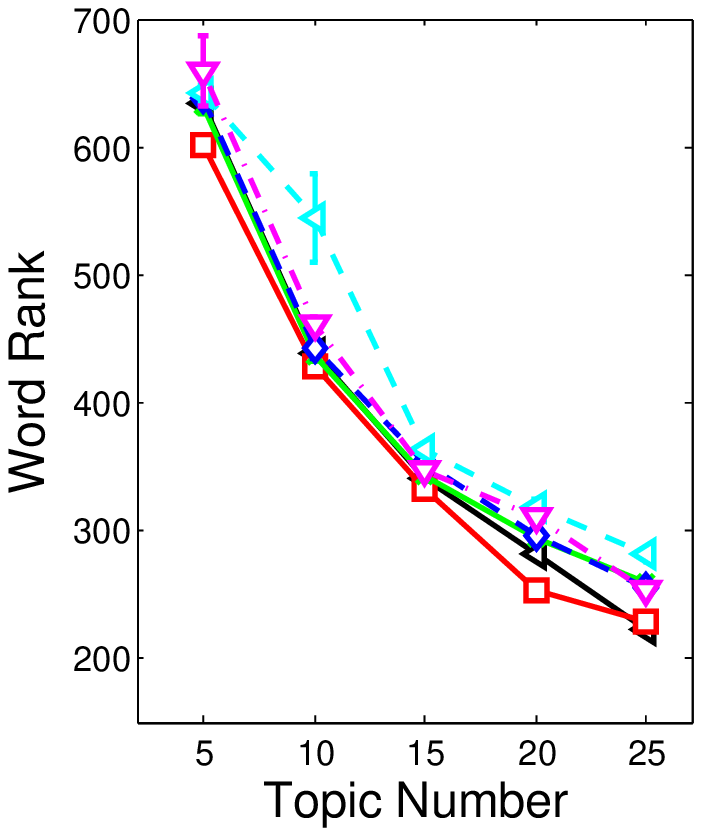}}\label{fig:cite_mmrtm}
\subfigure[AUC score]{\includegraphics[height=1.3in, width=1.25in]{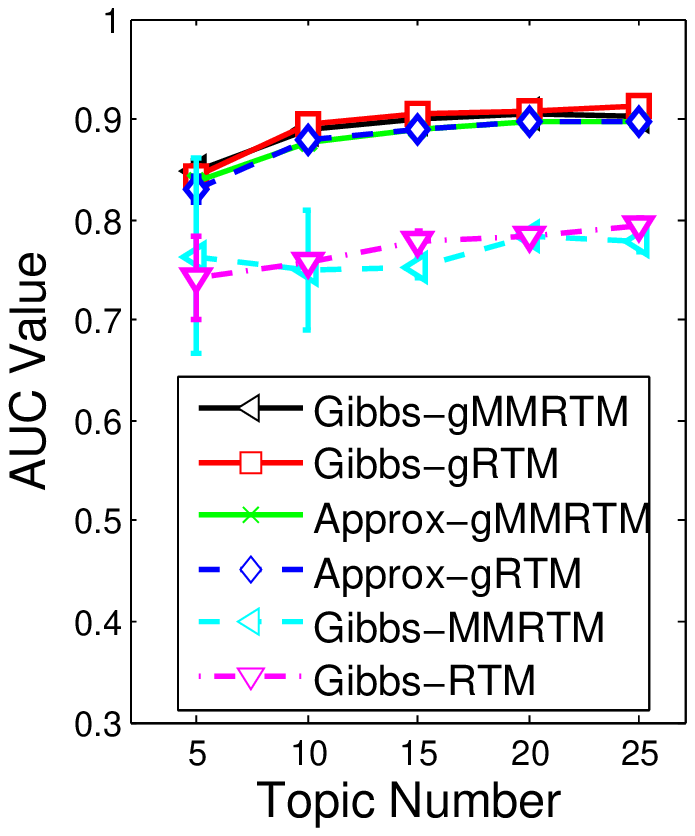}}\label{fig:cite_mmrtm}
\subfigure[train time]{\includegraphics[height=1.3in, width=1.25in]{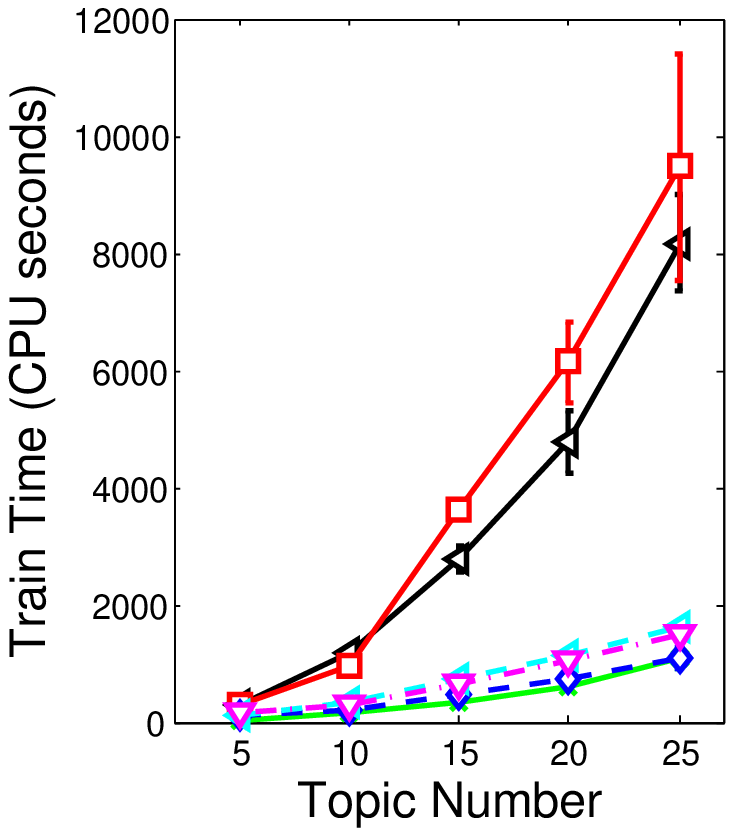}}\label{fig:cite_mmrtm}
\subfigure[test time]{\includegraphics[height=1.3in, width=1.25in]{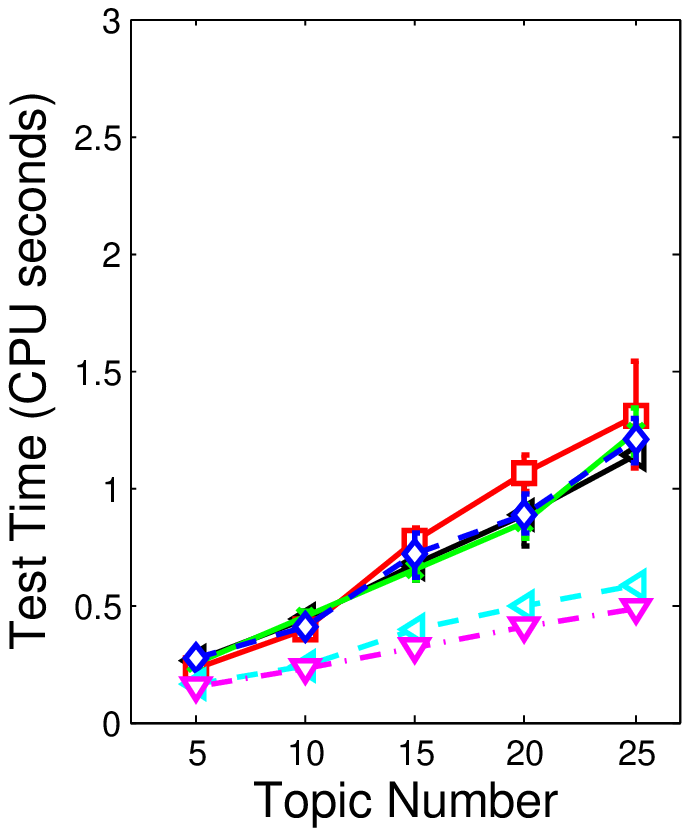}}\label{fig:cite_mmrtm}
\caption{Results of various models with different numbers of topics on the Citeseer dataset.}
\label{fig:citeseer_mmrtm}
\end{figure*}

The outstanding performance of Gibbs-gRTM is due to many possible factors. For example, the superior performance of Gibbs-gRTM over the diagonal Gibbs-RTM demonstrates that it is important to consider all pairwise topic interactions to fit real network data; and the superior performance of Gibbs-RTM over Var-RTM shows the benefits of using the regularization parameter $c$ in the regularized Bayesian framework and a collapsed Gibbs sampling algorithm without restricting mean-field assumptions\footnote{Gibbs-RTM doesn't outperform Var-RTM on Citeseer because they use different strategies of drawing negative samples. If we use the same strategy (e.g., randomly drawing 1\% negative samples), Gibbs-RTM significantly outperforms Var-RTM.}.

\begin{table}
\caption{Split of training time on Cora dataset.}\label{table:split}\vspace{-.1cm}
\begin{center}
\scalebox{1}{
\begin{tabular}{c|ccc}
\hline
{} & Sample $\Zv$  & Sample $\lambdav$ & Sample $U$  \\
\hline
K=10     & 331.2 (73.55\%) & 55.3 (12.29\%) & 67.8 (14.16\%) \\
K=15     & 746.8 (76.54\%) & 55.0 (5.64\%) & 173.9 (17.82\%) \\
K=20     & 1300.3 (74.16\%) & 55.4 (3.16\%) & 397.7 (22.68\%) \\
\hline
\end{tabular}}
\end{center}\vspace{-.2cm}
\end{table}

To single out the influence of the proposed Gibbs sampling algorithm, we also present the results of Var-RTM and Gibbs-RTM with $c=1$, both of which randomly sample $0.2\%$ unobserved links\footnote{Var-RTM performs much worse if using $1\%$ negative links, while Gibbs-RTM could obtain similar performance (see Fig.~\ref{fig:SampleRatio}) due to its effectiveness in dealing with imbalance.} as negative examples on the Cora dataset. We can see that by using Gibbs sampling without restricting mean-field assumptions, Gibbs-RTM (neg $0.2\%$) outperforms Var-RTM (neg $0.2\%$) that makes mean-field assumptions when the number of topics is larger than 10. We defer more careful analysis of other factors in the next section, including $c$ and the subsampling ratio.

We also note that the cost we pay for the outstanding performance of Gibbs-gRTM is on training time, which is much longer than that of Var-RTM because Gibbs-gRTM has $K^2$ latent features in the logistic likelihood and more training link pairs, while Var-RTM has $K$ latent features and only uses the sparse positive links as training examples. Fortunately, we can apply a simple approximate method in sampling $\Zv$ as in Approx-gRTM to significantly improve the training efficiency, while the prediction performance is not sacrificed much. In fact, Approx-gRTM is still significantly better than Var-RTM in all cases, and it has comparable link prediction performance with Gibbs-gRTM on the WebKB dataset, when $K$ is large. Table~\ref{table:split} further shows the training time spent on each sub-step of the Gibbs sampling algorithm of Gibbs-gRTM. We can see that the step of sampling $\Zv$ takes most of the time ($>70\%$); and the steps of sampling $\Zv$ and $\etav$ take more time as $K$ increases, while the step of sampling $\lambdav$ takes almost a constant time when $K$ changes.



\begin{figure}
\centering
\subfigure[Time for drawing $\lambda$]{\includegraphics[height=1.5in, width=1.6in]{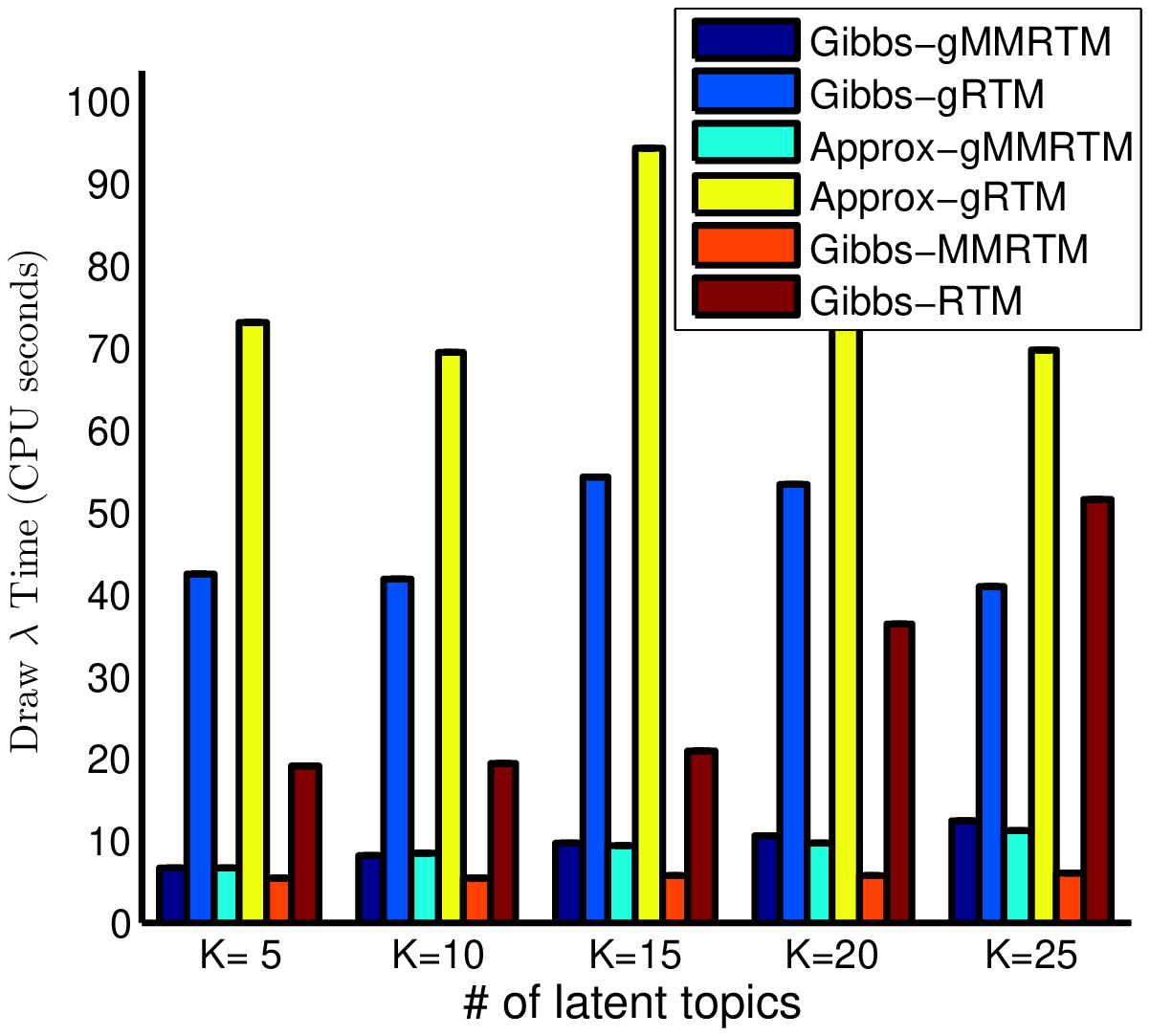}\label{fig:cite_lambda}}
\subfigure[Time for drawing $\eta$]{\includegraphics[height=1.5in, width=1.6in]{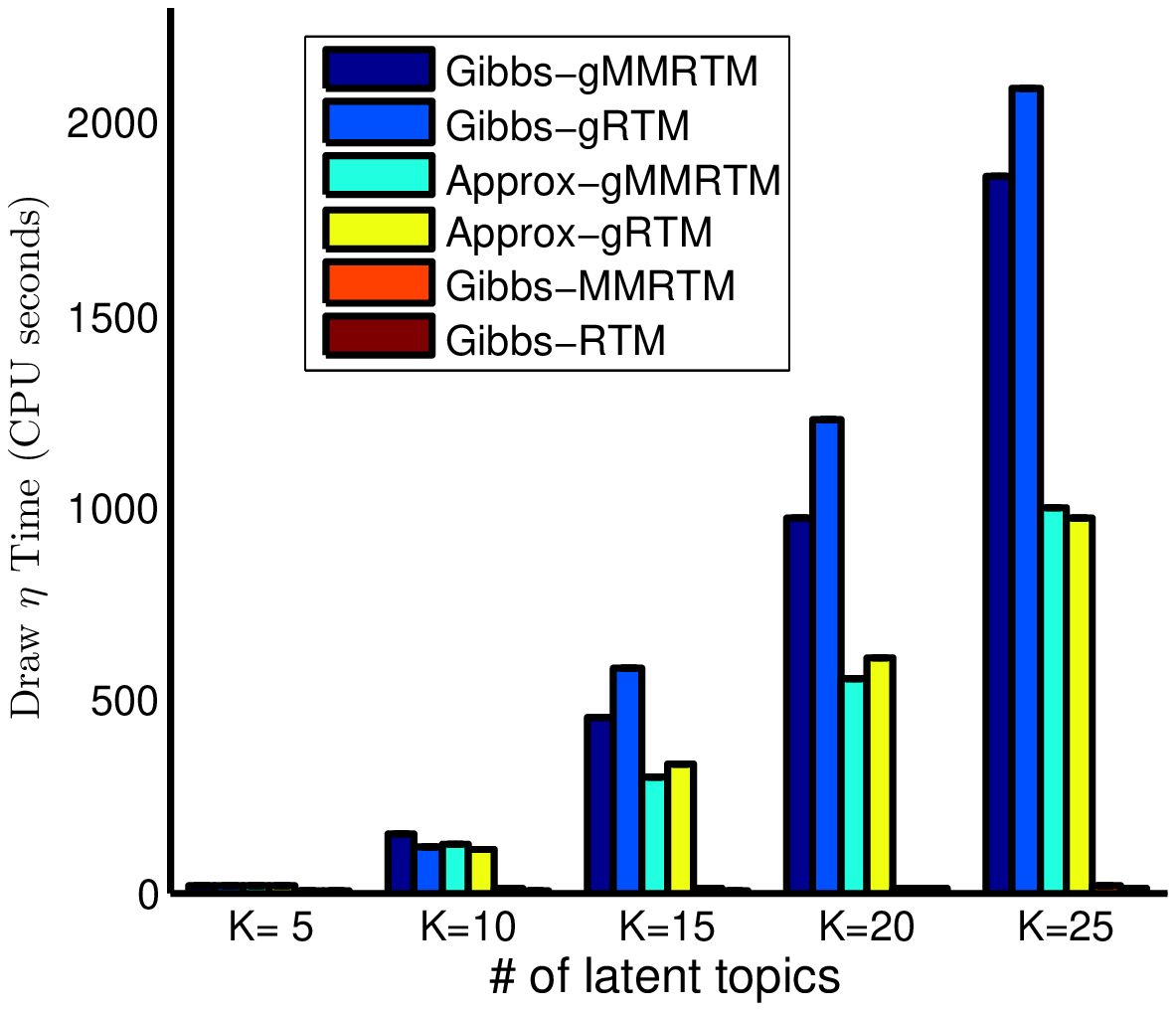}\label{fig:cite_eta}}
\caption{Time complexity of drawing $\lambdav$ and $\etav$ on the Citeseer dataset.}
\label{fig:timecomparison}
\end{figure}

\vspace{-0.2cm}
\subsubsection{Results with the Hinge Loss}\label{section:resultHingeLoss}

Fig.~\ref{fig:cora_mmrtm} and Fig.~\ref{fig:citeseer_mmrtm} show the 5-fold average results with standard deviations of the discriminative RTMs with hinge loss, comparing with the RTMs with log-loss on Cora and Citeseer datasets\footnote{The result on WebKB dataset is similar, but omitted for saving space. Please refer to Fig.~\ref{fig:webkb_mmrtm} in Appendix.}. 
We can see that the discriminative RTM models with hinge loss (i.e., Gibbs-gMMRTM and Gibbs-MMRTM) obtain comparable predictive results (e.g., link rank and AUC scores) with the RTMs using log-loss (i.e., Gibbs-gRTM and Gibbs-RTM). And owing to the use of a full weight matrix, Gibbs-gMMRTM obtains superior performance over the diagonal Gibbs-MMRTM. These results verify the fact that the max-margin RTMs can be used as a competing alternative approach for statistical network link prediction. For word rank, all the RTM models using Gibbs sampling perform similarly.

As shown in Fig.~\ref{fig:timecomparison}, one superiority of the max margin Gibbs-gMMRTM is that the time cost of drawing $\lambdav$ is cheaper than that in Gibbs-gRTM with log-loss. Specifically, the time of drawing $\lambdav$ in Gibbs-gRTM is about 10 times longer than Gibbs-gMMRTM (Fig.~\ref{fig:cite_lambda}). This is because sampling from a Polya-gamma distribution in Gibbs-gRTM needs a few steps of iteration for convergence, which takes more time than the constant time sampler of an inverse Gaussian distribution~\cite{Michael:IG76} in Gibbs-gMMRTM. We also observe that the time costs for drawing $\etav$ (Fig.~\ref{fig:cite_eta}) in Gibbs-gRTM and Gibbs-gMMRTM are comparable\footnote{Sampling $\Zv$ also takes comparable time. Omitted for saving space.}. As most of the time is spent on drawing $\Zv$ and $\etav$, the total training time of the RTMs with the two types of losses are similar (gMMRTM is slightly faster on Citeseer). Fortunately, we can also develop Approx-gMMRTM by using a simple approximate method in sampling $\Zv$ to greatly improve the time efficiency (Fig.~\ref{fig:cora_mmrtm} and Fig.~\ref{fig:citeseer_mmrtm}), and the prediction performance is still very compelling, especially on the Citeseer dataset.


\begin{figure}[t]
\centering
\subfigure[link rank]{\includegraphics[height=1.3in, width=1in]{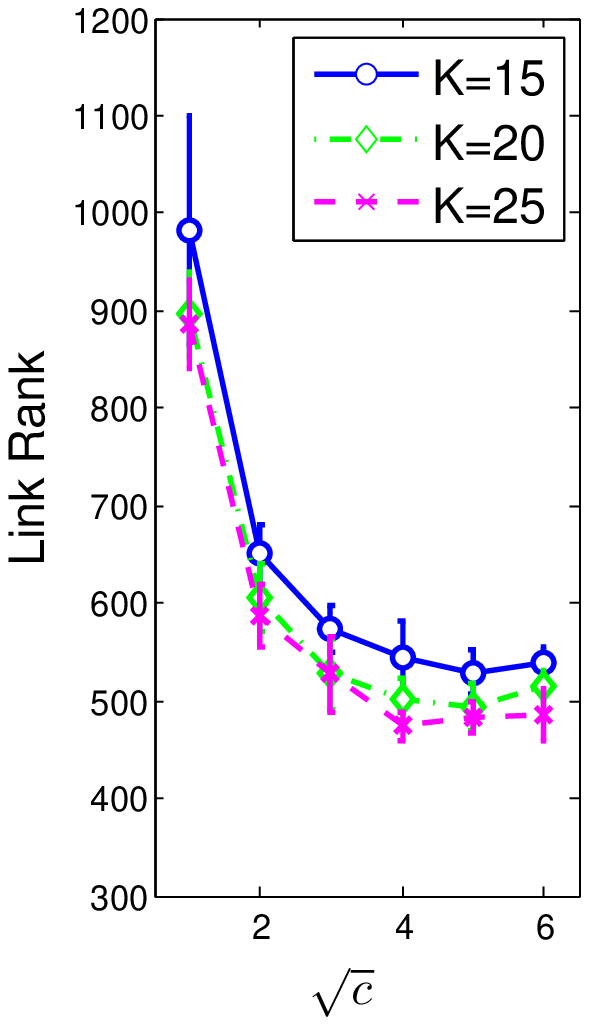}}\label{fig:Csen_link_rtm}
\subfigure[AUC score]{\includegraphics[height=1.3in, width=1in]{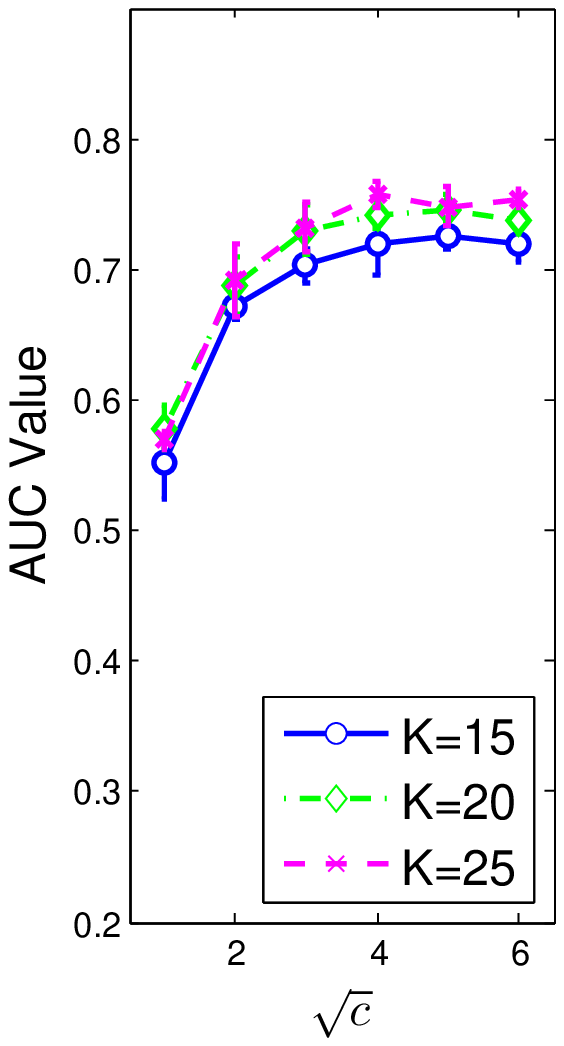}}\label{fig:Csen_AUC_rtm}
\subfigure[word rank]{\includegraphics[height=1.3in, width=1in]{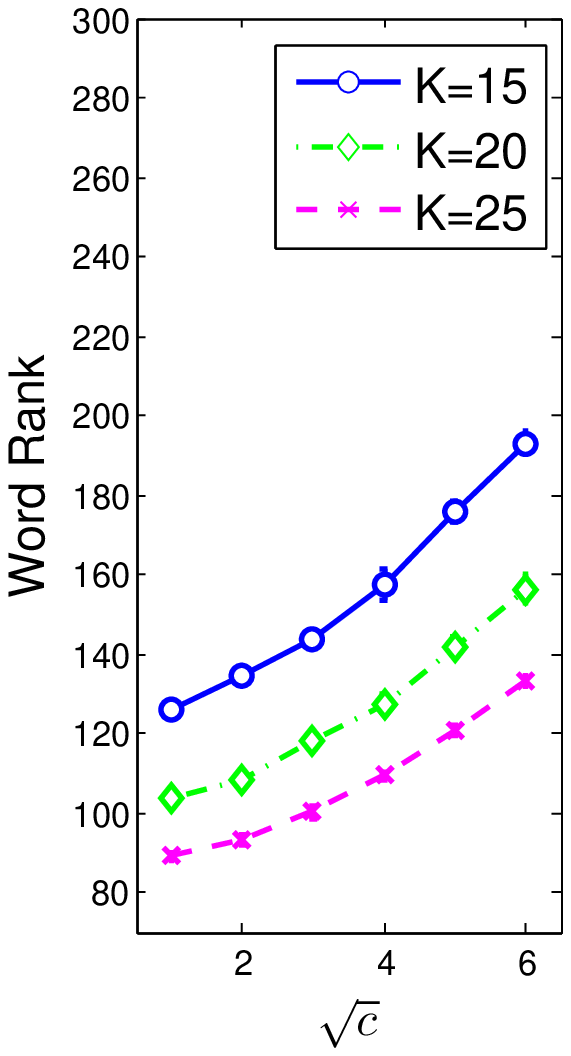}}\label{fig:Csen_word_rtm}
\caption{Performance of Gibbs-RTM with different $c$ values on the Cora dataset.}
\label{fig:Csen_rtm}
\end{figure}

\begin{figure}
\centering
\subfigure[link rank]{\includegraphics[height=1.3in, width=1in]{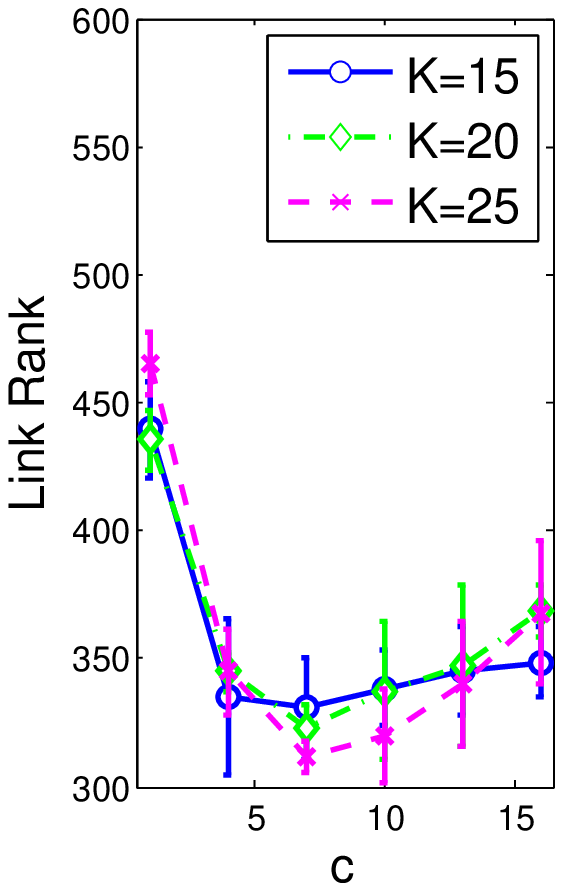}}\label{fig:Csen_link}
\subfigure[AUC score]{\includegraphics[height=1.3in, width=1in]{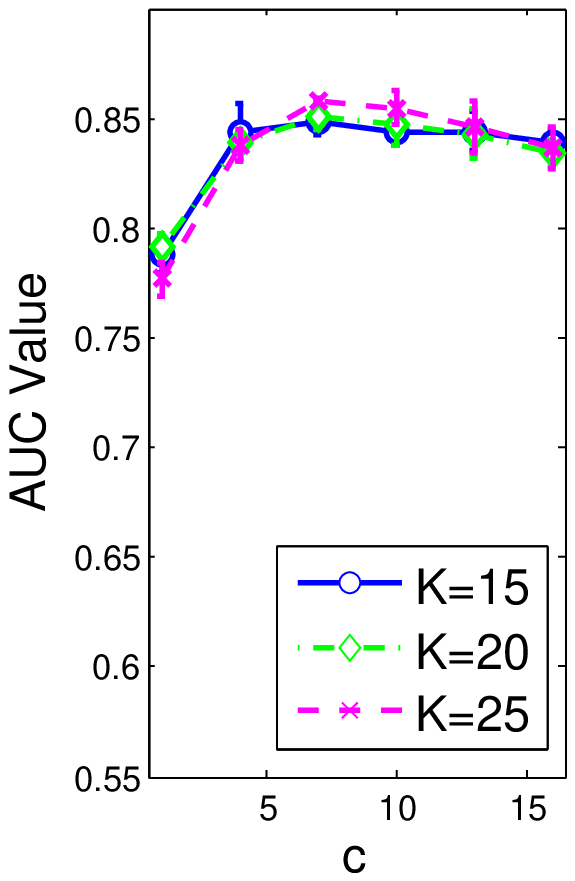}}\label{fig:Csen_word}
\subfigure[word rank]{\includegraphics[height=1.3in, width=1in]{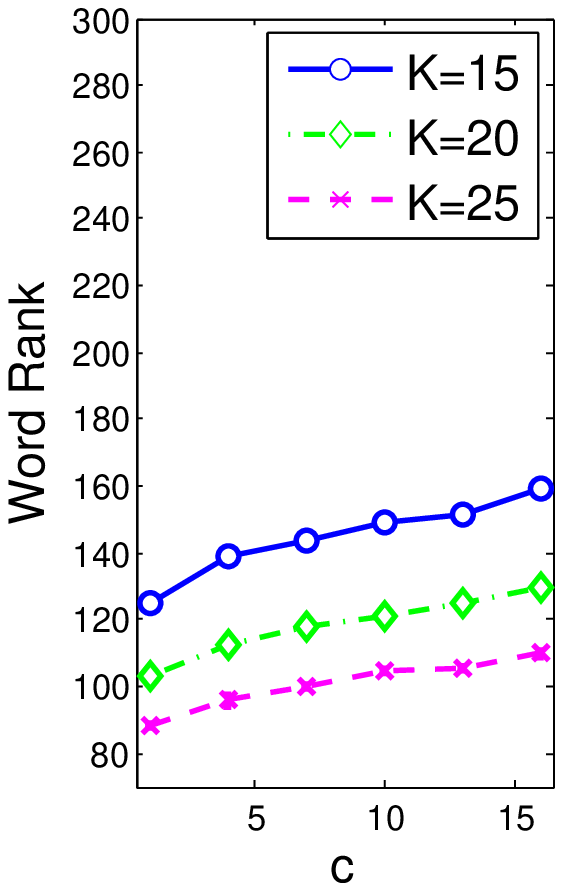}}\label{fig:Csen_AUC}
\caption{Performance of Gibbs-gRTM with different $c$ values on the Cora dataset.}
\label{fig:Csen}
\end{figure}

\begin{figure}[t]
\centering
\subfigure[link rank]{\includegraphics[height=1.3in, width=1in]{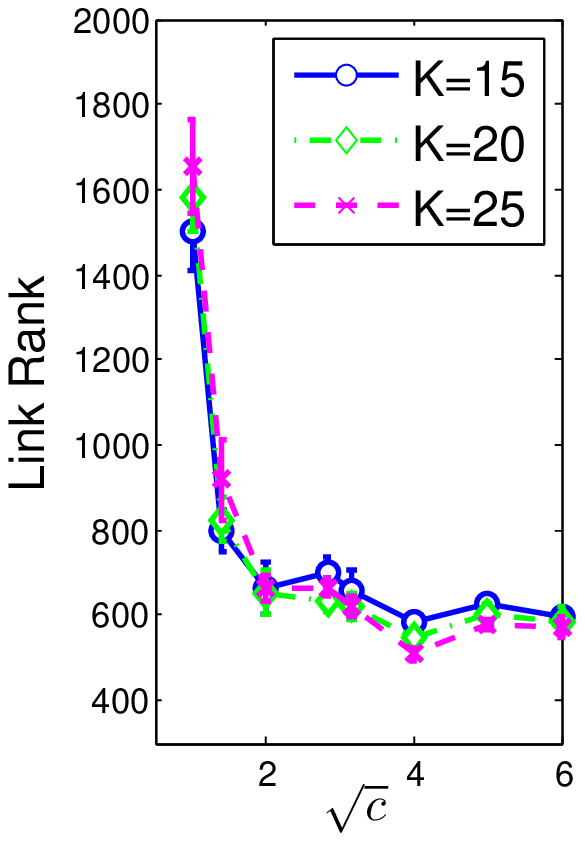}}\label{fig:Csen_link_mmrtm}
\subfigure[AUC score]{\includegraphics[height=1.3in, width=1in]{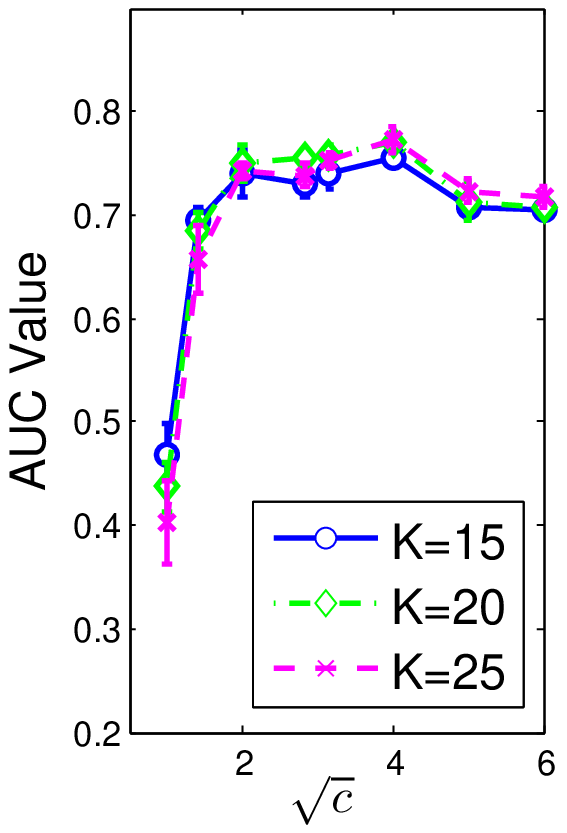}}\label{fig:Csen_AUC_mmrtm}
\subfigure[word rank]{\includegraphics[height=1.3in, width=1in]{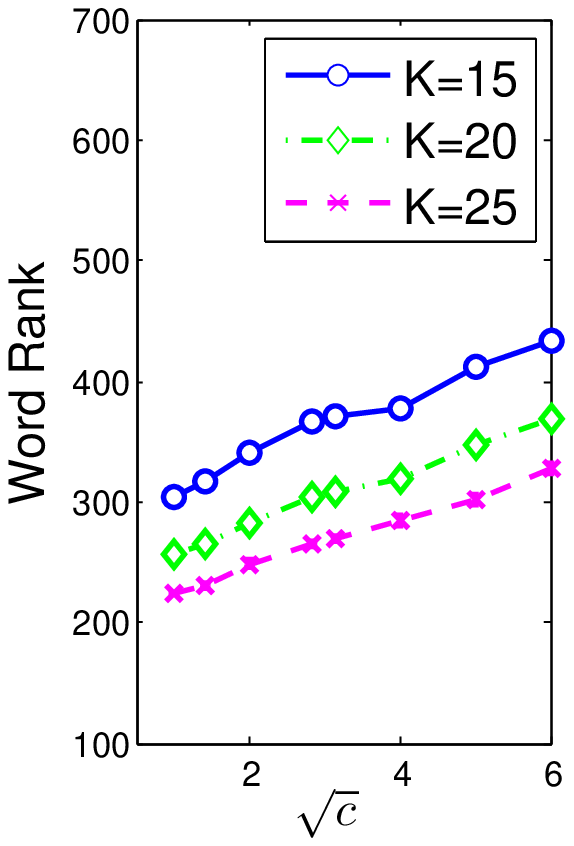}}\label{fig:Csen_word_mmrtm}
\caption{Performance of Gibbs-MMRTM with different $c$ values on the Citeseer dataset.}
\label{fig:Csen_mmrtm}
\end{figure}

\begin{figure}
\centering
\subfigure[link rank]{\includegraphics[height=1.3in, width=1in]{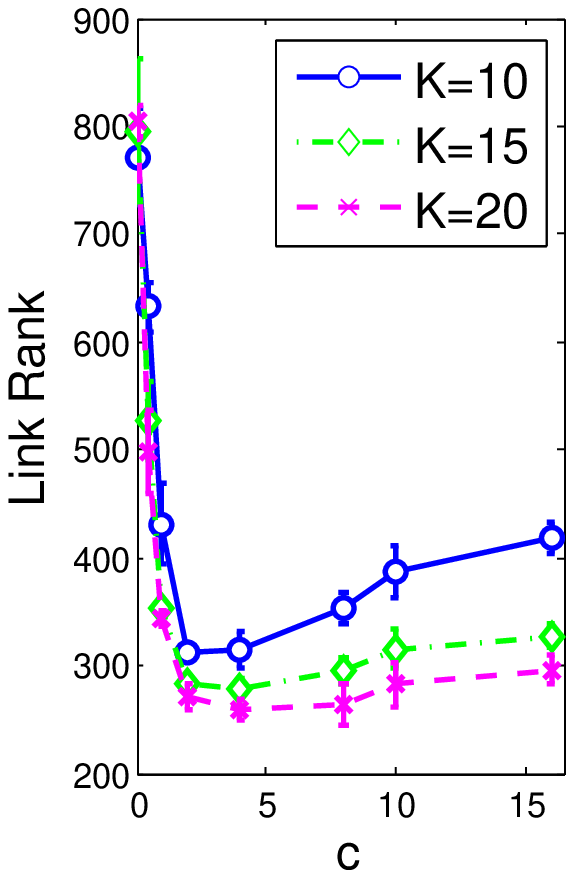}}\label{fig:Csen_link_mmgrtm}
\subfigure[AUC score]{\includegraphics[height=1.3in, width=1in]{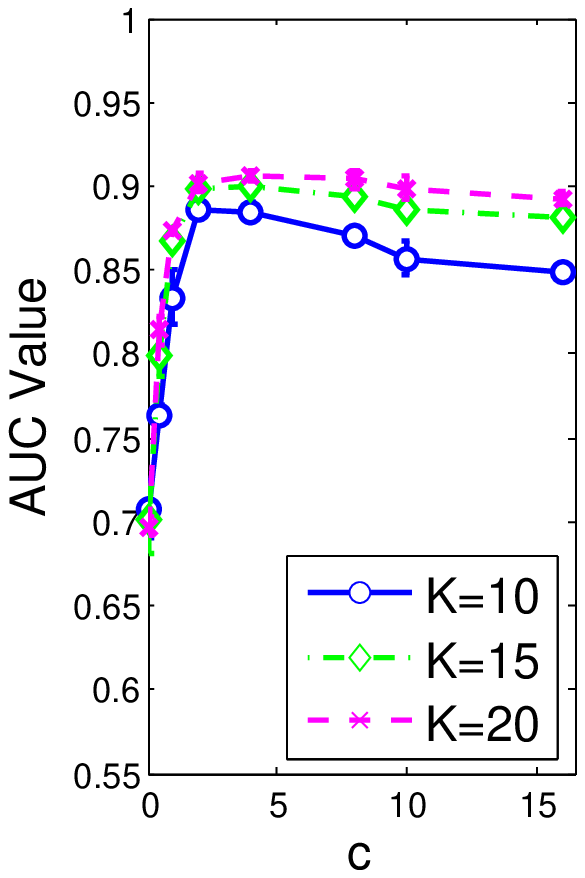}}\label{fig:Csen_AUC_mmgrtm}
\subfigure[word rank]{\includegraphics[height=1.3in, width=1in]{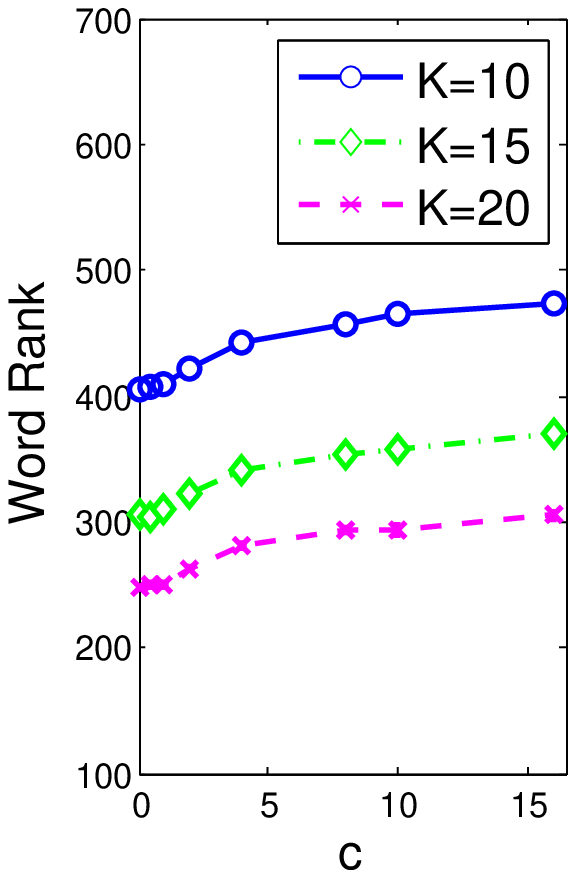}}\label{fig:Csen_word_mmgrtm}
\caption{Performance of Gibbs-gMMRTM with different $c$ values on the Citeseer dataset.}
\label{fig:Csen_mmgrtm}
\end{figure}

\vspace{-0.1cm}
\subsection{Sensitivity Analysis}

To provide more insights about the behaviors of our discriminative RTMs, we present a careful analysis of various factors.

\begin{figure}[t]
\centering
\subfigure[link rank]{\includegraphics[height=1.3in, width=1in]{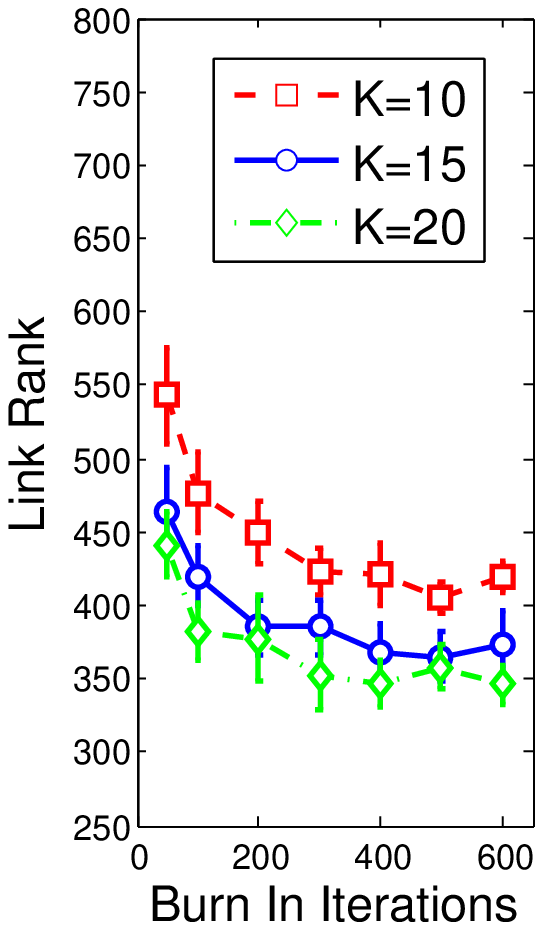}}\label{fig:burnin_link}
\subfigure[AUC score]{\includegraphics[height=1.3in, width=1in]{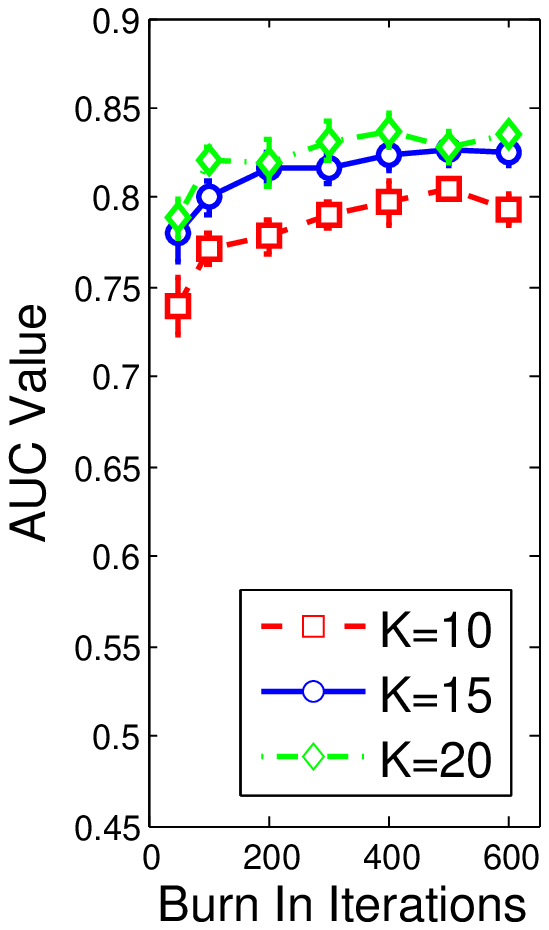}}\label{fig:burnin_AUC}
\subfigure[train time]{\includegraphics[height=1.3in, width=1in]{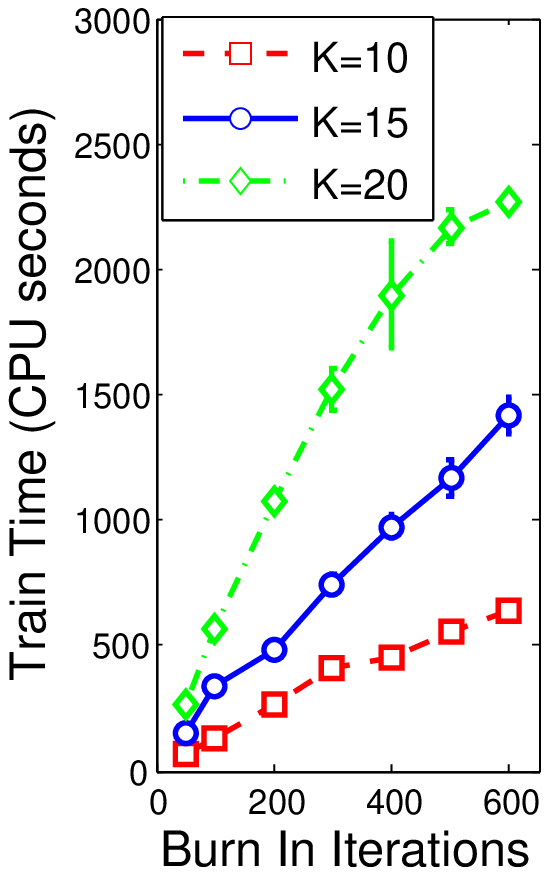}}\label{fig:burnin_trtime}
\caption{Performance of Gibbs-gRTM with different burn-in steps on the Cora dataset.}
\label{fig:burnin_rank}
\end{figure}

\begin{figure}
\centering
\subfigure[link rank]{\includegraphics[height=1.3in, width=1in]{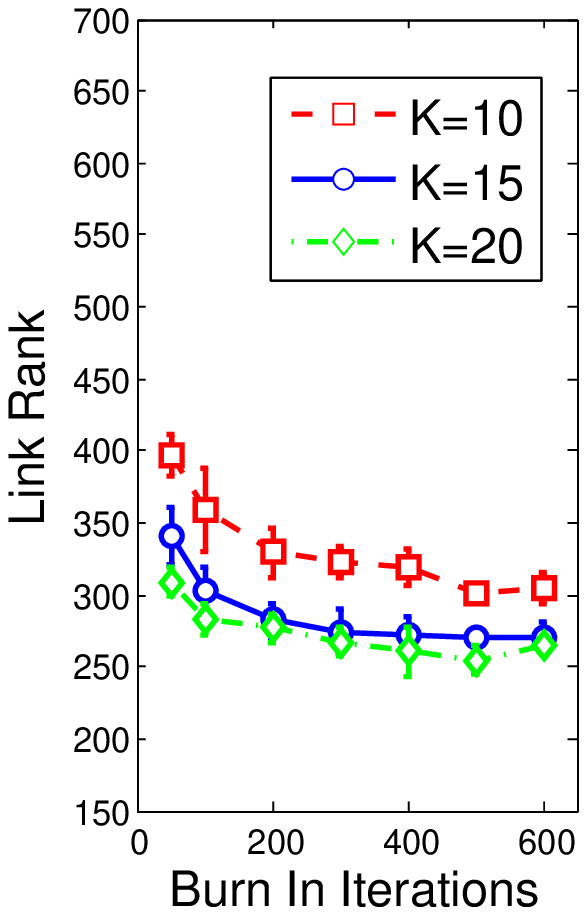}\label{fig:Burnin_mmgrtm_link}}
\subfigure[AUC]{\includegraphics[height=1.3in, width=1in]{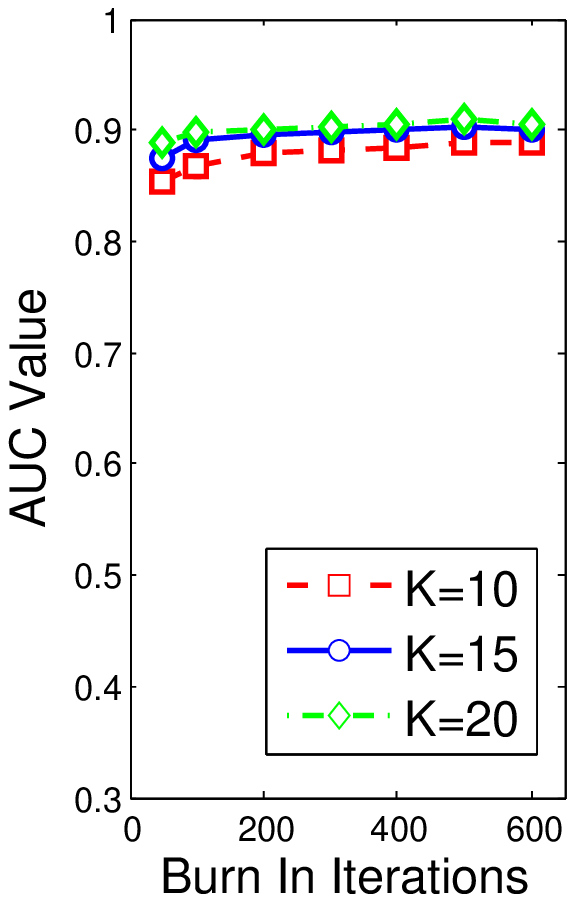}\label{fig:Burnin_mmgrtm_AUC}}
\subfigure[train time]{\includegraphics[height=1.3in, width=1in]{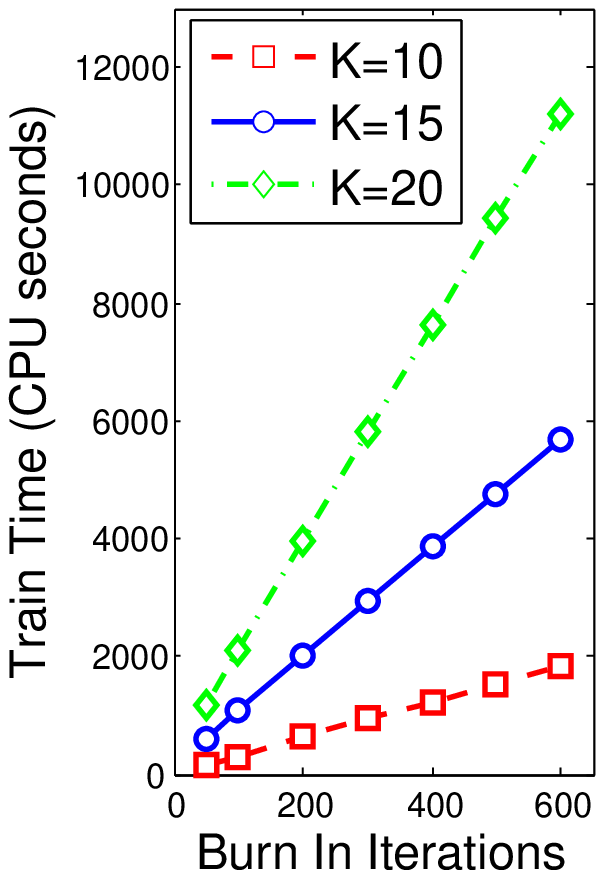}\label{fig:Burnin_mmgrtm_trtime}}
\caption{Performance of Gibbs-gMMRTM with different Burin-In steps on the Citeseer dataset.}
\label{fig:burnin_mmgrtm}
\end{figure}

\begin{figure}[t]
\centering
\subfigure[link rank]{\includegraphics[height=1.3in, width=1in]{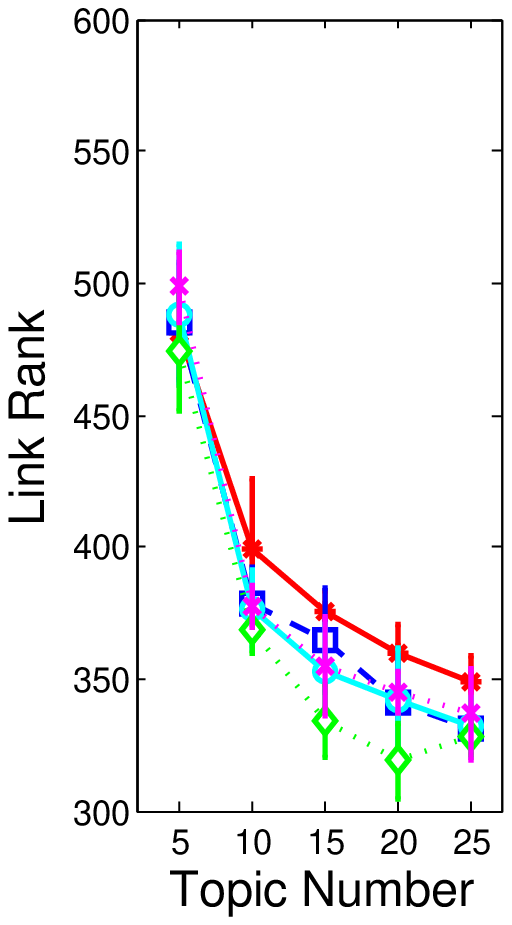}}\label{fig:SampleRatio-link}
\subfigure[AUC score]{\includegraphics[height=1.3in, width=1in]{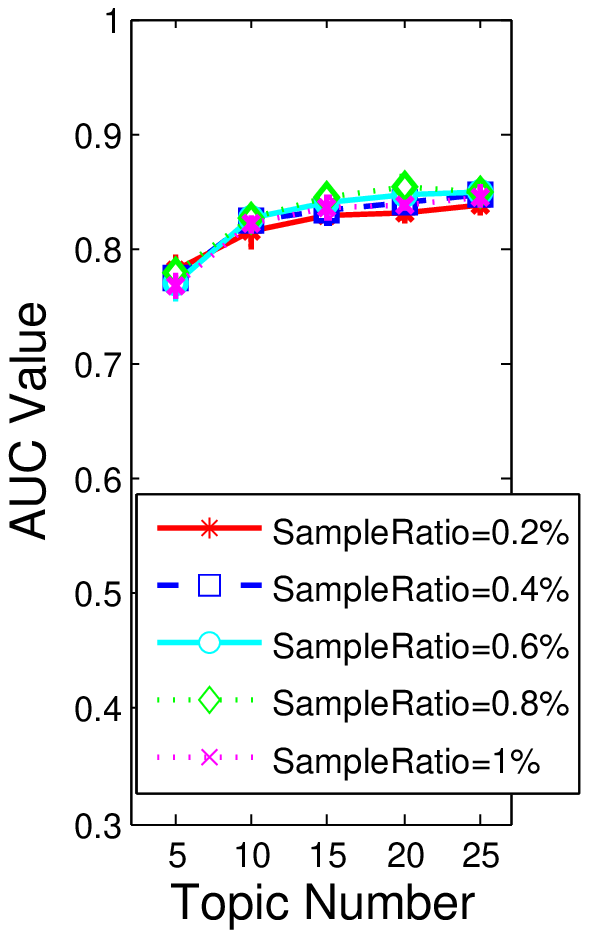}}\label{fig:SampleRatio-AUC}
\subfigure[train time]{\includegraphics[height=1.3in, width=1in]{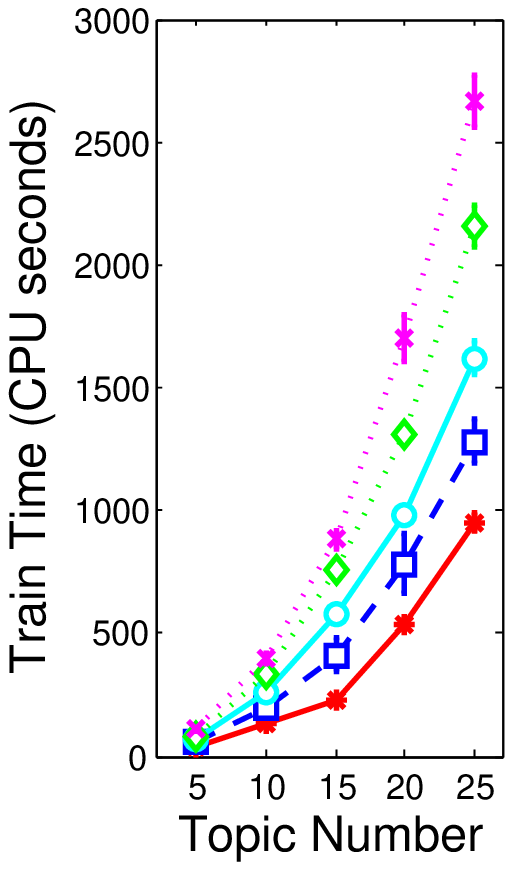}}\label{fig:SampleRatio-time}
\caption{Performance of Gibbs-gRTM with different numbers of negative training links on the Cora dataset.}
\label{fig:SampleRatio}
\end{figure}

\begin{figure}
\centering
\subfigure[link rank]{\includegraphics[height=1.3in, width=1in]{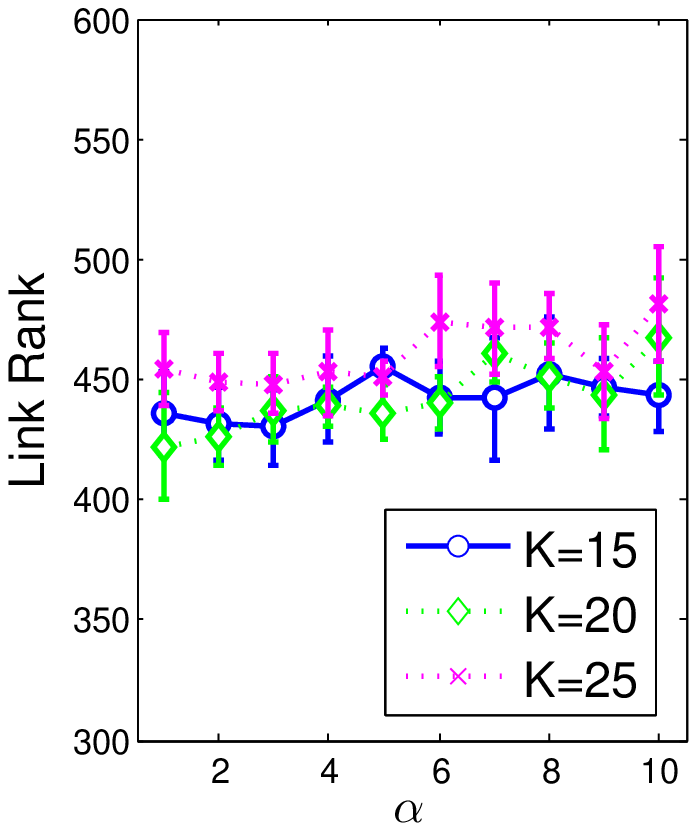}}\label{fig:Asen_link}
\subfigure[AUC score]{\includegraphics[height=1.3in, width=1in]{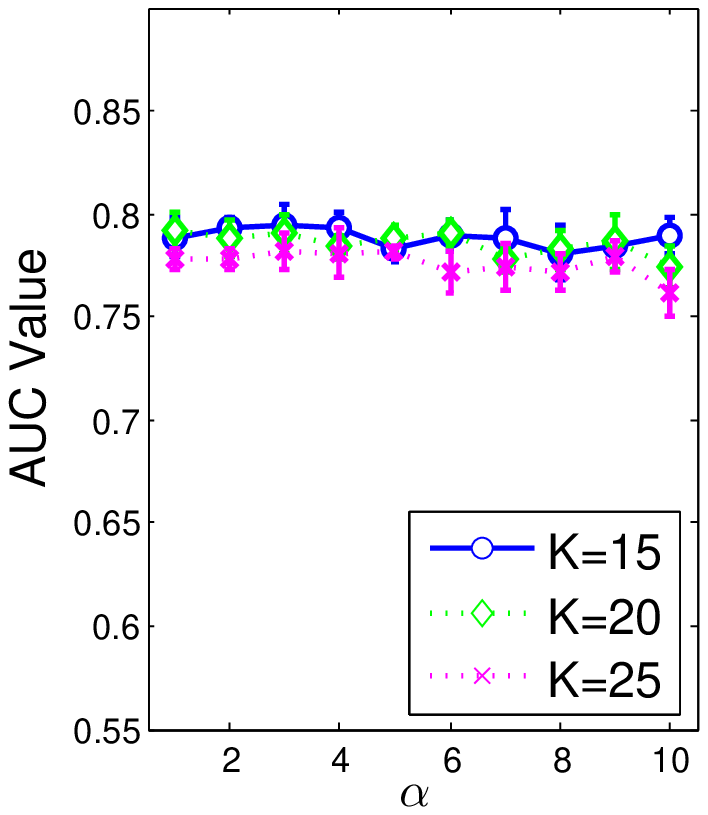}}\label{fig:Asen_AUC}
\subfigure[word rank]{\includegraphics[height=1.3in, width=1in]{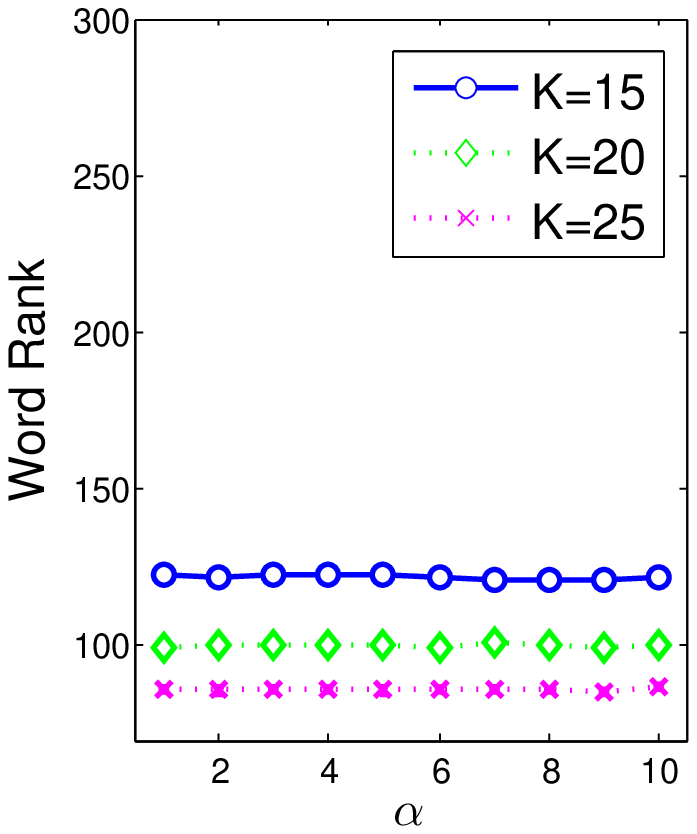}}\label{fig:Asen_word}
\caption{Performance of Gibbs-gRTM ($c=1$) with different $\alpha$ values on the Cora dataset.} 
\label{fig:Asen}
\end{figure}

\vspace{-0.1cm}
\subsubsection{Hyper-parameter $c$}\label{section:sensitivityC}

Fig.~\ref{fig:Csen_rtm} and~\ref{fig:Csen_mmrtm} show the prediction performance of the diagonal Gibbs-RTM and Gibbs-MMRTM with different $c$ values on both Cora and Citeseer datasets\footnote{We have similar observations on the WebKB dataset, again omitted for saving space.}, and Fig.~\ref{fig:Csen} and ~\ref{fig:Csen_mmgrtm} show the results of the generalized Gibbs-gRTM and Gibbs-gMMRTM. For Gibbs-RTM and Gibbs-MMRTM, we can see that the link rank decreases and AUC scores increase when $c$ becomes larger and the prediction performance is stable in a wide range (e.g., $2\leq\sqrt{c}\leq6$). But the RTM model (i.e., $c=1$) using Gibbs sampling doesn't perform well due to its ineffectiveness in dealing with imbalanced network data. In Fig.~\ref{fig:Csen} and \ref{fig:Csen_mmgrtm}, we can observe that when $2\leq c \leq 10$, the link rank and AUC scores of Gibbs-gRTM achieve the local optimum, which performs much better than the performance of Gibbs-gRTM when $c=1$. In general,
we can see that both Gibbs-gRTM and Gibbs-gMMRTM need a smaller $c$ to get the best performance. This is because by allowing all pairwise topic interactions, Gibbs-gRTM and Gibbs-gMMRTM are much more expressive than Gibbs-RTM and Gibbs-MMRTM with a diagonal weight matrix; and thus easier to over-fit when $c$ gets large.

For all the proposed models, the word rank increases slowly with the growth of $c$. This is because a larger $c$ value makes the model more concentrated on fitting link structures and thus the fitness of observed words sacrifices a bit. But if we compare with the variational RTM (i.e., Var-RTM) as shown in Fig.~\ref{fig:cora} and Fig.~\ref{fig:citeseer}, the word ranks of all the four proposed RTMs using Gibbs sampling are much lower for all the $c$ values we have tested. This suggests the advantages of the collapsed Gibbs sampling algorithms. 

%

\vspace{-0.1cm}
\subsubsection{Burin-In Steps}
Fig.~\ref{fig:burnin_rank} and Fig.~\ref{fig:burnin_mmgrtm} show the sensitivity of Gibbs-gRTM and Gibbs-gMMRTM with respect to the number of burn-in iterations, respectively. 
We can see that the link rank and AUC scores converge fast to stable optimum points with about 300 iterations. The training time grows almost linearly with respect to the number of burn-in iterations. We have similar observations for the diagonal Gibbs-RTM, Gibbs-MMRTM and Approximate RTMs with fast approximation. In the previous experiments, we have set the burn-in steps at 400 for Cora and Citeseer, which is sufficiently large.


\vspace{-0.1cm}
\subsubsection{Subsample ratio}
Fig.~\ref{fig:SampleRatio} shows the influence of the subsample ratio on the performance of Gibbs-gRTM on the Cora data. In total, less than $0.1\%$ links are positive on the Cora networks. We can see that by introducing the regularization parameter $c$, Gibbs-gRTM can effectively fit various imbalanced network data and the different subsample ratios have a weak influence on the performance of Gibbs-gRTM. Since a larger subsample ratio leads to a bigger training set, the training time increases as expected. We have similar observations on Gibbs-gMMRTM and other models.

\vspace{-0.1cm}
\subsubsection{Dirichlet prior $\alpha$}
Fig.~\ref{fig:Asen} shows the sensitivity of the generalized Gibbs-gRTM model and diagonal Gibbs-RTM on the Cora dataset with different $\alpha$ values. We can see that the results are quite stable in a wide range of $\alpha$ (i.e., $1\leq\alpha\leq10$) for three different topic numbers. We have similar observations for Gibbs-gMMRTM. In the previous experiments, we set $\alpha$ = 5 for both Gibbs-gRTM and Gibbs-gMMRTM.

\begin{table*}[t]
\caption{Top 8 link predictions made by Gibbs-gRTM and Var-RTM on the Cora dataset. (Papers with titles in bold have ground-truth links with the query document.)}\label{table:LinkSuggestion}
\begin{center}
\scalebox{.95}{\setlength{\tabcolsep}{1.8pt}
       \begin{tabular}{|c|c|}
       \hline
        \hline
        {{\color{blue}\bf Query: Competitive environments evolve better solutions for complex tasks }}& {}       \\
        \hline
        {\bf{Coevolving High Level Representations}}&{\multirow{8}{*}{\rotatebox{270}{Gibbs-gRTM}}}\\
        {\bf{Strongly typed genetic programming in evolving cooperation strategies}}&{}\\
        {\bf{Genetic Algorithms in Search, Optimization and Machine Learning}}&{}\\
        {Improving tactical plans with genetic algorithms}&{}\\
        {\bf{Some studies in machine learning using the game of Checkers}}&{}\\
        {Issues in evolutionary robotics: From Animals to Animats}&{}\\
        {Strongly Typed Genetic Programming}&{}\\
        {Evaluating and improving steady state evolutionary algorithms on constraint satisfaction problems}&{}\\
        \hline
        {\bf{Coevolving High Level Representations}}&{\multirow{8}{*}{\rotatebox{270}{Var-RTM}}}\\
        {A survey of Evolutionary Strategies}&{}\\
        {\bf{Genetic Algorithms in Search, Optimization and Machine Learning}}&{}\\
        {\bf{Strongly typed genetic programming in evolving cooperation strategies}}&{}\\
        {Solving combinatorial problems using evolutionary algorithms}&{}\\
        {A promising genetic algorithm approach to job-shop scheduling, rescheduling, and open-shop scheduling problems}&{}\\
        {Evolutionary Module Acquisition}&{}\\
        {An Empirical Investigation of Multi-Parent Recombination Operators in Evolution Strategies}&{}\\
        \hline
        \hline
        \hline
        {\color{blue}\bf{Query: Planning by Incremental Dynamic Programming}}&{}\\
        \hline
        {\bf{Learning to predict by the methods of temporal differences}}&{\multirow{8}{*}{\rotatebox{270}{Gibbs-gRTM}}}\\
        {Neuronlike adaptive elements that can solve difficult learning control problems}&{}\\
        {Learning to Act using Real- Time Dynamic Programming}&{}\\
        {A new learning algorithm for blind signal separation}&{}\\
        {Planning with closed-loop macro actions}&{}\\
        {\bf{Some studies in machine learning using the game of Checkers}}&{}\\
        {Transfer of Learning by Composing Solutions of Elemental Sequential Tasks}&{}\\
        {Introduction to the Theory of Neural Computation}&{}\\
        \hline
        {Causation, action, and counterfactuals}&{\multirow{8}{*}{\rotatebox{270}{Var-RTM}}}\\
        {Learning Policies for Partially Observable Environments}&{}\\
        {Asynchronous modified policy iteration with single-sided updates}&{}\\
        {Hidden Markov models in computational biology: Applications to protein modeling}&{}\\
        {Exploiting structure in policy construction}&{}\\
        {Planning and acting in partially observable stochastic domains}&{}\\
        {A qualitative Markov assumption and its implications for belief change}&{}\\
        {Dynamic Programming and Markov Processes}&{}\\
        \hline
        \end{tabular}}
\end{center}
\end{table*}

\vspace{-0.1cm}
\subsection{Link Suggestion}

As in \cite{Chang:RTM09}, Gibbs-gRTM could perform the task of suggesting links for a new document (i.e., test data) based on its text contents. Table~\ref{table:LinkSuggestion} shows the example suggested citations for two query documents: 1) ``Competitive environments evolve better solutions for complex tasks" and 2) ``Planning by Incremental Dynamic Programming" in Cora data using Gibbs-gRTM and Var-RTM. The query documents are not observed during training, and suggestion results are ranked by the values of link prediction likelihood between the training documents and the given query. We can see that Gibbs-gRTM outperforms Var-RTM in terms of identifying more ground-truth links. For query 1, Gibbs-gRTM finds 4 truly linked documents (5 in total) in the top-8 suggested results, while Var-RTM finds 3. For query 2, Gibbs-gRTM finds 2 while Var-RTM does not find any. In general, Gibbs-gRTM outperforms Var-RTM on the link suggestion task across the whole corpus. We also observe that the suggested documents which are not truly linked to the query document are also very related to it semantically.

\section{Conclusions and Discussions}

We have presented discriminative relational topic models (gRTMs and gMMRTMs) which consider all pairwise topic interactions and are suitable for asymmetric networks. We perform regularized Bayesian inference that introduces a regularization parameter to control the imbalance issue in common real networks and gives a freedom to incorporate two popular loss functions (i.e., logistic log-loss and hinge loss). We also presented a simple ``augment-and-collapse" sampling algorithm for the proposed discriminative RTMs without restricting assumptions on the posterior distribution. Experiments on real network data demonstrate significant improvements on prediction tasks. The time efficiency can be significantly improved with a simple approximation method.

For future work, we are interested in making the sampling algorithm scalable to large networks by using distributed architectures~\cite{Smola:vldb10} or doing online inference~\cite{Hoffman:olda10}. Moreover, developing nonparametric RTMs to avoid model selection problems (i.e., automatically resolve the number of latent topics in RTMs) is an interesting direction. Finally, our current focus in on static networks, and it is interesting to extend the models to deal with dynamic networks, where incorporating time varying dependencies is a challenging problem to address.

\ifCLASSOPTIONcompsoc
  \section*{Acknowledgments}
\else
  \section*{Acknowledgment}
\fi

This work is supported by National Key Project for Basic Research of China (Grant Nos. 2013CB329403, 2012CB316301), and Tsinghua Self-innovation Project (Grant Nos: 20121088071, 20111081111), and China Postdoctoral Science Foundation Grant (Grant Nos: 2013T60117, 2012M520281).

\ifCLASSOPTIONcaptionsoff
  \newpage
\fi

\bibliographystyle{plain}
\bibliography{dpmed}\vspace{-0.5cm}

\begin{thebibliography}{10}

\bibitem{Gruber:2008}
M.~Rosen-zvi A.~Gruber and Y.~Weiss.
\newblock {Latent Topic Models for Hypertext}.
\newblock In {\em Proceedings of Uncertainty in Artificial Intelligence}, 2008.

\bibitem{Airoldi:nips08}
E.~Airoldi, D.~M. Blei, S.~E. Fienberg, and E.~P. Xing.
\newblock {Mixed Membership Stochastic Blockmodels}.
\newblock In {\em Advances in Neural Information Processing Systems}, 2008.

\bibitem{Akbani:ecml04}
R.~Akbani, S.~Kwek, and N.~Japkowicz.
\newblock {Applying Support Vector Machines to Imbalanced Datasets}.
\newblock In {\em European Conference on Machine Learning}, 2004.

\bibitem{backstrom}
L.~Backstrom and J.~Leskovec.
\newblock {Supervised Random Walks: Predicting and Recommending Links in Social
  Networks}.
\newblock In {\em International Conference on Web Search and Data Mining},
  2011.

\bibitem{SDM:2011}
R.~Balasubramanyan and W.~Cohen.
\newblock {Block-LDA: Jointly Modeling Entity-Annotated Text and Entity-entity
  Links}.
\newblock In {\em Proceeding of the SIAM International Conference on Data
  Mining}, 2011.

\bibitem{Bengio:2012}
Y.~Bengio, A.~Courville, and P.~Vincent.
\newblock {Representation Learning: A Review and New Perspectives}.
\newblock {\em arXiv:1206.5538v2}, 2012.

\bibitem{Blei:03}
D.~Blei, A.~Ng, and M.~I. Jordan.
\newblock {Latent Dirichlet Allocation}.
\newblock {\em Journal of Machine Learning Research}, (3):993--1022, 2003.

\bibitem{Chang:RTM09}
J.~Chang and D.~Blei.
\newblock {Relational Topic Models for Document Networks}.
\newblock In {\em International Conference on Artificial Intelligence and
  Statistics}, 2009.

\bibitem{Chen:ijcai13}
N.~Chen, J.~Zhu, F.~Xia, and B.~Zhang.
\newblock {Generalized Relational Topic Models with Data Augmentation}.
\newblock In {\em International Joint Conference on Artificial Intelligence},
  2013.

\bibitem{Craven:1998}
M.~Craven, D.~Dipasquo, D.~Freitag, and A.~McCallum.
\newblock {Learning to Extract Symbolic Knowledge from the World Wide Web}.
\newblock In {\em AAAI Conference on Artificial Intelligence}, 1998.

\bibitem{Dempster1977}
A.~P. Dempster, N.~M. Laird, and D.~B. Rubin.
\newblock {Maximum Likelihood Estimation from Incomplete Data via the EM
  Algorithm}.
\newblock {\em Journal of the Royal Statistical Society, Ser. B}, (39):1--38,
  1977.

\bibitem{Devroye:book1986}
L.~Devroye.
\newblock {\em {Non-uniform random variate generation}}.
\newblock Springer-Verlag, 1986.

\bibitem{Dietz:2007}
L.~Dietz, S.~Bickel, and T.~Scheffer.
\newblock {Unsupervised Prediction of Citation Influences}.
\newblock In {\em Proceedings of the 24th Annual International Conference on
  Machine Learning}, 2007.

\bibitem{DykMeng2001}
D.~Van Dyk and X.~Meng.
\newblock {The Art of Data Augmentation}.
\newblock {\em Journal of Computational and Graphical Statistics}, 10(1):1--50,
  2001.

\bibitem{ParallelCholeskyDeCompo:1986}
A.~George, M.~Heath, and J.~Liu.
\newblock {Parallel Cholesky Factorization on a Shared-memory Multiprocessor}.
\newblock {\em Linear Algebra and Its Applications}, 77:165--187, 1986.

\bibitem{Germain:icml09}
P.~Germain, A.~Lacasse, F.~Laviolette, and M.~Marchand.
\newblock {PAC-Bayesian Learning of Linear Classifiers}.
\newblock In {\em International Conference on Machine Learning}, pages
  353--360, 2009.

\bibitem{Goldenberg:2010}
A.~Goldenberg, A.~X. Zheng, S.~E. Fienberg, and E.~M. Airoldi.
\newblock {A Survey of Statistical Network Models}.
\newblock {\em Foundations and Trends in Machine Learning}, 2(2):129--233,
  2010.

\bibitem{Griffiths:04}
T.~L. Griffiths and M.~Steyvers.
\newblock {Finding Scientific Topics}.
\newblock {\em Proceedings of the National Academy of Sciences}, 2004.

\bibitem{Hasan:2006}
M.~A. Hasan, V.~Chaoji, S.~Salem, and M.~Zaki.
\newblock {Link prediction using supervised learning}.
\newblock In {\em SIAM Workshop on Link Analysis, Counterterrorism and
  Security}, 2006.

\bibitem{Hoff:02}
P.~D. Hoff, A.~E. Raftery, and M.~S. Handcock.
\newblock {Latent Space Approaches to Social Network Analysis}.
\newblock {\em Journal of American Statistical Association}, 97(460), 2002.

\bibitem{Hoff:07}
P.D. Hoff.
\newblock {Modeling Homophily and Stochastic Equivalence in Symmetric
  Relational Data}.
\newblock In {\em Advances in Neural Information Processing Systems}, 2007.

\bibitem{Hoffman:olda10}
M.~Hoffman, D.~Blei, and F.~Bach.
\newblock {Online Learning for Latent {Dirichlet} Allocation}.
\newblock In {\em Advances in Neural Information Processing Systems}, 2010.

\bibitem{Hofmann:99}
T.~Hofmann.
\newblock {Probabilistic Latent Semantic Analysis}.
\newblock In {\em Proceedings of Uncertainty in Aritificial Intelligence},
  1999.

\bibitem{Jordan:99}
M.~Jordan, Z.~Ghahramani, T.~Jaakkola, and L.~Saul.
\newblock {\em {An introduction to variational methods for graphical models}}.
\newblock MIT Press, Cambridge, MA, 1999.

\bibitem{liben_nowell}
D.~Liben-Nowell and J.M. Kleinberg.
\newblock {The Link Prediction Problem for Social Networks}.
\newblock In {\em ACM Conference of Information and Knowledge Management},
  2003.

\bibitem{Lichtenwalter:2010}
R.~N. Lichtenwalter, J.~T. Lussier, and N.~V. Chawla.
\newblock {New Perspectives and Methods in Link Prediction}.
\newblock In {\em Proceedings of the 19th ACM SIGKDD Conference on Knowledge
  Discovery and Data Mining}, 2010.

\bibitem{LiuYan:09}
Y.~Liu, A.~Niculescu-Mizil, and W.~Gryc.
\newblock {Topic-Link LDA: Joint Models of Topic and Author Community}.
\newblock In {\em International Conference on Machine Learning}, 2009.

\bibitem{McAllester2003}
D.~McAllester.
\newblock {PAC-Bayesian} stochastic model selection.
\newblock {\em Machine Learning}, 51:5--21, 2003.

\bibitem{McCallum:cora2000}
A.~McCallum, K.~Nigam, J.~Rennie, and K.~Seymore.
\newblock {Automating the Construction of Internet Portals with Machine
  Learning}.
\newblock {\em Information Retrieval}, 2000.

\bibitem{Michael:IG76}
J.R. Michael, W.R. Schucany, and R.W. Haas.
\newblock {Generating Random Variates Using Transformations with Multiple
  Roots}.
\newblock {\em The American Statistician}, 30(2):88--90, 1976.

\bibitem{Miller:nips09}
K.~Miller, T.~Griffiths, and M.~Jordan.
\newblock {Nonparametric Latent Feature Models for Link Prediction}.
\newblock In {\em Advances in Neural Information Processing Systems}, 2009.

\bibitem{Cohen:08}
R.~Nallapati and W.~Cohen.
\newblock {Link-PLSA-LDA: A New Unsupervised Model for Topics and Influence in
  Blogs}.
\newblock In {\em Proceedings of International Conference on Weblogs and Social
  Media}, 2008.

\bibitem{Nallappati:2008}
R.~M. Nallapati, A.~Ahmed, E.~P. Xing, and W.~Cohen.
\newblock {Joint Latent Topic Models for Text and Citations}.
\newblock In {\em Proceeding of the ACM SIGKDD International Conference on
  Knowledge Discovery and Data Mining}, 2008.

\bibitem{Polson:arXiv12}
N.~G. Polson, J.~G. Scott, and J.~Windle.
\newblock {Bayesian Inference for Logistic Models using {Polya-Gamma} Latent
  Variables}.
\newblock {\em arXiv:1205.0310v1}, 2012.

\bibitem{Polson:BA11}
N.~G. Polson and S.~L. Scott.
\newblock {Data Augmentation for Support Vector Machines}.
\newblock {\em Bayesian Analysis}, 6(1):1--24, 2011.

\bibitem{Rosasco:04}
L.~Rosasco, E.~De Vito, A.~Caponnetto, M.~Piana, and A.~Verri.
\newblock {Are Loss Functions All the Same?}
\newblock {\em Neural Computation}, (16):1063--1076, 2004.

\bibitem{sen:aimag08}
P.~Sen, G.~Namata, M.~Bilgic, and L.~Getoor.
\newblock Collective classification in network data.
\newblock {\em AI Magazine}, 29(3):93--106, 2008.

\bibitem{Smola:vldb10}
A.~Smola and S.~Narayanamurthy.
\newblock {An Architecture for Parallel Topic Models}.
\newblock {\em International Conference on Very Large Data Bases}, 2010.

\bibitem{Strauss:1990}
D.~Strauss and M.~Ikeda.
\newblock {Pseudolikelihood Estimation for Social Networks}.
\newblock {\em Journal of American Statistical Association}, 85(409):204--212,
  1990.

\bibitem{Tanner:1987}
M.~A. Tanner and W.~H. Wong.
\newblock {The Calculation of Posterior Distributions by Data Augmentation}.
\newblock {\em Journal of the Americal Statistical Association},
  82(398):528--540, 1987.

\bibitem{Zhu:ICML12}
J.~Zhu.
\newblock {Max-Margin Nonparametric Latent Feature Models for Link Prediction}.
\newblock In {\em International Conference on Machine Learning}, 2012.

\bibitem{Zhu:ICML13}
J.~Zhu, N.~Chen, H.~Perkins, and B.~Zhang.
\newblock {Gibbs Max-margin Topic Models with Fast Sampling Algorithms}.
\newblock In {\em International Conference on Machine Learning}, 2013.

\bibitem{Zhu:nips11}
J.~Zhu, N.~Chen, and E.P. Xing.
\newblock {Infinite Latent SVM for Classification and Multi-task Learning}.
\newblock In {\em Advances in Neural Information Processing Systems}, 2011.

\bibitem{Zhu:arXiv12}
J.~Zhu, N.~Chen, and E.P. Xing.
\newblock {Bayesian Inference with Posterior Regularization and applications to
  Infinite Latent SVMs}.
\newblock {\em arXiv:1210.1766v2}, 2013.

\bibitem{Zhu:ACL13}
J.~Zhu, X.~Zheng, and B.~Zhang.
\newblock {Bayesian Logistic Supervised Topic Models with Data Augmentation}.
\newblock In {\em Proceedings of the 51st Annual Meeting of the Association for
  Computational Linguistics}, 2013.

\end{thebibliography}

\begin{IEEEbiography}[{\includegraphics[width=.9in,height=1.15in]{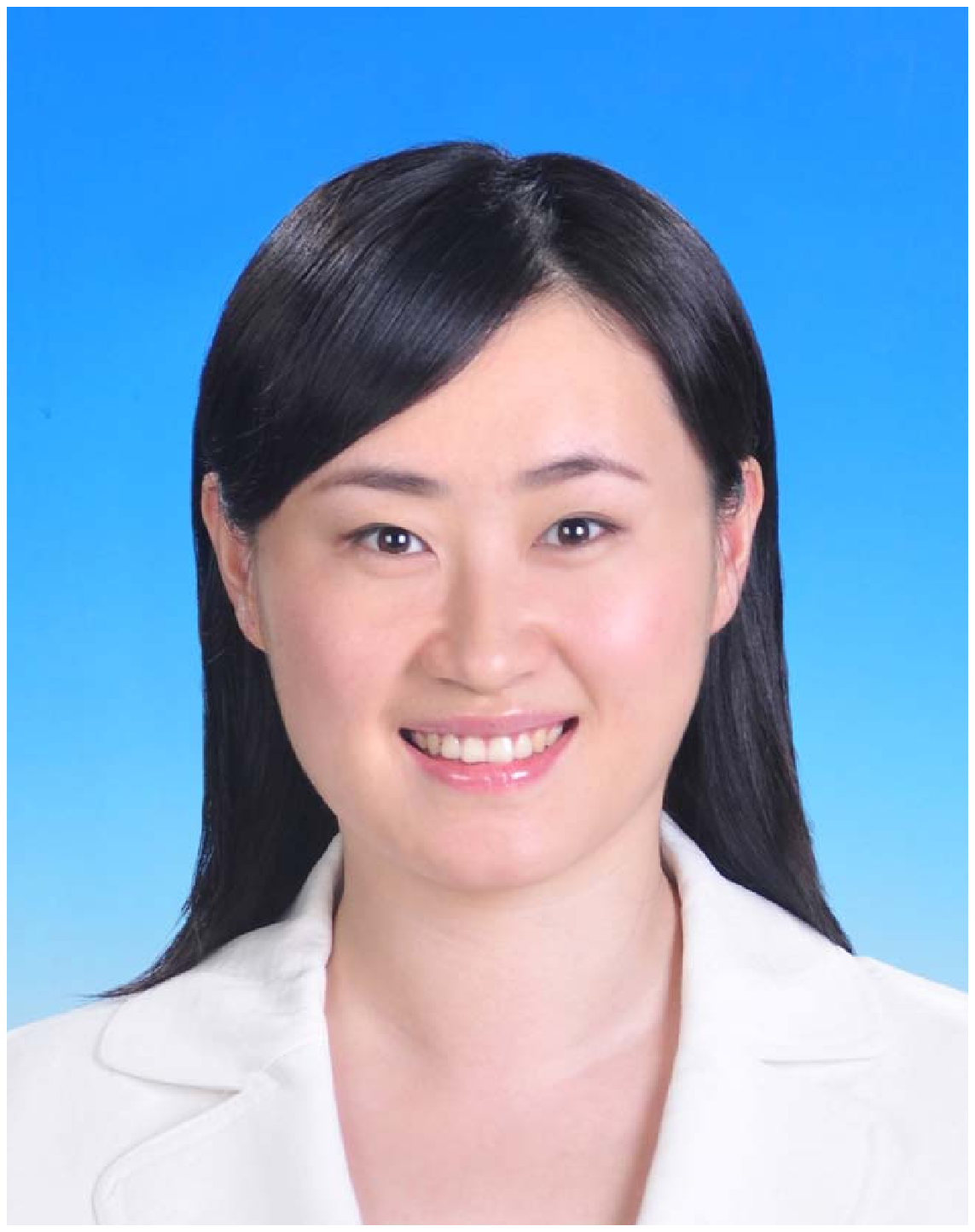}}]{Ning Chen}
received her BS from China Northwestern Polytechnical University, and PhD degree in the Department of Computer Science and Technology at Tsinghua University, China, where she is currently a post-doc fellow. She was a visiting researcher in the Machine Learning Department of Carnegie Mellon University. Her research interests are primarily in machine learning, especially probabilistic graphical models, Bayesian Nonparametrics with applications on data mining and computer vision.
\end{IEEEbiography}\vspace{-1cm}

\begin{IEEEbiography}[{\includegraphics[width=.9in,height=1.1in]{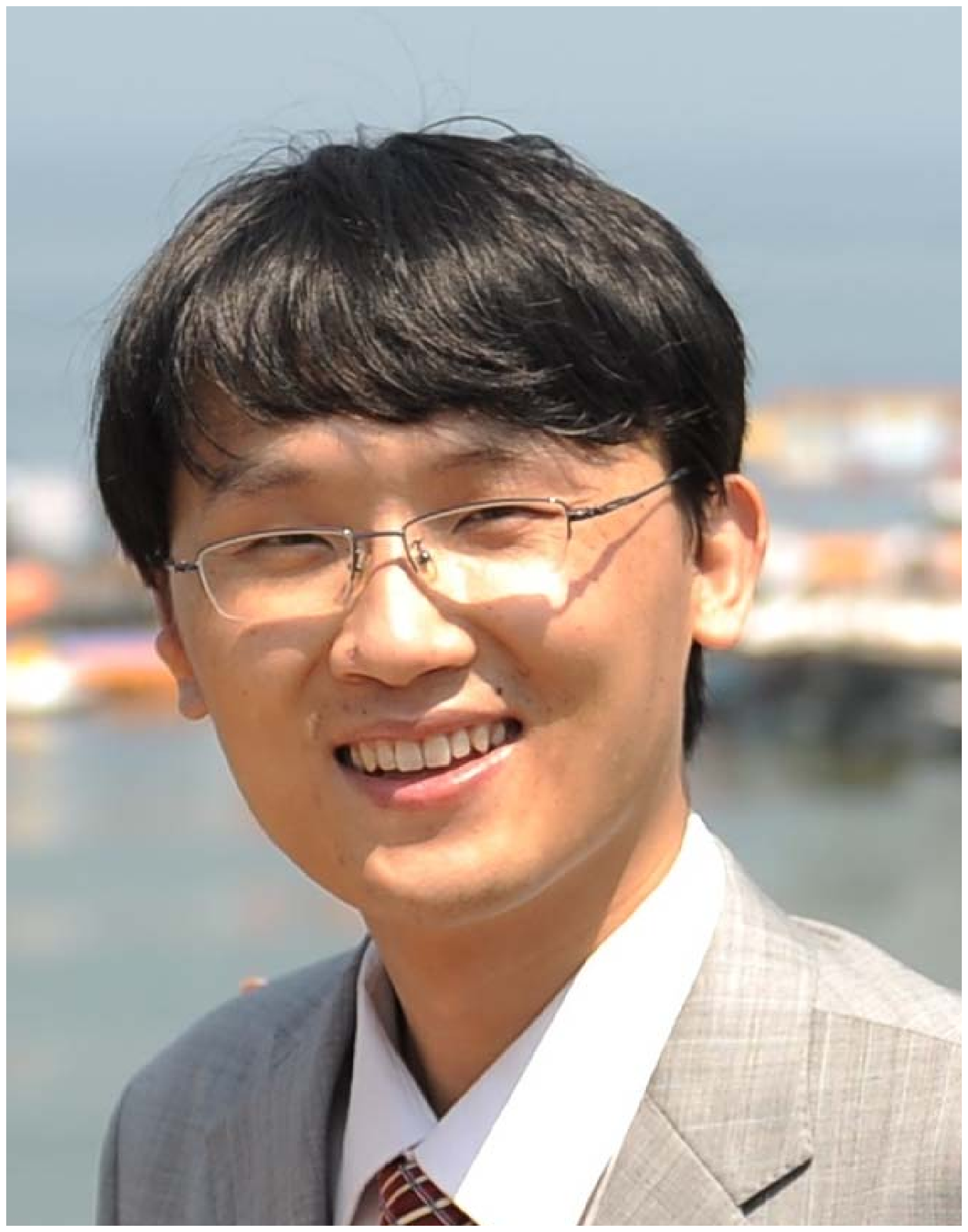}}]{Jun Zhu}
received his BS, MS and PhD degrees all from the Department of Computer Science and Technology in Tsinghua University, China, where he is currently an associate professor. He was a project scientist and postdoctoral fellow in the Machine Learning Department, Carnegie Mellon University. His research interests are primarily on developing statistical machine learning methods to understand scientific and engineering data arising from various fields. He is a member of the IEEE.
\end{IEEEbiography}\vspace{-1cm}

\begin{IEEEbiography}[{\includegraphics[width=.9in,height=1.1in]{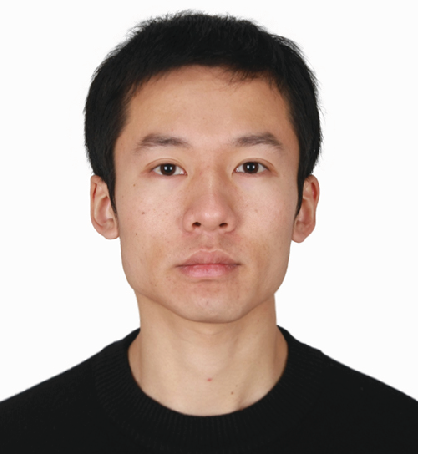}}]{Fei Xia}
received his BS from School of Software, Tsinghua University, China. He is currently working toward his MS degree in the Language Technologies Institute, School of Computer Science, Carnegie Mellon University, USA. His research interests are primarily on machine learning especially on probabilistic graphical models, Bayesian nonparametrics and data mining problems such as social networks.
\end{IEEEbiography}\vspace{-1cm}

\begin{IEEEbiography}[{\includegraphics[width=.9in,height=1.15in]{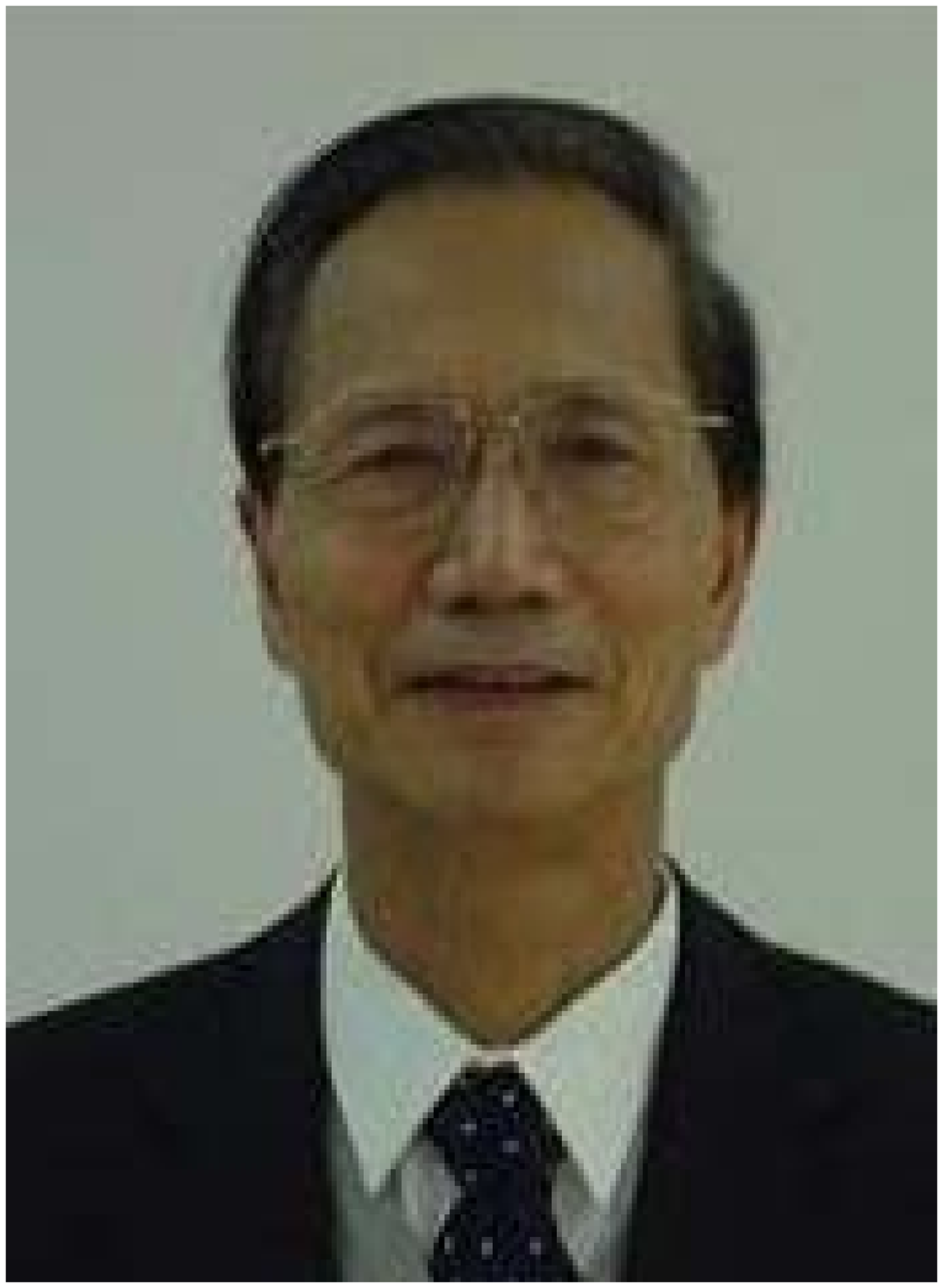}}]{Bo Zhang} graduated from the Department of Automatic Control, Tsinghua University, Beijing, China, in 1958. Currently, he is a Professor in the Department of Computer Science and Technology, Tsinghua University and a Fellow of Chinese Academy of Sciences, Beijing, China. His main interests are artificial intelligence, pattern recognition, neural networks, and intelligent control. He has published over 150 papers and four monographs in these fields.
\end{IEEEbiography}





\section*{Appendix}

In this section, we present additional experimental results.

\subsection*{A.1~~Prediction Performance on WebKB Dataset}

Fig.~\ref{fig:webkb_mmrtm} shows the 5-fold average results with standard deviations of the discriminative RTMs (with both the log-loss and hinge loss) on the WebKB dataset. We have similar observations as shown in Section~\ref{section:resultHingeLoss} on the other two datasets. Discriminative RTMs with the hinge loss (i.e., Gibbs-gMMRTM and Gibbs-MMRTM) obtain comparable predictive results with the RTMs using the log-loss (i.e., Gibbs-gRTM and Gibbs-RTM). And generalized gRTMs achieve superior performance over the diagonal RTMs, especially when the topic numbers are less than 25. We also develop Approx-gMMRTM and Approx-gRTM by using a simple approximation method in sampling $\Zv$ (see Section~\ref{section:dataset}) to greatly improve the time efficiency without sacrificing much prediction performance.

\begin{figure*}[t]
\centering
\subfigure[link rank]{\includegraphics[height=1.3in, width=1.25in]{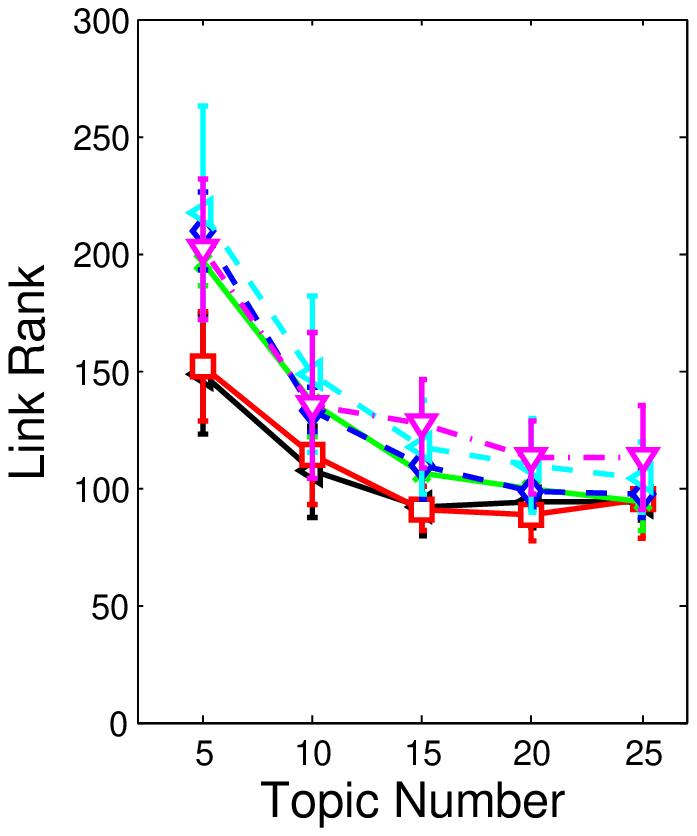}}\label{fig:webkb_mmrtm}
\subfigure[word rank]{\includegraphics[height=1.3in, width=1.25in]{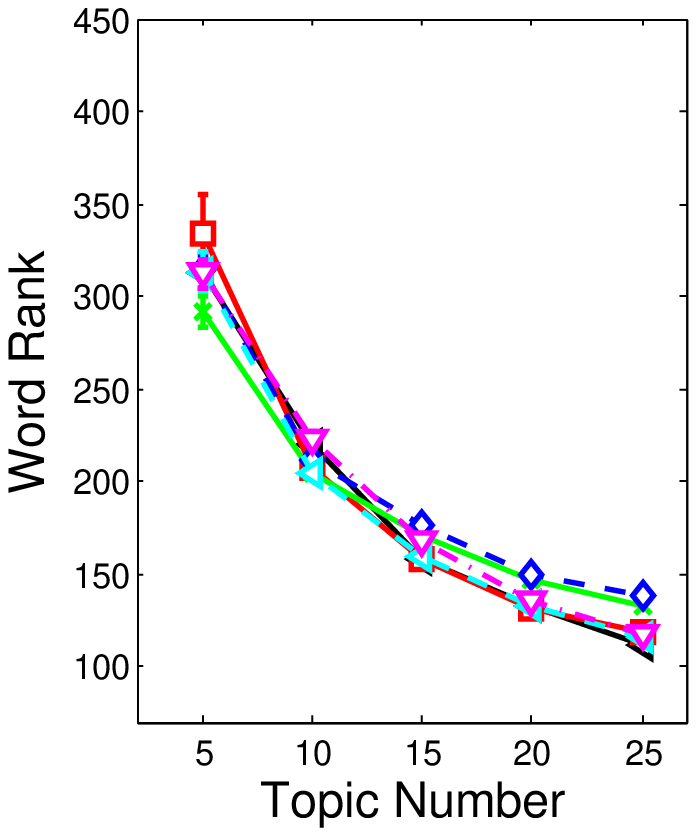}}\label{fig:webkb_mmrtm}
\subfigure[AUC score]{\includegraphics[height=1.3in, width=1.25in]{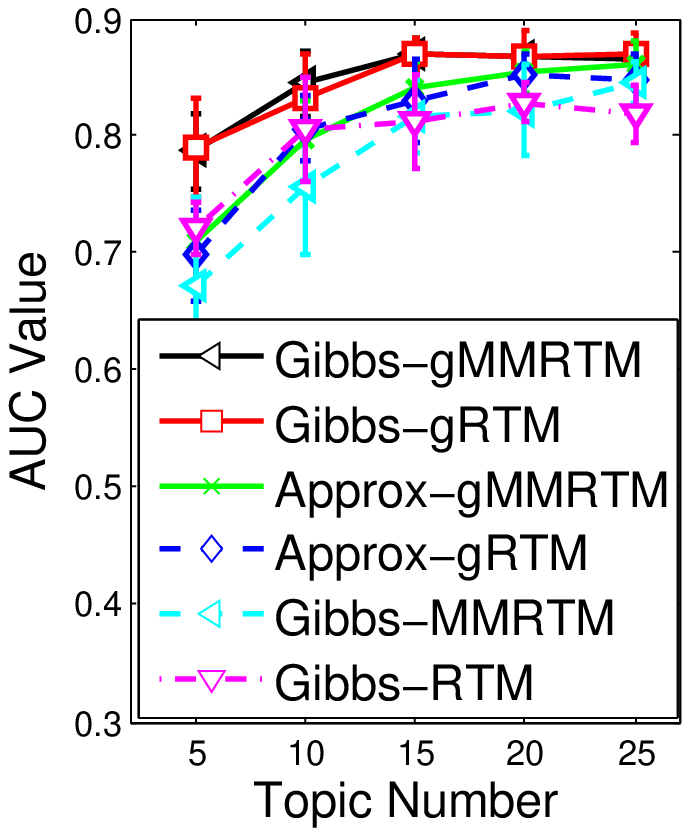}}\label{fig:webkb_mmrtm}
\subfigure[train time]{\includegraphics[height=1.3in, width=1.25in]{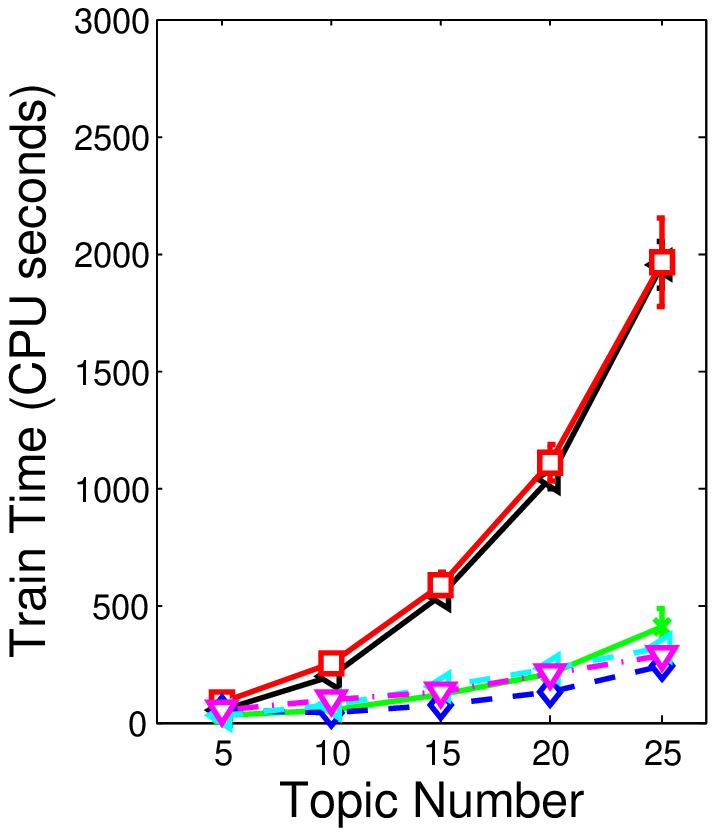}}\label{fig:webkb_mmrtm}
\subfigure[test time]{\includegraphics[height=1.3in, width=1.25in]{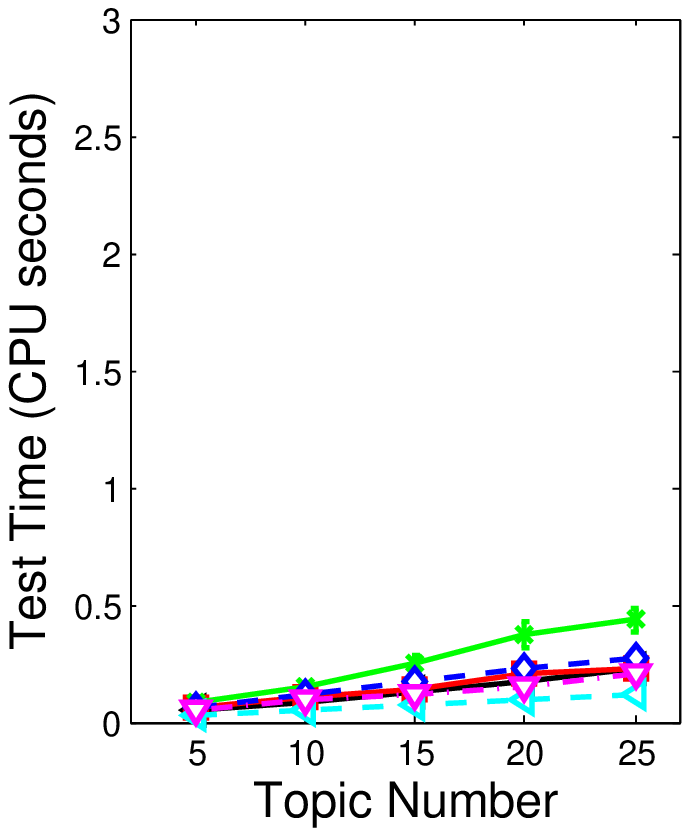}}\label{fig:webkb_mmrtm}
\caption{Results of various models with different numbers of topics on the WebKB dataset.}
\label{fig:webkb_mmrtm}
\end{figure*}

\subsection*{A.2~~Topic Discovery}
Table.~\ref{table:latentTopic_appendix} shows 7 example topics discovered by the 10-topic Gibbs-gRTM on the Cora dataset. For each topic, we show the 6 top-ranked document titles that yield higher values of $\Thetav$. In order to qualitatively illustrate the semantic meaning of each topic among the documents from 7 categories\footnote{The seven categories are {\it Case Based}, {\it Genetic Algorithms}, {\it Neural Networks}, {\it Probabilistic Methods}, {\it Reinforcement Learning}, {\it Rule Learning} and {\it Theory}.}, in the left part of Table~\ref{table:latentTopic_appendix}, we show the average probability of each category distributed on the particular topic. Note that category labels are not considered in all the models in this paper, we use it here just to visualize the discovered semantic meanings of the proposed Gibbs-gRTM. We can observe that most of the discovered topics are representative for documents from one or several categories. For example, topics T1 and T2 tend to represent documents about ``Genetic Algorithms" and ``Rule Learning", respectively. Similarly, topics T3 and T6 are good at representing documents about ``Reinforcement Learning" and ``Theory", respectively.
\begin{table*}[t]
\caption{Example topics discovered by a 10-topic Gibbs-gRTM on the Cora dataset. For each topic, we show 6 top-ranked documents as well as the average probabilities of that topic on representing documents from 7 categories. \label{table:latentTopic_appendix}}
\begin{center}
\scalebox{0.95}{\setlength{\tabcolsep}{2pt}
       \begin{tabular}{|c|l|}
       \hline
        \hline
        { Topic }& Top-6 Document Titles                          \\
        \hline
{T1: Genetic Algorithms} & {1. Stage scheduling: A tech. to reduce the register requirements of a modulo schedule.}        \\
{\multirow{5}{*}{\includegraphics[height=1.6cm, width=6cm]{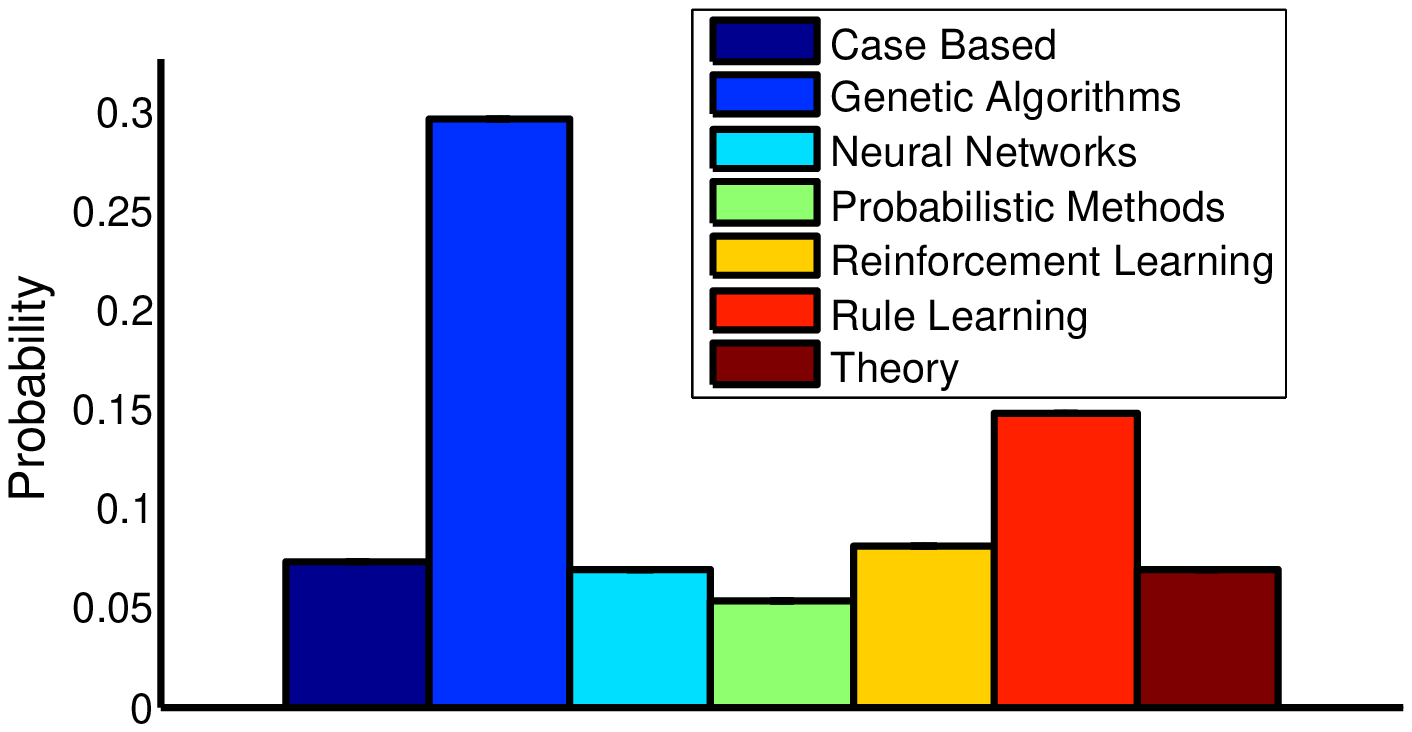}}}& {2. Optimum modulo schedules for minimum register requirements.}\\
{}& {3. Duplication of coding segments in genetic programming.}\\
{}&{4. Genetic programming and redundancy.}\\
{}&{5. A cooperative coevolutionary approach to function optimization. }\\
{}&{6. Evolving graphs and networks with edge encoding: Preliminary report. }\\
        \hline
{T2: Rule Learning} & {1. Inductive Constraint Logic.}        \\
{\multirow{5}{*}{\includegraphics[height=1.57cm, width=6cm]{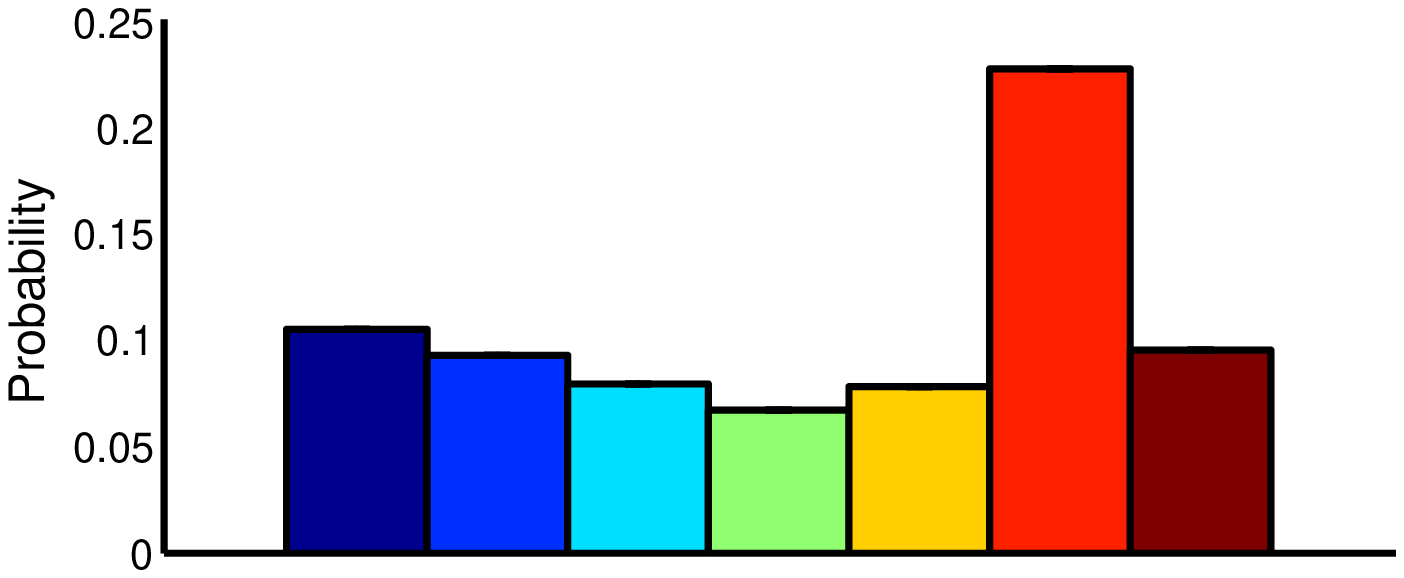}}}& {2. The difficulties of learning logic programs with cut.}\\
{}& {3. Learning se-mantic grammars with constructive inductive logic programming.}\\
{}&{4. Learning Singly Recursive Relations from Small Datasets.}\\
{}&{5. Least generalizations and greatest specializations of sets of clauses.}\\
{}&{6. Learning logical definitions from relations. }\\
        \hline
{T3: reinforcement learning} & {1. Integ. Architect. for Learning, Planning $\&$ Reacting by Approx. Dynamic Program.}        \\
{\multirow{5}{*}{\includegraphics[height=1.57cm, width=6cm]{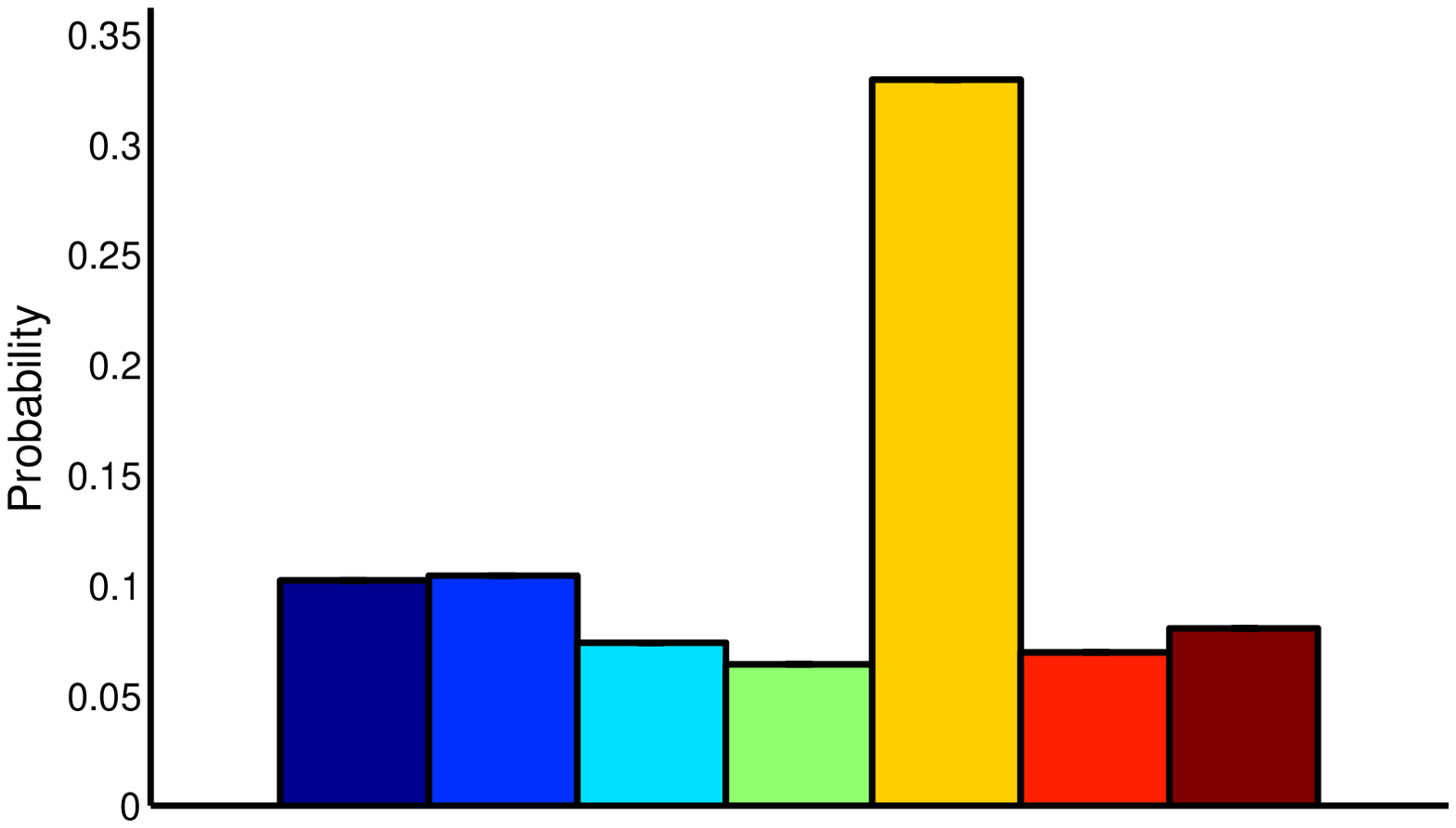}}}& {2. Multiagent reinforcement learning: Theoretical framework and an algorithm.}\\
{}& {3. Learning to Act using Real- Time Dynamic Programming.}\\
{}&{4. Learning to predict by the methods of temporal differences.}\\
{}&{5. Robot shaping: Developing autonomous agents though learning.}\\
{}&{6. Planning and acting in partially observable stochastic domains.}\\
        \hline
{T6: Theory} & {1. Learning with Many Irrelevant Features.  }        \\
{\multirow{5}{*}{\includegraphics[height=1.57cm, width=6cm]{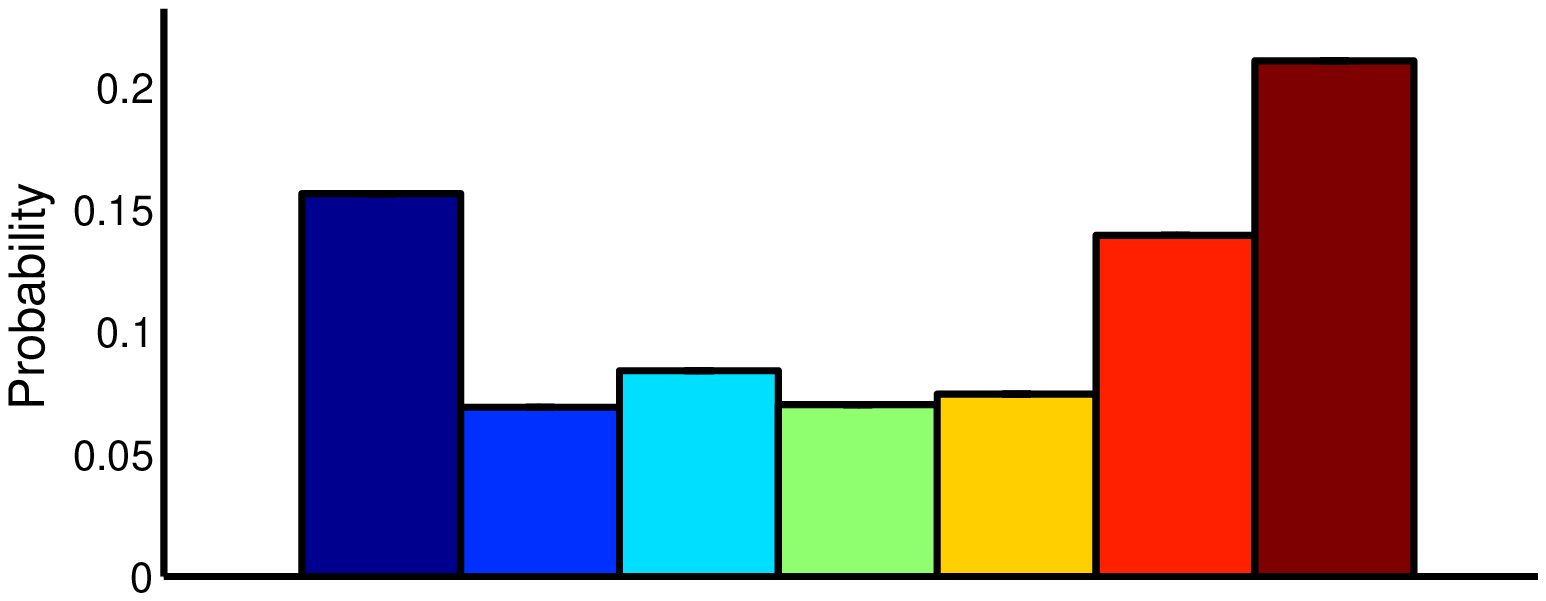}}}& {2. Learning decision lists using homogeneous rules.}\\
{}& {3. An empirical comparison of selection measures for decision-tree induction.}\\
{}&{4. Learning active classifiers.}\\
{}&{5. Using Decision Trees to Improve Case-based Learning.}\\
{}&{6. Utilizing prior concepts for learning.}\\
      \hline
{T7: Neural Networks} & {1. Learning factorial codes by predictability minimization.  }        \\
{\multirow{5}{*}{\includegraphics[height=1.57cm, width=6cm]{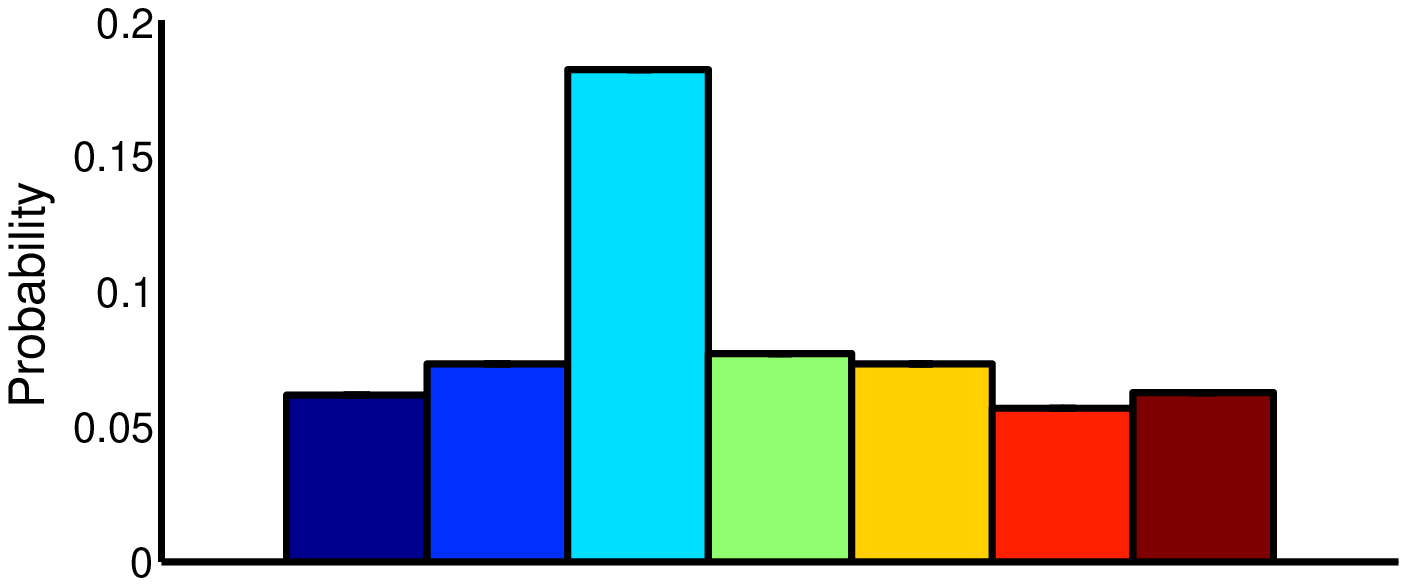}}}& {2. The wake-sleep algorithm for unsupervised neural networks.}\\
{}& {3. Learning to control fast-weight memories: An alternative to recurrent nets.}\\
{}&{4. An improvement over LBG inspired from neural networks.}\\
{}&{5. A distributed feature map model of the lexicon.}\\
{}&{6. Self-organizing process based on lateral inhibition and synaptic resource redistribution.}\\
        \hline
{T8: Probabilistic Methods} & {1. Density estimation by wavelet thresholding.}        \\
{\multirow{5}{*}{\includegraphics[height=1.57cm, width=6cm]{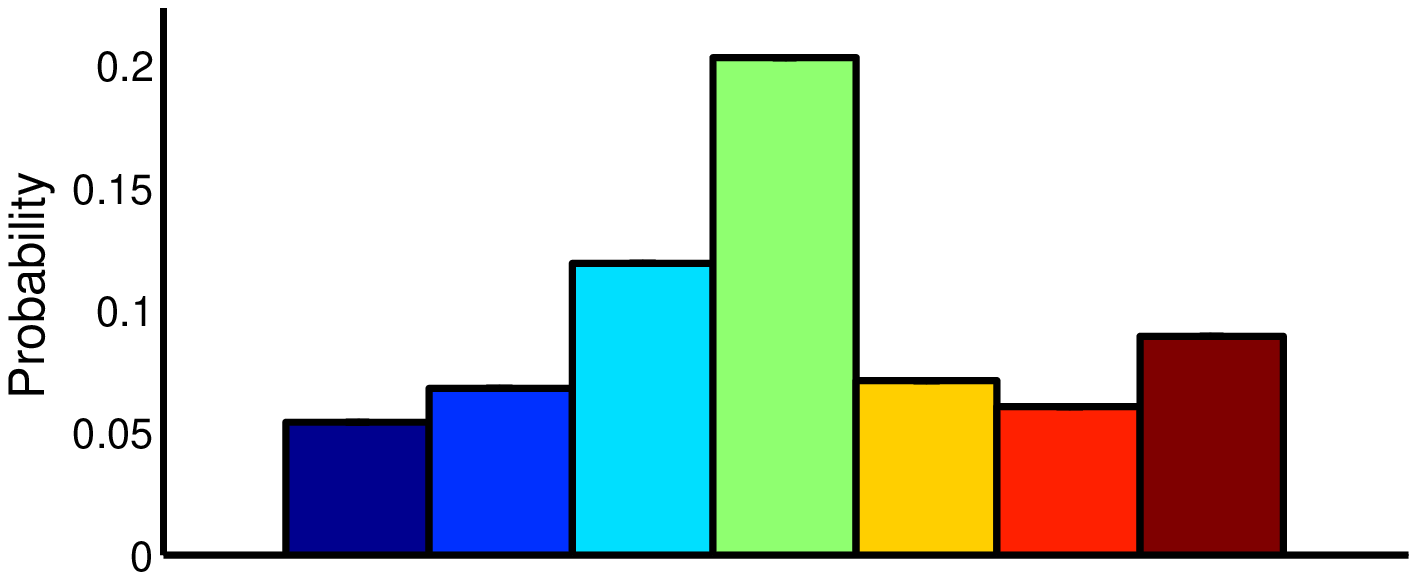}}}& {2. On Bayesian analysis of mixtures with an unknown number of components.}\\
{}& {3. Markov chain Monte Carlo methods based on "slicing" the density function.}\\
{}&{4. Markov chain Monte Carlo convergence diagnostics: A comparative review.}\\
{}&{5. Bootstrap C-Interv. for Smooth Splines $\&$ Comparison to Bayesian C-Interv.}\\
{}&{6. Rates of convergence of the Hastings and Metropolis algorithms.}\\
        \hline
{T9: Case Based} & {1. Case Retrieval Nets: Basic ideas and extensions.}        \\
{\multirow{5}{*}{\includegraphics[height=1.57cm, width=5.6cm]{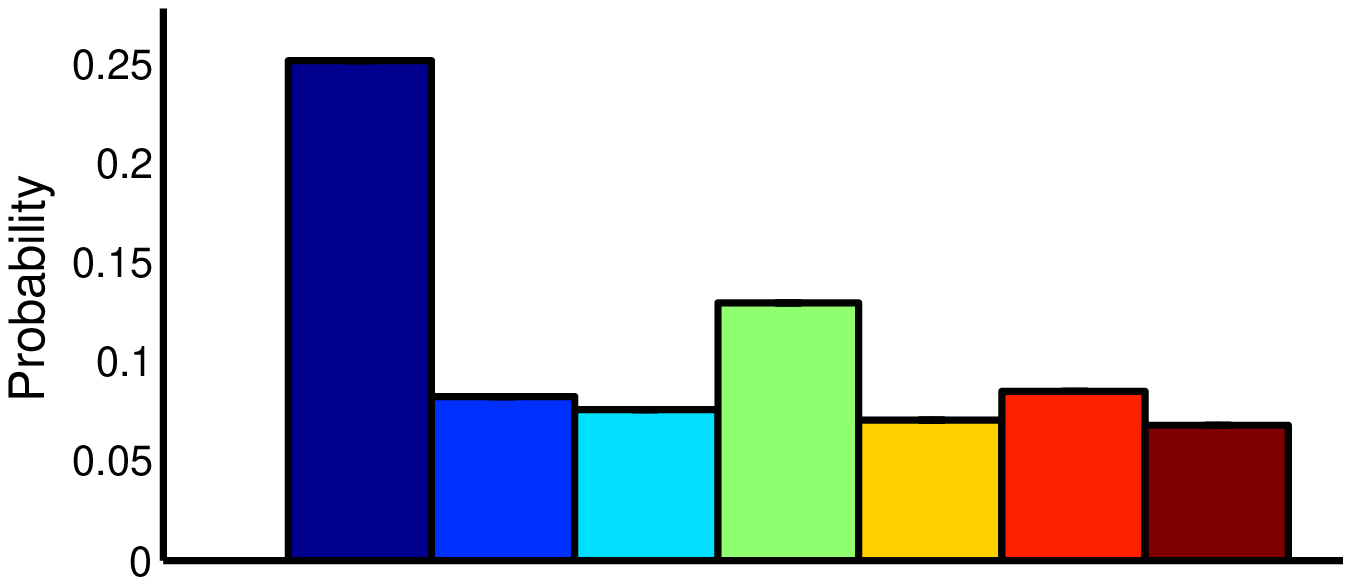}}}& {2. Case-based reasoning: Foundat. issues, methodological variat., $\&$ sys. approaches.}\\
{}& {3. Adapter: an integrated diagnostic system combining case-based and abduct. reasoning.}\\
{}&{4. An event-based abductive model of update.}\\
{}&{5. Applying Case Retrieval Nets to diagnostic tasks in technical domains.}\\
{}&{6. Introspective Reasoning using Meta-Explanations for Multistrategy Learning.}\\
   \hline
        \hline
        \end{tabular}}
\end{center}
\end{table*}

\end{document}